\documentclass[]{fairmeta}
\usepackage{amsfonts}
\usepackage{enumitem}
\usepackage{subcaption}
\usepackage{xcolor}
\usepackage{colortbl}
\usepackage[utf8]{inputenc}
\usepackage[T1]{fontenc}

\usepackage{charter}

\usepackage{tikz}
\usetikzlibrary{arrows.meta,positioning,fit,calc,decorations.pathreplacing,backgrounds}
\usepackage{pgfplots}
\pgfplotsset{compat=1.18}
\usepackage[most]{tcolorbox}
\usepackage{amssymb}
\usepackage{listings}

\RequirePackage{listings}
\RequirePackage{tcolorbox}
\tcbuselibrary{listings}  
\usepackage{xspace}
\newcommand{\name}{\textcolor{black}{KernelEvolve}\xspace}

\newcommand{\textred}[1]{\textcolor{red}{#1}}
\ifx\noeditingmarks\undefined
  \newcommand{\pgwrapper}[2]{\textred{#1: }\textit{#2}}
\else
  \newcommand{\pgwrapper}[2]{}
\fi

\definecolor{speedup}{RGB}{230, 255, 230}
\definecolor{slowdown}{RGB}{255, 220, 220}
\definecolor{highlight}{RGB}{255, 255, 200}
\definecolor{othershape}{RGB}{230, 230, 250}  
\usepackage{newunicodechar}
\newunicodechar{❌}{\textbf{[X]}}

\RequirePackage[scale=0.85]{sourcecodepro}

\definecolor{keywordgreen}{RGB}{0,128,0}       
\definecolor{builtingreen}{RGB}{0,128,0}       
\definecolor{operatormagenta}{RGB}{170,34,255} 
\definecolor{commentgray}{RGB}{96,139,78}      
\definecolor{stringred}{RGB}{186,33,33}        
\definecolor{metabg}{RGB}{248,249,250}         
\definecolor{metablue}{RGB}{0,102,204}         

\lstdefinestyle{pythonminted}{
    language=Python,
    basicstyle=\footnotesize\ttfamily\linespread{1.15}\selectfont,
    keywordstyle=\bfseries\color{keywordgreen},           
    keywordstyle=[2]\bfseries\color{operatormagenta},     
    keywordstyle=[3]\bfseries\color{builtingreen},        
    commentstyle=\itshape\color{commentgray},             
    stringstyle=\itshape\color{stringred},                
    showstringspaces=false,
    breaklines=true,
    tabsize=4,
    numbersep=8pt,
    numberstyle=\tiny\color{gray},
    morekeywords={def,class,return,yield,if,elif,else,for,while,with,as,import,from,assert},
    morekeywords=[2]{and,or,not,is,in},
    morekeywords=[3]{None,True,False}
}

\newtcblisting{pythoncode}{
    listing engine=listings,
    listing options={
        style=pythonminted,
        language=Python
    },
    colback=metabg,
    colframe=metablue,
    left=2mm,
    right=2mm,
    top=2mm,
    bottom=2mm,
    boxrule=0.5pt,
    arc=2pt,
    listing only,
    breakable,
    enhanced jigsaw
}

\definecolor{metabg}{RGB}{245, 247, 250}
\definecolor{metablue}{RGB}{59, 130, 246}
\definecolor{feat0color}{RGB}{59, 130, 246}      
\definecolor{feat1color}{RGB}{16, 185, 129}      
\definecolor{headercolor}{RGB}{239, 68, 68}      

\lstdefinestyle{mbdtstyle}{
    basicstyle=\ttfamily\scriptsize\color{black},
    frame=single,
    framerule=0.5pt,
    rulecolor=\color{metablue},
    backgroundcolor=\color{metabg},
    moredelim=[is][\color{headercolor}\bfseries]{@}{@},
    moredelim=[is][\color{feat0color}\bfseries]{!}{!},
    moredelim=[is][\color{feat1color}\bfseries]{*}{*},
    columns=fullflexible,
    keepspaces=true,
    xleftmargin=2mm,
    xrightmargin=2mm,
    aboveskip=2mm,
    belowskip=2mm,
}

\title{KernelEvolve: Scaling Agentic Kernel Coding for Heterogeneous AI Accelerators at Meta}

\author{
\vspace{5mm}
KernelEvolve Team, Meta Platforms
}

\abstract{\lipsum[1]}

\date{\today}
\metadata[ISCA 2026 Paper]{\url{https://gangliao.me/assets/pdf/kernelevolve_isca26_paper.pdf}}

\correspondence{Gang Liao: \email{gangliao@meta.com}, Carole-Jean Wu: \email{carolejeanwu@meta.com}, Gaoxiang Liu: \email{gaoxiang@meta.com\\}
}

\begin{document}

\abstract{
Making deep learning recommendation model (DLRM) training and inference fast and efficient is important. However, this presents three key system challenges – model architecture diversity, kernel primitive diversity, and hardware generation and architecture heterogeneity. The combination of the three diversity dimensions leads to a complex optimization space. 

This paper presents \name – an agentic kernel coding framework – to tackle heterogeneity at–scale for DLRM training and inference. \name is designed to take kernel specifications as input and automate the process of kernel generation and optimization for recommendation model across heterogeneous hardware architectures through multiple programming abstractions, including Triton, CuTe DSL, and low-level hardware diagnostic languages, spanning the full hardware-software optimization stack. The kernel optimization process is described as graph-based search with selection policy, universal operator, fitness function, and termination rule, dynamically adapts to runtime execution context through 
retrieval-augmented prompt synthesis. The system integrates a persistent knowledge 
base encoding hardware-specific constraints for heterogeneous AI accelerators, 
enabling effective kernel generation even for proprietary architectures absent 
from LLM training corpora.  

We designed, implemented, and deployed \name to optimize a wide variety of production recommendation models across generations of NVIDIA and AMD GPUs, as well as Meta's latest-generation AI accelerators (MTIA v3). We validate \name on the publicly-available KernelBench suite, achieving 100\% pass rate on all 250 problems across three difficulty levels, and 160 PyTorch ATen operators across three heterogeneous hardware platforms, demonstrating 100\% correctness over all 480 operator-platform configurations. 
\name reduces development time from weeks to hours and achieves substantial performance improvements by up to 17 times over PyTorch baselines across diverse production use cases and for heterogeneous AI systems at-scale.
Beyond performance efficiency improvements, \name significantly mitigates the programmability barrier for new AI hardware by enabling automated kernel generation for proprietary accelerators. We hope the insights and deployment experience presented in this paper will shed new light on the design of AI systems and optimization at-scale.}

\noindent
\begin{minipage}{\textwidth}
    KernelEvolve Technical Report
    \hfill \today \\
    \rule{\textwidth}{1pt}
\end{minipage}
\vspace{1.5em}

\maketitle

\newpage
\tableofcontents
\newpage

\section{Introduction}
\label{sec:intro}

Modern online advertising systems demand massive-scale machine learning inference under stringent sub-second latency constraints. At Meta's operational scale – serving more than hundreds of trillions of ad-ranking inferences daily 
across global data centers consuming hundreds of megawatts – the infrastructure deploys an extensive ensemble of multi-stage ranking models (retrieval, early-stage ranking, and second-stage ranking, often exceeding 1500 distinct models) to address diverse prediction tasks. These models execute across a heterogeneous fleet of AI accelerators, including Meta Training and Inference Accelerator (MTIA), AMD GPUs, and NVIDIA hardware. Within this ecosystem, computational kernels—the low-level primitive functions implementing tensor operations, data transformations, and feature engineering—constitute a dual bottleneck affecting both training and serving architecture.

From a performance perspective, ads serving operates under strict real-time constraints where microsecond-level kernel efficiency directly impacts user experience and revenue generation, as well as Meta's Total Cost of Ownership (TCO). The economic implications are substantial: marginal kernel-level performance improvements translate to multi-million dollar reductions in infrastructure operating costs while simultaneously enhancing user engagement metrics that correlate directly with advertising revenue. From an architectural perspective, kernel coverage on target hardware fundamentally determines system deployment viability. When critical operators lack native implementations on AI accelerators, production systems may architect disaggregated serving topologies where missing operators execute on separate CPU tiers, or halt deployment until kernels are implemented. This fragmentation introduces cross-node network overhead (adding millisecond-scale latency), serialization costs, reduced reliability from cascading failures. Moreover, incomplete kernel coverage blocks model launches, directly constraining algorithmic innovation and competitive differentiation. This confluence of latency sensitivity, architectural constraints, and massive deployment scale elevates kernel optimization to a first-order system imperative for ads model serving infrastructure.

\subsection{The Curse of Dimensionality}

The computational landscape of production ads ranking systems exhibits unprecedented complexity across three critical dimensions: model architecture diversity, kernel primitive diversity and hardware generation and architecture heterogeneity. This combinatorial explosion creates a fundamental scalability crisis that manual kernel development approaches cannot address.

\begin{figure}
    \centering
    \includegraphics[width=0.9\linewidth]{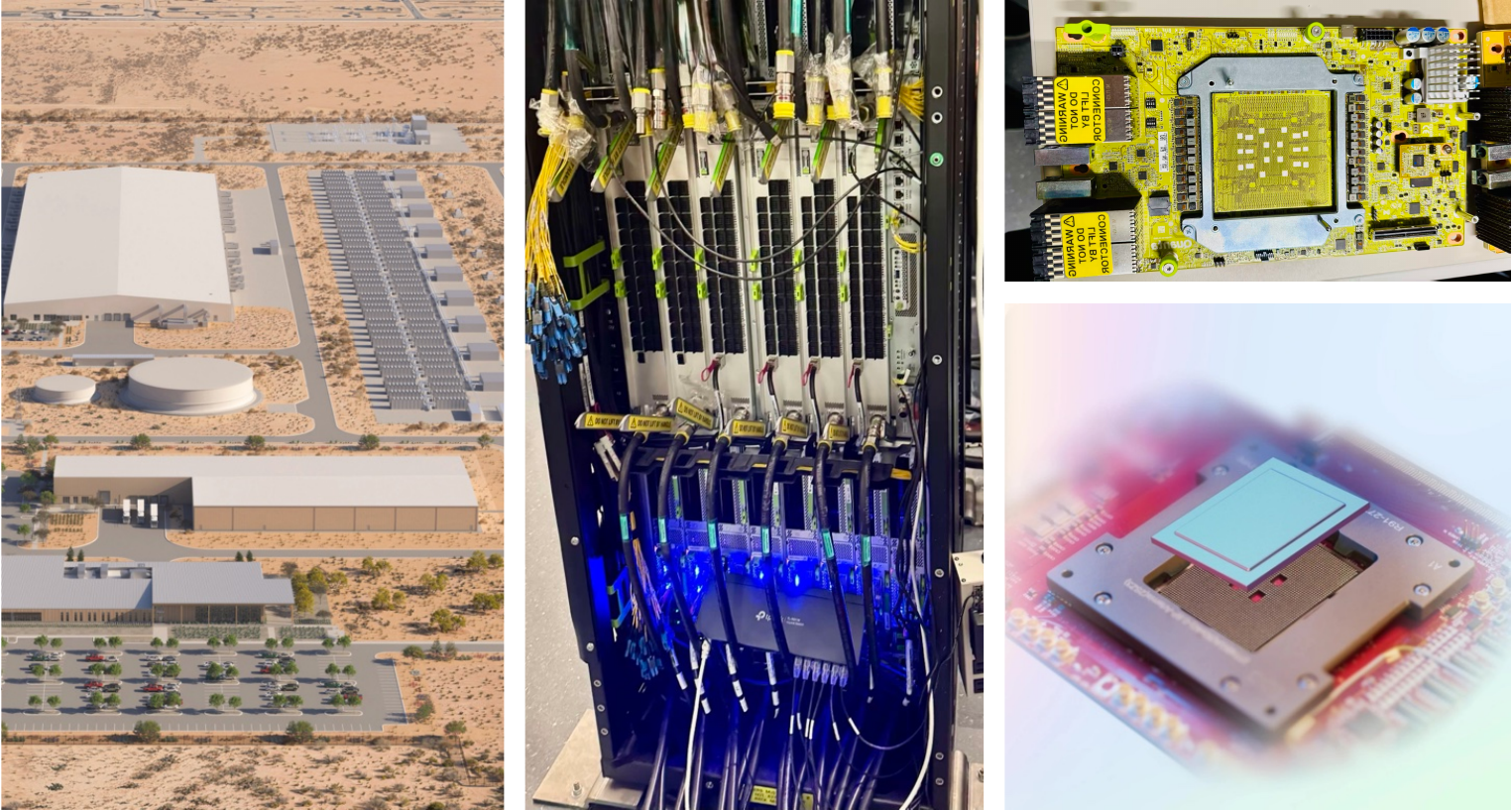}
    \caption{Meta's MTIA (Meta Training and Inference Accelerator) is a custom-designed chip optimized for AI workloads. The figure illustrates the MTIA hardware from four perspectives: its integration within a data center or server farm, highlighting the overall facility layout and environment; its deployment within a rack-mounted system for high-bandwidth applications; a close-up of the chip’s circuitry and board connections; and a detailed view of the chip core. Together, these images highlight MTIA’s advanced design, connectivity, and its role in enhancing the performance and efficiency of AI tasks across Meta’s platforms.}
    \label{fig:mtia_chips}
\end{figure}

\textbf{Hardware Diversity Across Vendors and Generations.} Production deployment spans heterogeneous accelerators including NVIDIA GPUs, AMD GPUs, and Meta's custom MTIA chips~\cite{mtia-blog} in Figure~\ref{fig:mtia_chips}—each exhibiting distinct architectural characteristics that prevent direct kernel portability. 

(1) Memory hierarchy heterogeneity: Architectures differ fundamentally in cache organization and bandwidth characteristics. NVIDIA architectures employ multi-level cache hierarchies with tens of megabytes of L2 cache, AMD architectures feature large Infinity Cache structures serving as shared L3-equivalent storage, while MTIA architectures implement custom on-chip SRAM subsystems optimized for recommendation inference patterns with distinct on-chip and off-chip bandwidth profiles. These differences necessitate architecture-specific memory access patterns and tiling strategies.

(2) Programming model fragmentation: Each platform exposes hardware through incompatible abstractions—CUDA's thread-block model, Triton's tile-based DSL~\cite{tillet2019triton}, AMD's ROCm/HIP extensions, CuTe's layout algebra for NVIDIA Hopper~\cite{nvdia-cute}, MTIA's C++ kernel DSL, and emerging frameworks like TileLang/TLX/Gluon~\cite{wang2504tilelang,tlx,gluon}. These abstractions differ not merely in syntax but in fundamental execution models—from Triton's automatic memory coalescing to CUDA's explicit shared memory management—necessitating complete algorithmic restructuring rather than syntactic translation.

(3) Generational architectural discontinuities: Even within a single vendor family, generational transitions introduce fundamentally different execution models. For example, NVIDIA's Ampere-to-Hopper evolution introduced the Tensor Memory Accelerator (TMA) for asynchronous bulk tensor transfers between global and shared memory, added a new 128-thread warp-group execution model to support WGMMA tensor operations, and exposed multiple asynchronous execution pipelines using mbarriers and producer–consumer synchronization. These features require kernel developers to adopt new pipeline structures that differ significantly from traditional warp-centric Ampere kernels.  

This hardware diversity renders manual cross-platform kernel development economically infeasible, as each hardware platform requires complete reimplementation of thousands of distinct kernel variants with hardware-specific optimization strategies that do not transfer across architectural boundaries. Each platform-specific implementation demanding 2-8 weeks of expert optimization effort. 
The resulting implementation matrix—O($\text{operators} \times \text{hardware platforms}$)—creates an unsustainable maintenance burden, further exacerbated by 12-18 month hardware generation update cycles that invalidate existing optimizations and necessitate complete kernel redesigns to exploit new architectural features.

\textbf{Model Diversity Across Ranking Stages.} As shown in Table~\ref{table:ranking_stage}, modern ads ranking employs multi-stage architectures where computational characteristics vary dramatically across pipeline stages. Retrieval models process millions of candidates using lightweight scoring functions optimized for both throughput and latency, employing approximate nearest neighbor search and efficient embedding operations. Early-stage ranking models apply moderate-complexity neural networks to thousands of candidates, balancing computational cost against filtering accuracy through pruned architectures and quantized operations. Second-stage ranking models execute heavyweight deep neural networks on hundreds of candidates with strict sub-100ms latency requirements, increasingly incorporating Transformer-based architectures that introduce distinct computational patterns.  
These Transformer-based ranking models~\cite{zhai2024actions,zeng2025interformer,zhang2024wukong} introduce 10-100$\times$ computational complexity increases per request compared to traditional dense architectures, requiring specialized attention kernels and sequence processing operations. Orthogonally, production recommendation models employ significantly larger embedding tables—often exceeding 100GB—that stress memory capacity and bandwidth, necessitating specialized embedding lookup and aggregation kernels optimized for irregular memory access patterns.

Each stage demands distinct kernel optimization strategies: retrieval favors batched operations maximizing memory bandwidth utilization; early-stage ranking requires balanced compute-memory workloads; final-stage ranking demands maximum single-request performance through aggressive operator fusion and specialization. This architectural diversity—spanning from traditional MLPs to attention-based sequence models—prevents one-size-fits-all kernel optimization approaches and necessitates model-aware kernel generation strategies that can synthesize implementations for both established operators and rapidly emerging architectural patterns.

\begin{table*}[t]
    \centering
    \small
    \begin{NiceTabular}{ccccc}
    \CodeBefore
    \Body
    \toprule
        \textbf{Ranking Stages} & \textbf{Description} & \textbf{Production Models} & \textbf{Total Models} & \textbf{Model Complexity}\\
    \midrule
         Early stage ranking & Rank 10K ads & $\ge$ 150 models & $\ge$ 500 models & 0.01-0.1 GFLOPS/request \\
         Late stage ranking$^{\dagger}$ & Rank 100 ads & $\ge$ 200 models & $\ge$ 1000 models & 0.2-2 GFLOPS/request \\
    \bottomrule
    \end{NiceTabular}
    \caption{Examples of some production models distribution at Meta. $^{\dagger}$Transformer-based sequence models require significantly higher computational complexity at $\sim$80 GFLOPS/request, representing a 10-100× increase over traditional dense ranking models.}
    \label{table:ranking_stage}
\end{table*}



\textbf{Kernel Diversity Beyond GEMM.} While dense matrix multiplication (GEMM) operations benefit from mature, highly-optimized libraries (cuBLAS, FBGEMM, DeepGEMM)~\cite{fbgemm,deepgemm}, production ads ranking workloads exhibit fundamentally different computational characteristics that necessitate broader kernel coverage. Ads ranking models execute over 200 distinct data preprocessing operators as integral components of model inference pipelines, where raw features undergo transformation before feeding subsequent neural network layers. These transformations~\cite{zhao2022understanding} span feature derivation (bucketization, set operations, n-gram hashing), dense normalization (variance-stabilizing transformations, one-hot encoding), and sparse normalization (cryptographic hashing with modulo reduction and type downcasting to map categorical features to embedding indices, top-k truncation with score-based ranking). 

While individually exhibiting low arithmetic intensity compared to GEMM operations, kernel availability for these preprocessing operators fundamentally determines deployment architecture. The lack of optimized preprocessing kernels on AI accelerators creates a binary constraint: without native accelerator implementations, models become ineligible for unified accelerator deployment. Executing preprocessing on accelerator host CPUs proves infeasible: (1) host CPUs in accelerator servers are underprovisioned compared to dedicated CPU tiers, lacking capacity for preprocessing-intensive workloads; (2) heterogeneous preprocessing demands—varying compute and memory bandwidth requirements—contend with accelerator I/O and system management for limited host resources; (3) allocating additional host CPU capacity negates accelerator consolidation's economic and power efficiency benefits.

\begin{table*}[t]
    \small
    \centering
    \setlength{\tabcolsep}{22pt}
    \begin{NiceTabular}{ccccc}
    \CodeBefore
    \Body
    \toprule
            \multicolumn{5}{c}{Serving Paradigm 1: Overall Latency Ms} \\
  \midrule
        \textbf{Compute Tier} & \textbf{P50} & \textbf{P75} & \textbf{P90} & \textbf{P99} \\
      \midrule
        Client $\rightarrow$ MTIA Tier & 39 & 44 & 46 & 61 \\
    \toprule
        \multicolumn{5}{c}{Serving Paradigm 2: Overall Latency Ms} \\
  \midrule
        \textbf{Compute Tier} & \textbf{P50} & \textbf{P75} & \textbf{P90} & \textbf{P99} \\
    \midrule
        $\alpha$: Client $\rightarrow$ CPU Tier & 58 & 65 & 73  & 97 \\
      \midrule
        $\beta$: CPU Tier $\rightarrow$ MTIA Tier & 42 & 48 & 51 & 57 \\
      \midrule
        $\gamma$: Data Preproc Execution & 4 & 7 & 10 & 16 \\
       \midrule
        $\delta$: Extra Network Latency & \multicolumn{4}{c}{\textbf{$\delta = \alpha - \beta - \gamma = 10\sim20 ms$}} \\
    \bottomrule
    \end{NiceTabular}
    \caption{Latency comparison of monolithic versus disaggregated serving paradigms for a production MTIA model. Paradigm 1 executes preprocessing client-side, achieving 61ms P99 latency with direct client$\rightarrow$remote MTIA communication. Paradigm 2 introduces a dedicated CPU tier for preprocessing scalability, incurring additional network hops that increase P99 latency to 97ms. The extra network latency ($\delta = \alpha - \beta - \gamma \approx 10\sim20$ms) represents pure architectural overhead with no computational benefit, demonstrating the cost of disaggregated serving when preprocessing operators lack native accelerator implementations.}
    \label{table:mtia_model}
\end{table*}

This constraint forces disaggregated serving architectures partitioning models across compute tiers: preprocessing on CPU servers, neural network computation on accelerators (GPU, AMD, MTIA). While enabling independent tier scaling—a desirable microservice property~\cite{audibert2023tf,murray2021tf}—disaggregation \textbf{imposes severe penalties} for latency-critical models at multi-megawatt serving capacity. 
Table~\ref{table:mtia_model} quantifies these costs for a production MTIA deployment. Paradigm 1 executes preprocessing client-side before accelerator invocation (P99: 61ms). Paradigm 2 introduces a dedicated CPU preprocessing tier, incurring additional network hops increasing P99 to 97ms—a 25\% degradation (compared to expected $57+16\approx73$ms with zero network overhead). The 10-20ms network overhead represents pure architectural tax consuming a substantial portion of the sub-100ms latency budget required for ads serving. This fragmentation compounds through: (1) cross-node serialization and communication costs; (2) cascading failure modes across service tiers; (3) operational complexity from synchronized deployments and version compatibility; (4) increased TCO from redundant CPU infrastructure.

The architectural implications prove severe: preprocessing kernel absence creates a binary deployment constraint rather than incremental performance loss. A single missing preprocessing operator—regardless of arithmetic intensity—blocks entire model deployments on new accelerators, forcing disaggregated architectures with system-level costs exceeding individual kernel inefficiency. This constraint paradoxically elevates preprocessing operators to equal or greater priority than compute-intensive kernels: while suboptimal GEMM implementations degrade performance incrementally, missing preprocessing kernels impose architectural penalties preventing monolithic deployment entirely. Consequently, comprehensive kernel coverage enabling monolithic accelerator deployment—where preprocessing and neural network computation co-locate—emerges as a first-order requirement for cost-effective serving. This architectural imperative motivates \name's automated generation framework synthesizing optimized implementations across the complete operator-hardware matrix, treating data preprocessing as first-class optimization targets alongside traditional compute kernels.
\begin{figure}
    \centering
    \includegraphics[width=0.6\linewidth]{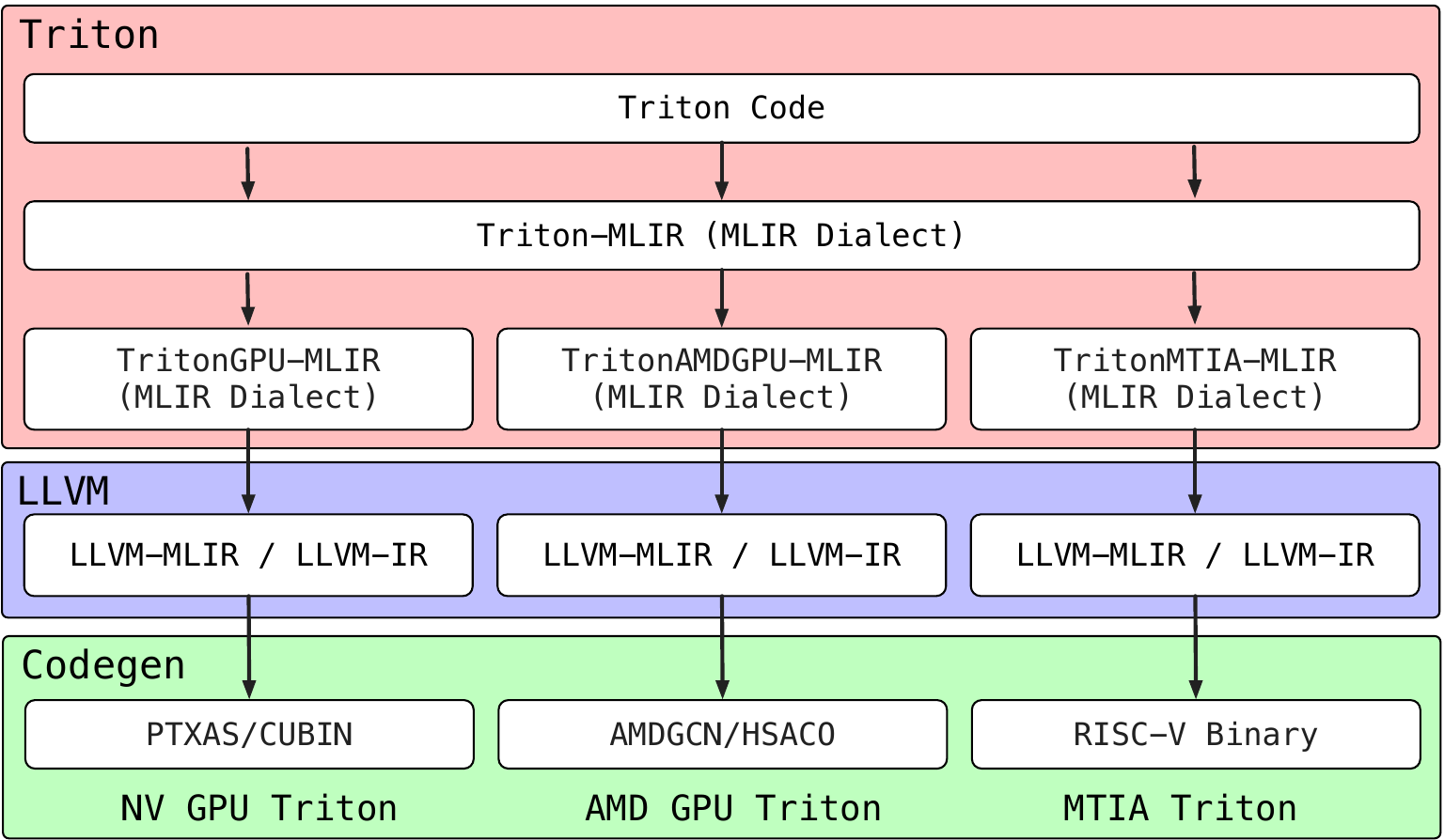}
\caption{Triton multi-target compilation architecture. Source code transforms through progressive MLIR lowering stages—platform-independent Triton-MLIR, hardware-specific GPU/AMDGPU/MTIA dialects, LLVM-IR—generating native binaries for NVIDIA (PTX/CUBIN), AMD (AMDGCN/HSACO)~\cite{amd2025triton}, and MTIA (RISC-V) platforms.}
\label{fig:triton_compilation}
\end{figure}

\textbf{Emerging AI-Powered Kernel Authoring.} Recent advances in large language models have catalyzed a wave of AI-powered coding agent systems, with several targeting GPU kernel optimization. Academic efforts include KernelBench~\cite{ouyang2025kernelbench}, which benchmarks LLM capabilities across three difficulty levels (operator, kernel fusion, full model) spanning canonical GPU operators, and AutoTriton~\cite{li2025autotriton} from Tsinghua, which applies reinforcement learning to Triton programming. Industry initiatives span multiple organizations: Meta's KernelLLM~\cite{kernelllm2025} and CWM~\cite{carbonneaux2025cwm} explore code generation with world models; Amazon's TritonRL~\cite{woo2025tritonrl} trains LLMs for Triton synthesis; AMD's GEAK-agent~\cite{wang2025geak} targets Triton kernel generation with agentic workflows on AMD MI300X and MI250; Cognition AI's Kevin~\cite{baronio2025kevin} employs multi-turn RL for CUDA kernel generation. These systems demonstrate that AI agents can generate competitive implementations on isolated OSS benchmarks through iterative refinement and learned optimization strategies.

However, these remain research prototypes with fundamental limitations for production deployment. (1) narrow optimization scope: systems target isolated subproblems—AutoTriton focuses on RL post-training for Triton, Kevin optimizes CUDA generation—without addressing end-to-end kernel lifecycle management from synthesis to deployment. (2) synthetic evaluation: benchmarks use canonical operators with static tensor shapes rather than production workloads exhibiting dynamic batching, variable sequence lengths, and domain-specific transformations (e.g., jagged tensor operations, cryptographic hashing for feature engineering). (3) single-platform focus: most target homogeneous NVIDIA environments without cross-platform synthesis for heterogeneous accelerator fleets (NVIDIA, AMD, custom ASICs). (4) limited agent capabilities: existing systems lack fully autonomous workflows encompassing automated synthesis, multi-level correctness verification (unit tests, integration tests, numerical accuracy), hierarchical profiling feedback (system, kernel and intra-kernel granularities), and persistent knowledge bases via filesystem that enable context-aware prompt synthesis by dynamically retrieving relevant optimization patterns, hardware specifications, and historical profiling data. (5) absence of inference-time scaling: no system employs large-scale search strategies (greedy, Monte Carlo Tree Search, evolutionary algorithms) that iterate hundreds to thousands of optimization steps per kernel, a capability critical for achieving expert-level performance. (6) no checkpointing support: systems restart from scratch on failure rather than resuming from intermediate states, making multi-hour optimization runs brittle and resource-inefficient. These gaps—spanning workload realism, hardware heterogeneity, agent autonomy, search scalability, and operational robustness—prevent existing prototypes from meeting production requirements.

\begin{figure}
    \centering
    \includegraphics[width=1\linewidth]{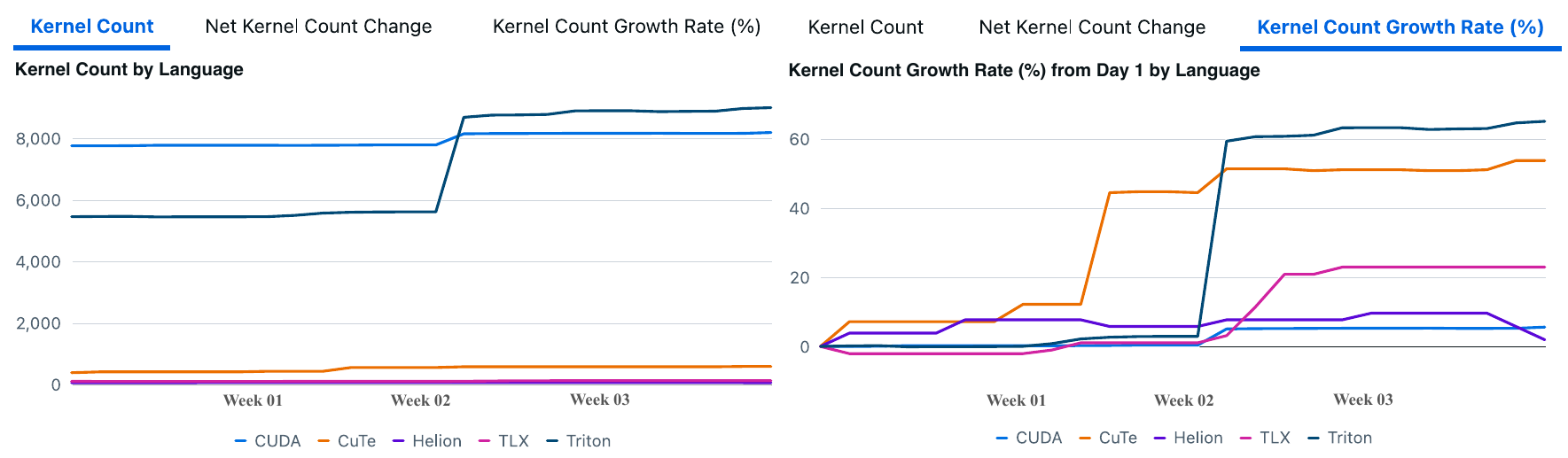}    
\caption{Triton overtakes CUDA as dominant kernel programming model at Meta. Left: Triton has grown to over 8,000 kernels, surpassing CUDA's stagnant legacy codebase, while emerging DSLs (CuTe, TLX, Helion) remain under 600. Right: Growth trajectories show Triton's 60\% expansion rate driving this transition, with CuTe at 50\% following November deployment. This shift toward higher-level DSLs—while maintaining legacy CUDA and introducing new abstraction (TLX)—creates programming model fragmentation across 5+ languages, motivating \name's automated synthesis approach.}
\label{fig:kernel_languages}
\end{figure}

\textbf{\name.} To address these fundamental limitations, we propose \name, an agent-based framework that automates the generation and optimization of high-performance compute kernels for ads ranking model serving across heterogeneous hardware architectures. Responding to the programming model evolution in Figure~\ref{fig:kernel_languages}—where Triton has emerged as the dominant DSL with 60\% growth—\name adopts Triton as its primary target, capitalizing on its cross-platform support across NVIDIA, AMD, and MTIA accelerators (see Figure~\ref{fig:triton_compilation}). We additionally target Triton-TLX, enabling low-level hardware-specific tuning while maintaining Triton's portability advantages.

\begin{figure}
    \centering
    \includegraphics[width=1\linewidth]{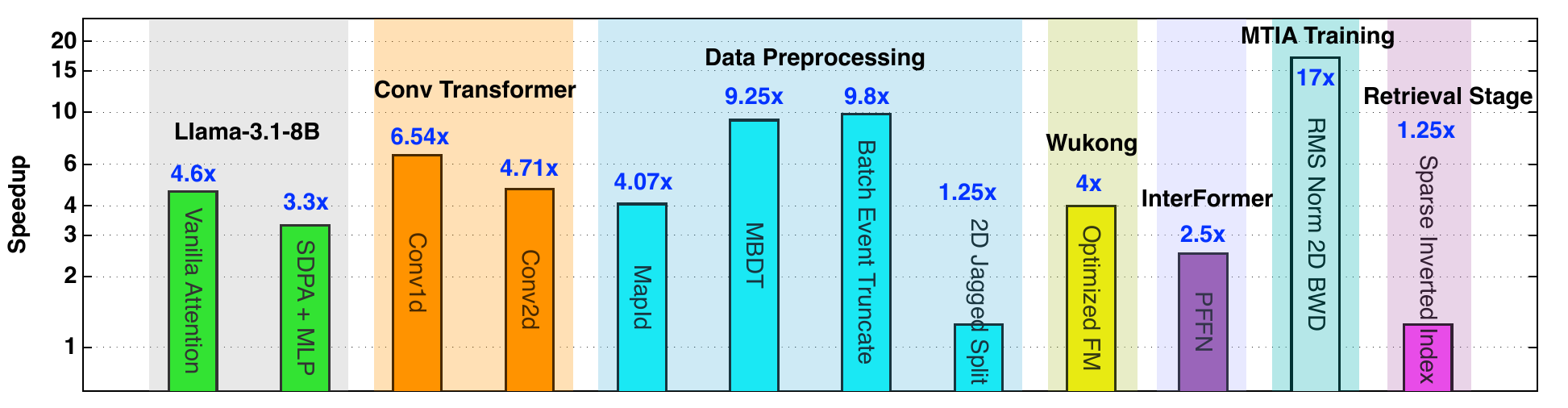}
    \caption{\name achieves 1.25-17× speedups across Meta LLMs and production use cases, spanning convolutional Transformers, data preprocessing operators, and recommendation systems, over heterogeneous AI hardware.}
    \label{fig:kernel-evolve-speedups}
\end{figure}

Figure~\ref{fig:kernel-evolve-speedups} demonstrates \name's effectiveness across diverse production workloads and hardware platforms. \name achieves substantial speedups spanning LLM inference workloads (Llama-3.1-8B: Vanilla Attention 4.6×, SDPA-MLP 3.3×), convolutional transformers (conv1d: 6.5×, conv2d: 4.7×), memory-bound data preprocessing operators critical for model enablement (MapId: 4.1×, MBDT: 9.3×, Batch Event Truncate: 9.8×), compute-intensive fusion kernels in ranking models (WuKong Optimized FM: 4.0×, InterFormer PFFN: 2.5×), MTIA-specific optimizations (RMSNorm 2D backward: 17×), and retrieval operations (Sparse Inverted Index: 1.25×). These results validate \name's ability to generate production-grade implementations across the full spectrum of recommendation and LLM inference workloads—from compute-bound fusion to memory-bound preprocessing—while addressing hardware-specific optimization challenges on proprietary accelerators. Detailed evaluation and analysis are presented in Section~\ref{sec:eval}.

\name leverages agentic AI capabilities to establish an end-to-end kernel generation service that autonomously handles kernel synthesis, compilation, profiling, correctness verification, and performance benchmarking. This approach fundamentally transforms traditional kernel development from a manual, expertise-dependent process to an automated, scalable service maintaining production-grade performance standards while adapting to evolving model architectures and hardware diversity.

This paper makes the following contributions:

\begin{itemize}[leftmargin=10pt]
\item 
We present \name, the first production-grade AI-powered kernel optimization system deployed at industrial scale for recommendation model training and inference. \name operates continuously in Meta's production infrastructure, autonomously generating optimized Triton kernels for hundreds of models serving billions of users daily. Our agent-based approach achieves production-grade reliability while exploring optimization strategies infeasible through manual development, and demonstrates how structured knowledge bases encoding hardware-specific constraints enable effective kernel generation for proprietary architectures absent from LLM training corpora.

\item We demonstrate \name's autonomous kernel optimization achieving competitive performance with expert manual implementations while reducing development time from weeks to hours. Through diverse production use cases spanning ads training and serving workloads, \name-generated kernels achieve 1.2 to 17 times speedup over the PyTorch baselines, demonstrating that automated synthesis can exceed state-of-the-art compiler-generated code for domain-specific operators. We analyze optimization trajectories across diverse operator types and hardware platforms, identifying patterns in successful kernel generation strategies that inform future automated optimization systems.

\item  We share insights from operating \name in production environments, including failure mode analysis, debugging strategies for incorrect kernel generation, performance validation methodologies ensuring production safety, and organizational integration patterns for adopting automated kernel generation workflows.

\end{itemize}


\section{Background}
\subsection{Meta's Recommendation Model}
\label{subsec:ads-models}

Meta's products—Facebook, Instagram, Reels, Threads—rely on deep learning recommendation models (DLRM)~\cite{naumov2019deep} delivering personalized content including advertisements, short videos, and posts. Llama-based generative AI additionally powers advertising features such as image and text generation~\cite{dubey2024llama}. Recommendation infrastructure processes trillions of daily inferences through multi-stage pipelines exhibiting substantial architectural diversity driven by stage-specific computational and latency constraints.

\textbf{Multi-Stage Ranking Pipeline.} The recommendation funnel comprises three stages with distinct characteristics: Retrieval processes millions of candidates narrowing to 10K-100K items through low-complexity models operating at large batch sizes with significant preprocessing overhead; Early-stage ranking refines thousands of candidates to hundreds using moderate-complexity models; Late-stage ranking scores hundreds of candidates with heavyweight models (up to 2 GFLOPS per sample) incorporating high-order feature interactions through hierarchical architectures (DHEN~\cite{zhang2022dhen}, Wukong~\cite{zhang2024wukong}). Production late-stage models exhibit over 60× complexity variation reflecting diverse quality-latency trade-offs.

\textbf{DLRM Architecture.} Traditional recommendation models combine sparse feature embeddings (categorical inputs: post IDs), dense feature transformations via MLP (continuous values: age, time), and interaction layers modeling cross-feature dependencies. Recent architectures incorporate Transformer-like components for sequence modeling—HSTU~\cite{zhai2024actions} processes user history through jagged attention mechanisms, while InterFormer~\cite{zeng2025interformer} enables bidirectional information flow between sequential and non-sequential features through personalized attention—introducing 10-100× per-request complexity increases versus traditional DLRM while demanding larger embedding tables stressing memory bandwidth.

\textbf{Emerging Generative Recommendation.} Recent recommendation systems explore generative approaches treating recommendation as sequence generation. Architectures like OneRec~\cite{zhou2025onerec,zhou2025onerecv2} employ quantization techniques (RQVAE~\cite{rajput2023recommender}, RQ-Kmean~[\cite{luo2025qarm})] converting continuous embeddings into discrete semantic IDs for LLM-based modeling. These introduce computational patterns—autoregressive decoding, token-based retrieval, quantization operations—requiring kernel support beyond traditional DLRM operators.

This architectural diversity creates heterogeneous kernel optimization requirements. Retrieval prioritizes throughput at large batch sizes; late-stage ranking demands compute-intensive fused interactions under sub-100ms latency; sequence models require jagged tensor operations for variable-length histories; generative recommendation introduces quantization and autoregressive decoding primitives. Each stage combines these specialized operators with data preprocessing transformations, creating a diverse operator landscape requiring hundreds of optimized kernel implementations.

\subsection{Data Preprocessing Operations}
\label{subsec:data-preprocessing}

Data preprocessing transforms raw features into model-ready inputs, executing transformations on batched data before feeding neural network stages. During training, preprocessing executes on distributed workers~\cite{zhao2022understanding}; during inference, preprocessing operators are embedded within model modules as integral components—executing synchronously with model invocation under strict latency constraints where preprocessing latency directly impacts end-to-end inference response time. 

\textbf{Feature Types and Memory Layout.} Ads ranking models consume four primary feature types with distinct memory representations~\cite{liao2025bullion}: (1) Dense features (\texttt{float32}) encode continuous user/item attributes (age, click-through rates, engagement scores); (2) Sparse features (\texttt{list<int64>}) represent categorical IDs (page IDs, post IDs) as variable-length lists requiring jagged tensor support; (3) Weighted sparse features (\texttt{list<\texttt{pair<int64, float32>}>}) associate scores with IDs for ranking-based operations; (4) Event-based features capture user history events as temporal sequences, increasingly adopted in production models for improved prediction quality through sequential interaction modeling.

\textbf{Preprocessing Transformation Categories.} Production preprocessing pipelines execute three operation classes: (1) Dense normalization applies statistical transformations (BoxCox, Logit) and one-hot encoding to grouped features, followed by linear scaling (shift, multiplication) across the feature matrix; (2) Sparse processing truncates ID lists via top-k selection (optionally sorted), applies cryptographic hashing mapping IDs to embedding table indices, and downcasts hashed values from int64 to int32 for memory efficiency; (3) Feature derivation computes new features from existing ones through complex operator compositions—bucketizing continuous values into categorical bins, set operations across multiple sparse lists, n-gram hashing for text features. Feature derivation comprises hundreds of operator branches in abstract syntax trees executing over feature subsets, creating substantial kernel diversity beyond standard neural network primitives.

This preprocessing complexity—hundreds of operators with irregular memory access patterns, data-dependent control flow, and sparse computation characteristics— fundamentally determines serving paradigm in production systems, as discussed in Section~\ref{sec:intro}. The diversity and computational significance of preprocessing operators motivate comprehensive kernel coverage as a first-order architectural requirement rather than focusing solely on compute-intensive operations like GEMM.

\begin{figure}
    \centering

    \begin{subfigure}[b]{1\textwidth}
        \includegraphics[width=\textwidth]{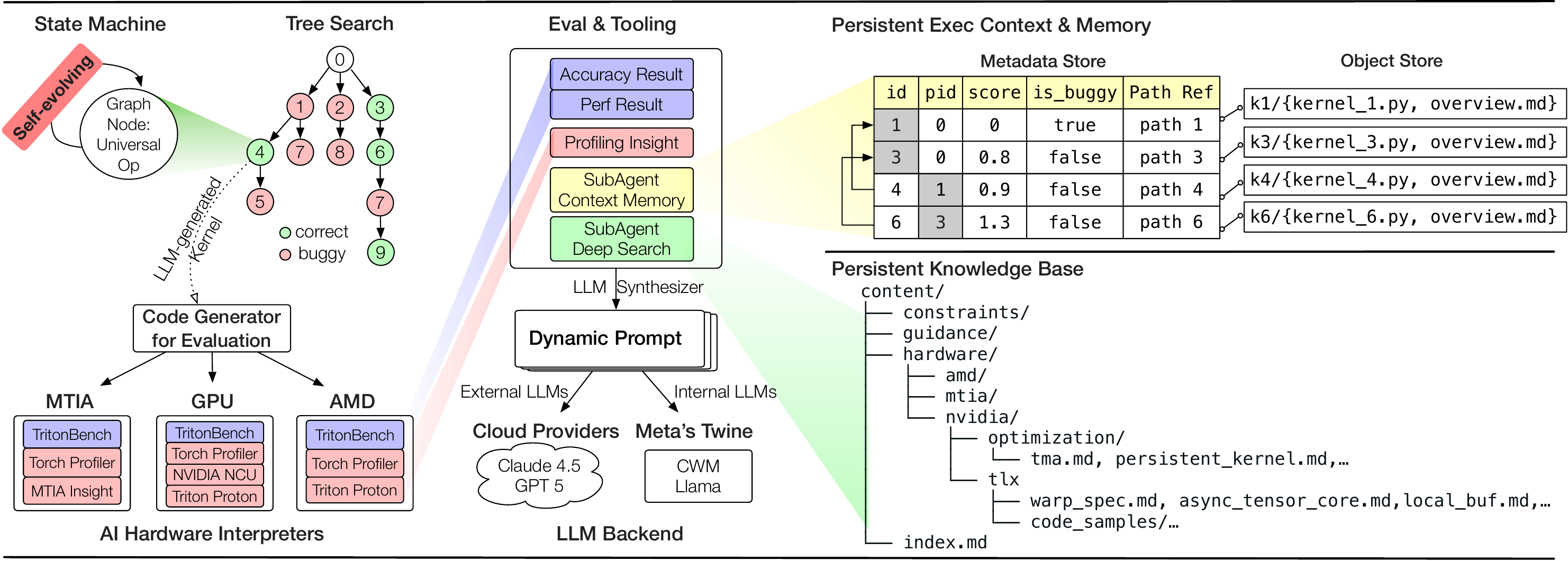}
    \end{subfigure}
    
    \vspace{0.5em}
    
    \begin{subfigure}[b]{1\textwidth}
        \includegraphics[width=\textwidth]{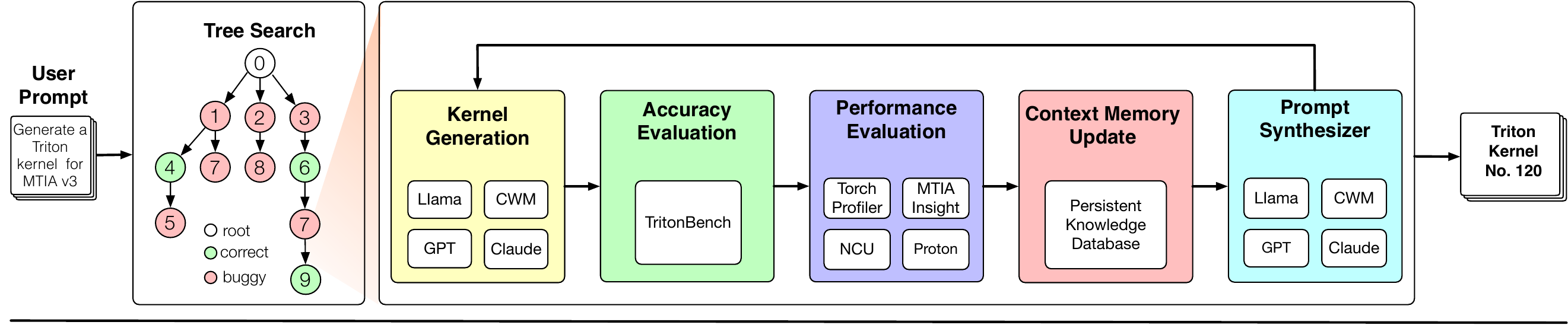}
    \end{subfigure}
    
    \caption{\name System Architecture (top) and Execution Workflow (down). \name employs a self-improving state machine with tree search to explore and validate kernel optimizations. The system integrates evaluation tooling (accuracy, performance, profiling), specialized sub-agents for context management and deep search, and AI hardware interpreters for MTIA, GPU, and AMD platforms. An LLM synthesizer generates dynamic prompts, which are then used by external (Claude 4.5, GPT-5) or internal (Meta's CWM) LLM backends to generate Triton kernel candidates. Persistent storage includes a metadata store tracking execution scores and parent-child relationships in the search tree (connected to the object store via path references), an object store for kernel files, and a knowledge base that serves as a retrieval system for hardware constraints and optimization guidance to support LLM context augmentation.}
    \label{fig:ak_system}
\end{figure}

\section{System Architecture}

Figure~\ref{fig:ak_system} illustrates the \name system architecture and its optimization workflow. The system centers around a self-improving state machine that drives exploration of the kernel optimization space through tree search strategies. Recent advancements~\cite{jiang2025aide} demonstrate state-of-the-art performance by conceptualizing iterative experimentation as tree search over potential solutions. Building on this insight, \name employs multiple search strategies including greedy search for rapid initial solutions, Monte Carlo Tree Search (MCTS)~\cite{kocsis2006bandit} for balanced exploration-exploitation, and evolutionary algorithms~\cite{de2017evolutionary} for population-based optimization. Each iteration begins with the LLM synthesizer generating dynamic prompts augmented with context from specialized sub-agents. The context memory sub-agent maintains relevant historical information from previous iterations and analyzes the current state of the search tree to guide exploration strategy, while the deep search sub-agent further enrich these prompts through retrieval from the persistent knowledge base, which organizes hardware-specific constraints for different accelerators (AMD, MTIA, NVIDIA), optimization guidelines, and code samples in a hierarchical file system structure. The augmented prompts are then processed by either external LLM backends (Claude 4.5, GPT-5) or internal models (Meta's CWM \cite{carbonneaux2025cwm} and Llama \cite{dubey2024llama} on Twine~\cite{tang2020twine}) to generate Triton kernel candidates.

Generated kernels undergo comprehensive evaluation across multiple dimensions. TritonBench validates correctness by comparing generated Triton kernels against PyTorch reference implementations while measuring speedup over baselines. System-level profiling via Torch Profiler captures CPU/GPU time, kernel launch overhead, and function execution durations to identify host-device bottlenecks. However, modern AI accelerator optimization faces a fundamental challenge: performance signals are fragmented across abstraction layers—DSLs, compiler IR, CUDA/PTX/SASS, runtime APIs, and hardware counters—requiring manual correlation across siloed tools. Intra-kernel tracers such as Triton Proton~\cite{zhou2025proton} expose instruction-level behavior, kernel profilers report occupancy and memory throughput, while system profilers capture execution timelines and communication patterns. No single tool provides complete stack visibility. To address this fragmentation, Meta developed MPP (Multi-Pass Profiler), a federated tooling framework that unifies instrumentation, compiler transforms, profiling, and trace synthesis as composable job graph tasks. Integrated with existing ecosystems including MLIR~\cite{lattner2020mlir}, Proton~\cite{zhou2025proton, guan2025kperfir}, NCU~\cite{nv-ncu}, and NVBit~\cite{villa2019nvbit}, MPP enables \name to automate cross-stack experiments and access metrics previously gated by tool fragmentation. Hardware-specific profilers (MTIA Insight and Triton Proton) provide platform-specific insights across MTIA, GPU, and AMD interpreters. The evaluation framework in \name produces accuracy validation, performance metrics, and profiling insights that collectively determine each search tree node's status (correct or buggy) and assign optimization insights to guide automatic exploration strategy.

The persistent storage layer enables continuous learning and knowledge accumulation across optimization runs. The metadata store maintains a comprehensive execution context for each kernel candidate, including unique identifiers, parent-child relationships that encode the search tree structure, quality scores from evaluation results, and boolean flags (\texttt{is\_buggy}) indicating whether the kernel has execution errors or accuracy failures. Path references in the metadata store link to the object store, which maintains the actual kernel implementations organized by unique identifiers along with associated Triton kernel files and overview documentation (\texttt{overview.md}) containing profiling results analysis and optimization recommendations. This separation of metadata and content allows efficient querying and comparison of kernel variants while preserving complete implementation history. The architecture's modularity supports flexible deployment configurations with different LLM backends and hardware targets while maintaining a consistent optimization methodology. By systematically exploring the solution space and learning from both successful optimizations and failed attempts, \name progressively generates higher-quality kernel implementations that approach theoretical performance limits for given hardware constraints.

The following sections detail the design and implementation of key system components.

\subsection{Tree Search and State Machine}

Given a kernel generation problem, \name maintains a search graph $G_t = (V_t, E_t)$ that evolves over multiple iterations $t = 0, 1, \ldots$, where each node $v \in V_t \subseteq \mathcal{S}$ represents a kernel implementation artifact belonging to the set of all possible artifacts $\mathcal{S}$, and each directed edge $(v_i, v_j) \in E_t$ represents a transformation from kernel $v_i$ to $v_j$. The root node $v_0$ represents the initial specification or baseline implementation. At each iteration, the agent executes three fundamental operations: (1) \textbf{selects} promising nodes via a selection policy $\pi_{\text{sel}}$, (2) \textbf{applies} a transformation operator to generate new kernel candidates, and (3) \textbf{scores} the resulting solutions via a fitness function $\mathcal{F}$ that evaluates correctness and performance.

Recent research demonstrates that LLMs alone are insufficient for effectively solving open-ended optimization tasks~\cite{nathani2025mlgym}. Performance significantly improves when augmented with external tools~\cite{qin2024tool,schick2023toolformer}, execution feedback~\cite{gehring2024rlef}, and solutions addressing context limitations. Developing high-performance kernels requires iterative experimentation where insights from prior attempts inform subsequent refinements. Building on recent advancements~\cite{jiang2025aide,toledo2025ai}, which achieve state-of-the-art results by conceptualizing iterative experimentation as tree search over potential solutions, we formalize kernel optimization as a graph-based search algorithm that systematically explores the solution space.

\textbf{Graph-Based Search Framework.} We model \name as a graph-based search algorithm specified by the tuple $(\mathcal{F}, \pi_{\text{sel}}, \mathcal{O}, \tau)$, where:

\begin{itemize}[leftmargin=12pt]
\item \textbf{Fitness Function} $\mathcal{F}: \mathcal{S} \rightarrow \mathbb{R}_{\geq 0}$ estimates the quality of a kernel implementation node $v \in V_t$ through the speedup achieved by the generated Triton kernel relative to the PyTorch compiled reference code~\cite{ansel2024pytorch}. Specifically, for a generated Triton kernel with execution time $t_{\text{triton}}$ and PyTorch compiled baseline with execution time $t_{\text{pytorch}}$, the fitness is computed as $\mathcal{F}(v) = \frac{t_{\text{pytorch}}}{t_{\text{triton}}}$. Kernels that fail correctness validation (numerical accuracy checks against reference code) or encounter compilation/runtime errors are assigned $\mathcal{F}(v) = 0$. This fitness measure directly captures the performance optimization objective while ensuring correctness as a hard constraint.

\item \textbf{Selection Policy} $\pi_{\text{sel}}: 2^{V_t} \rightarrow 2^{V_t}$ chooses a subset of nodes $U_t \subseteq V_t$ for expansion, guided by a heuristic function $h: V_t \rightarrow \mathbb{R}$ that assigns scalar estimates to each node. Different instantiations support various search strategies: greedy search selects the single highest-scoring node, Monte Carlo Tree Search (MCTS) balances exploration-exploitation via upper confidence bounds for trees (UCT), and evolutionary algorithms maintain populations of diverse candidates for crossover and mutation operations.

\item \textbf{Universal Operator} $\mathcal{O}: \mathcal{S} \times \mathcal{C} \rightarrow \mathcal{S}$ is a single transformation function that generates new kernel candidates from existing implementations, where $\mathcal{C}$ represents the contextual information (profiling results, error messages, hardware constraints, historical optimizations) that guides the transformation. Unlike traditional approaches that employ multiple specialized operators (\texttt{Draft}, \texttt{Debug}, \texttt{Improve}) with fixed prompting strategies, \name's universal operator dynamically adapts its behavior based on runtime context through retrieval-augmented prompt synthesis.

\item \textbf{Termination Rule} $\tau$ halts search when computational budgets (wall-clock time or maximum number of artifacts) are exhausted, progress stalls, or fitness thresholds are achieved.
\end{itemize}

\textbf{The Operator Bottleneck and Universal Operator Design.} Prior research indicates that performance bottlenecks in LLM-based code generation stem primarily from operator design rather than search algorithms~\cite{toledo2025ai}. Traditional multi-operator frameworks face a fundamental limitation: each operator (e.g., \texttt{Debug}, \texttt{Improve}) is associated with a static prompt template that cannot adapt to runtime execution context. For instance, \texttt{Debug} operators employ fixed error-focused prompts regardless of whether the underlying issue stems from algorithmic errors, memory access patterns, or hardware-specific constraints, while \texttt{Improve} operators use performance-focused prompts that remain unchanged despite varying bottleneck characteristics (compute-bound vs. memory-bound vs. synchronization overhead). This static prompting strategy imposes cognitive constraints on the model's reasoning process, potentially misleading the model by framing optimization problems through predefined lenses that may not align with the actual runtime context. The operator-specific prompting creates artificial boundaries in the solution space, preventing the model from simultaneously reasoning about correctness, performance, and architectural trade-offs based on the specific execution characteristics observed at runtime.

\begin{figure}
    \centering
    \includegraphics[width=1\linewidth]{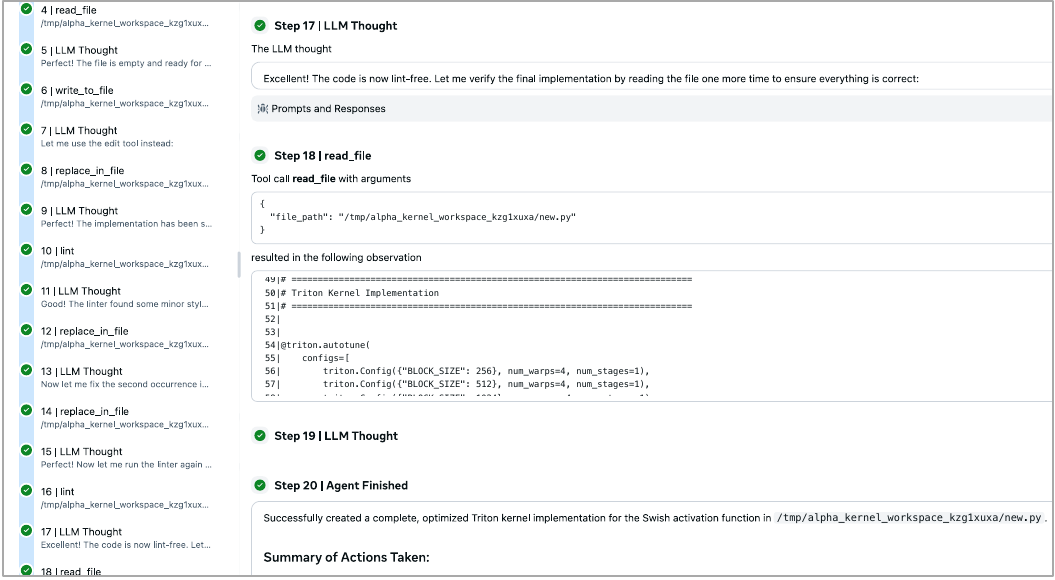}
    \caption{Universal operator's agentic kernel generation workflow for Swish activation. Agent iteratively reads, analyzes, modifies, and validates code through tool invocations (read\_file, write\_to\_file, replace\_in\_file, lint), guided by LLM reasoning at each step. The workflow completes after 20 steps, producing lint-free optimized Triton kernel with autotune configurations.}
    \label{fig:deep_search}
\end{figure}

To address this limitation, \name employs a \textbf{single universal operator} that dynamically adapts its behavior based on runtime context rather than predefined operator categories (Figure~\ref{fig:deep_search}). Instead of fixed prompt templates associated with specific operators, we introduce a retrieval-augmented dynamic prompting mechanism (detailed in Section~\ref{subsec:persistent-context}) that synthesizes contextually optimal prompts at each iteration. This design choice allows the LLM to reason holistically about kernel optimization without artificial constraints imposed by operator boundaries. By unifying operator functionality under a single context-aware interface, \name enables more flexible exploration strategies where each generation step can simultaneously address multiple optimization objectives---correcting numerical errors, improving memory access patterns, exploiting hardware-specific features, and refining algorithmic approaches.

\subsection{Agentic Retrieval \& Self-Managed Context}
\label{subsec:persistent-context}
\name employs a retrieval-augmented approach where contextual information is dynamically loaded into the LLM's context at runtime rather than maintaining complete historical knowledge in working memory. This design philosophy aligns with recent agentic coding systems such as Anthropic's Claude Code~\cite{anthropic2025contextengineering}, which demonstrates that complex analysis over large datasets can be performed through targeted queries and incremental loading rather than exhaustive context consumption. This approach reflects human cognitive patterns: rather than memorizing entire corpuses of information, humans introduce external systems—file hierarchies, databases, indexing structures—to retrieve relevant information on demand. For \name, this retrieval-based dynamic prompting eliminates cognitive constraints imposed by static prompt templates, enabling the LLM to reason freely about optimization strategies guided by actual runtime execution characteristics rather than predefined operator semantics.

\name operationalizes this architecture through a two-stage pipeline implemented by specialized sub-agents. The \textbf{context memory sub-agent} analyzes dynamic runtime artifacts—kernel implementations, profiling measurements, error diagnostics, correctness validation results—to diagnose performance bottlenecks and synthesize optimization directives. These analysis outputs parameterize the \textbf{deep search sub-agent}, which performs targeted retrieval from a persistent knowledge base containing hardware constraints, optimization patterns, and debugging methodologies. This staged design reflects a fundamental principle: effective knowledge retrieval requires runtime context to determine retrieval targets. The context memory sub-agent identifies which optimization challenges require attention; the deep search sub-agent retrieves domain knowledge addressing those specific challenges. By conditioning knowledge base retrieval on runtime analysis rather than static heuristics, \name ensures retrieved content directly targets observed bottlenecks, maintaining context window efficiency while avoiding irrelevant generic guidance.

\subsubsection{Deep Search Sub-Agent}
\label{subsubsec:knowledge-base}

The persistent knowledge base implements a hierarchical file system encoding domain expertise across hardware platforms, optimization strategies, debugging patterns, and language constraints. This organization exploits structural metadata—folder hierarchies, naming conventions, and file relationships—as retrieval signals that guide agentic search without explicit semantic annotation. A comprehensive index (\texttt{index.md}, Figure~\ref{fig:ak_system}) maintains the complete directory tree with inline module documentation, serving dual purposes: human-readable reference and machine-parseable navigation structure for automated retrieval.

\textbf{Hierarchical Taxonomy.} The knowledge base partitions content across three primary categories addressing distinct optimization concerns:
\begin{itemize}[leftmargin=12pt]
\item \textbf{Constraints} (\texttt{constraints/}): Enforces correctness requirements through anti-cheating rules (prohibiting cross-platform abstractions, external library dependencies), forbidden patterns (direct CUDA API usage, incomplete test coverage), and output format specifications. These constraints ensure generated kernels represent authentic Triton implementations rather than wrappers around pre-compiled libraries or high-level PyTorch operations that bypass kernel-level optimization.

\item \textbf{Guidance} (\texttt{guidance/}): Provides platform-agnostic optimization knowledge organized by concern: debugging methodologies (error interpretation, numerical stability), performance tuning (autotuning, block sizing, fusion strategies), and Triton language idioms (data types, indexing patterns, memory primitives). This cross-platform content transfers across hardware architectures, establishing foundational implementation patterns before platform-specific specialization.

\item \textbf{Hardware} (\texttt{hardware/}): Encodes platform-specific knowledge for NVIDIA GPUs, AMD GPUs, and MTIA accelerators. Each platform subtree maintains architectural documentation (compute hierarchies, memory subsystems, execution models), platform-specific debugging guidance (common pitfalls, precision issues), and advanced optimization techniques. NVIDIA content includes Hopper-generation features (Tensor Memory Accelerator, warp specialization via Triton Low-level Extensions (TLX)), persistent kernel patterns, and generation-specific optimizations. Hardware modules comprise 15-40 documents per platform—totaling $\geq$100 documents—reflecting the depth of architectural specialization required for production-grade kernel optimization.

\end{itemize}

\textbf{Index-Guided Retrieval.} The \texttt{index.md} file implements structured navigation enabling efficient content discovery. Upon receiving runtime feedback (profiling metrics, error diagnostics), the deep search sub-agent executes two-stage retrieval: (1) queries the index to identify relevant modules based on hardware platform, bottleneck type, and optimization phase; (2) fetches targeted content from identified modules. Memory bandwidth bottlenecks on NVIDIA H100 trigger index queries returning \texttt{hardware/nvidia/optimization/\{tma, shared\_memory, on\_device\_tma\}.md}. The hierarchical structure enables efficient pruning: top-level platform selection (\texttt{hardware/\{nvidia|amd|mtia\}/}) eliminates irrelevant architectures; folder hierarchies encode concern taxonomies (\texttt{arch/}, \texttt{debug/}, \texttt{optimization/}); file naming conventions signal specificity (\texttt{memory\_hierarchy.md} versus \texttt{on\_device\_tma.md}).

\textbf{Progressive Specialization.} Content organization supports iterative refinement throughout optimization trajectories. Initial generation retrieves broad guidance—Triton basics, general optimization principles, correctness requirements. Subsequent iterations navigate progressively specialized content guided by profiling feedback. A representative trajectory for compute-intensive GEMM on H100: (1) \texttt{hardware/nvidia/arch/tensor\_cores.md} establishing Tensor Core capabilities; (2) \texttt{hardware/nvidia/tlx/\{overview, warp\_specialization, async\_tensor\_core\_operations\}.md} introducing fine-grained control through producer-consumer warp patterns and asynchronous matrix operations; (3) \texttt{code\_samples/\{hopper-gemm-pipelined, hopper-gemm-ws\}.py} providing complete reference implementations. This progression from high-level tensor core usage to expert-level TLX pipelined and warp specialization exemplifies support for both novice implementations and production-grade optimizations approaching theoretical hardware limits.

\subsubsection{Context Memory Sub-Agent}
\label{subsubsec:context-memory}

The context memory sub-agent bridges the persistent knowledge base and runtime optimization state through a two-tier storage architecture separating metadata from content. As shown in Figure~\ref{fig:ak_system}, each explored search graph node persists to a metadata store (relational database) containing: unique identifiers (\texttt{id}), parent references (\texttt{pid}) encoding tree structure, fitness scores (\texttt{score}), correctness flags (\texttt{is\_buggy}), and path references (\texttt{path\_ref}) linking to an object store containing kernel implementations (\texttt{kernel\_n.py}), profiling results, and LLM-generated analysis reports (\texttt{overview.md}). This separation enables efficient metadata queries without loading large source files or profiling traces.

The relational storage architecture provides two critical capabilities beyond simple persistence. (1) \textbf{Distributed Concurrent Exploration.} It enables distributed concurrent exploration: multiple agents can simultaneously expand different nodes in the search graph, with the database providing transaction isolation and consistency guarantees. When \name scales to dozens or hundreds of concurrent agents exploring thousands of optimization steps, maintaining an in-memory graph representation becomes infeasible. The metadata store allows agents to operate independently, querying only relevant subgraphs on demand. 
(2) \textbf{Complex Contextual Queries.} It supports complex contextual queries through SQL operations that logically reconstruct graph views without materializing the entire structure in memory~\cite{liao2023filescale}. The relational schema enables efficient graph traversal via recursive Common Table Expressions (CTEs). These queries enable sophisticated context construction: analyzing sibling node outcomes, retrieving strategies from high-performing ancestors, identifying global best solutions for comparison.
(3) \textbf{Cross-Session Knowledge Reuse.} Persistence enables querying historically explored similar operators (matched by operator type, input shapes, hardware platform). When high-quality solutions exist, \name initializes search with these implementations rather than generating from scratch~\cite{liao2026experiencegraphsdatafoundation}. Consider a new GEMM variant for transformer attention on AMD MI350: metadata queries identify 15 historical GEMM kernels, three achieving $>1.5\times$ speedup through TLX warp specialization. \name retrieves the highest-performing implementation (score 1.5) with its optimization report documenting successful strategies (async tensor core operations, double-buffered shared memory), then focuses exploration on problem-specific adaptations (different dimensions, fusion opportunities) rather than rediscovering fundamental patterns. This approach provides: (reduced time-to-solution starting from proven implementations, inference cost reduction eliminating redundant token consumption, and environmental impact reduction through decreased computational overhead.
(4) \textbf{Fault Tolerance and Checkpointing.} Persistent storage enables fault-tolerant search with automatic checkpointing. When search processes crash or are interrupted, \name reconstructs the complete search state from metadata: loading explored nodes, their parent-child relationships, fitness scores, and associated artifacts. Tree search resumes from the last successful iteration rather than restarting from scratch. This proves critical for long-running optimization campaigns where generating production-grade kernels may require hundreds of iterations spanning hours: hardware failures, deployment updates, or resource preemption no longer discard accumulated exploration progress. The metadata store serves as continuous checkpoint, with each node insertion atomically persisting exploration state.

\textbf{Runtime Artifact Analysis.} At each search node, \name generates execution artifacts: kernel source, execution logs, correctness validation results, performance measurements (execution time, speedup), and profiling metrics (instruction latency, memory throughput, occupancy, synchronization overhead). The context memory sub-agent invokes the LLM to analyze these artifacts, producing structured reports diagnosing bottlenecks and recommending optimization strategies. Profiling revealing 30\% occupancy on H100 with high shared memory pressure triggers analysis identifying register spilling and bank conflicts as root causes, recommending register usage reduction through value recomputation and warp-level memory access optimizations.

\textbf{Dynamic Prompt Synthesis.} The context memory sub-agent composes prompts for the universal operator combining: (1) current kernel implementation and execution history, (2) LLM-generated analysis reports, (3) retrieved knowledge base content, (4) hardware-specific constraints. This implements self-managed context windows maintaining only task-relevant information within token budgets (64K-1M tokens depending on LLM backend). When multiple optimization opportunities exist, the sub-agent prioritizes based on profiling evidence, loading content for dominant bottlenecks while deferring secondary optimizations to subsequent iterations.

\textbf{Iterative Refinement.} The context memory maintains summaries of previous optimization attempts, enabling learning from successes and failures. If increasing block size fails to improve speedup or introduces correctness violations, subsequent iterations avoid that strategy and explore alternatives. This mirrors human debugging workflows where engineers track attempted fixes, analyze failures, and systematically explore solution spaces while avoiding dead ends. Combined with targeted knowledge retrieval, this enables efficient navigation of complex optimization landscapes compared to static prompts or naive trial-and-error approaches.

The persistent storage architecture scales efficiently: metadata queries execute in milliseconds even with millions of nodes, while object store access occurs selectively. Production deployments maintain search histories spanning months across hundreds of operator types and multiple platforms, creating continuously growing kernel expertise corpora benefiting all users while reducing aggregate inference costs.

\subsubsection{MTIA Knowledge Injection}
\label{subsubsec:mtia-knowledge}

MTIA presents unique challenges for LLM-based kernel generation: unlike widely-documented GPU architectures (NVIDIA CUDA, AMD ROCm), MTIA's proprietary architecture and programming model remain largely absent from public training corpora. Pretrained LLMs lack knowledge of MTIA-specific hardware features, extended Triton language constructs, and optimization patterns. \name addresses this knowledge gap through systematic injection of MTIA domain expertise into the persistent knowledge base, enabling the deep search sub-agent to retrieve MTIA-specific content that educates the LLM during kernel generation.

\begin{figure}[t]
    \centering
    \includegraphics[width=0.55\linewidth]{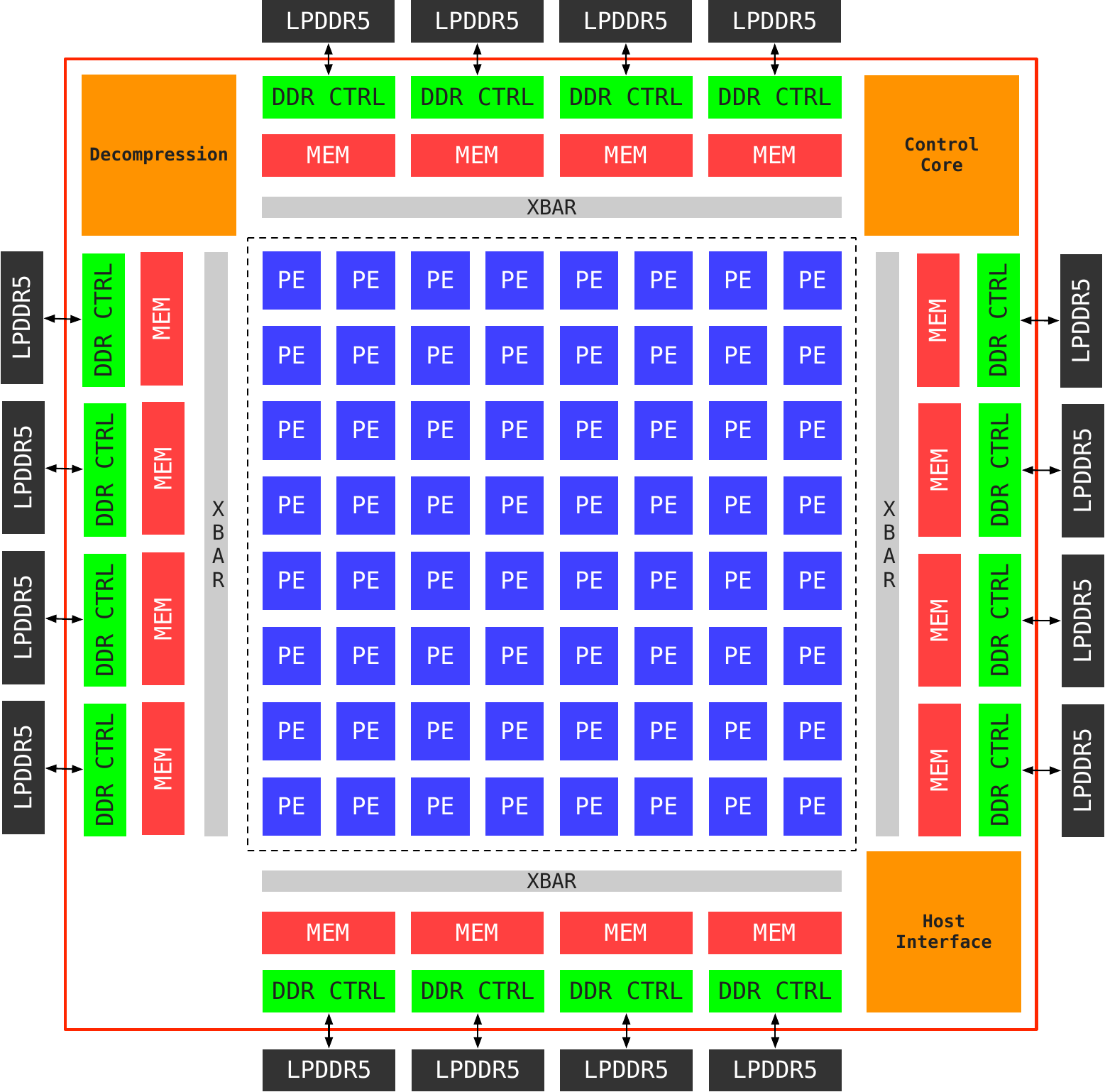}
    \caption{MTIA 2i architecture featuring an 8×8 processing element (PE) array interconnected via network-on-chip. Each PE contains dual RISC-V cores and specialized fixed-function units: Memory Layout Unit (MLU) for data transformation, Dot Product Engine (DPE) for matrix operations, Reduction Engine (RE) for aggregations, SIMD Engine (SE) for vector operations, and Command Processor (CP) for control flow (more details can be found in Section 3 of~\cite{coburn2025meta}).}
    \label{fig:mtia_pe}
\end{figure}

\textbf{MTIA Triton Extensions.} While Triton originated as a GPU-focused language, Triton-MTIA extends the base language to expose hardware-specific features fundamentally different from GPU programming models. The knowledge base documents three categories of extensions addressing distinct requirements:

\begin{itemize}[leftmargin=12pt]
\item \textbf{Hardware Feature Exposure}: As shown in Figure~\ref{fig:mtia_pe}, MTIA v2i architectures \cite{firoozshahian2023mtia,coburn2025meta} expose unique capabilities including Specialized Function Units (SFU) with lookup table (LUT) operations, inter-Processing Element (PE) communication primitives, and dual-core synchronization mechanisms. The knowledge base provides detailed documentation of libdevice APIs and hardware-specific compiler directives.

Libdevice APIs map Triton operations to hardware primitives: \texttt{tl.extra.libdevice.gelu(x)} compiles to SFU LUT queries rather than mathematical approximations, providing higher performance at potential accuracy cost. Documented operations include \texttt{exp}, \texttt{gelu}, \texttt{log}, \texttt{sigmoid}, \texttt{tanh}—each mapping to dedicated SFU instructions unavailable on GPU targets.

MTIA exposes hardware-specific performance tuning through compilation options during kernel invocation. Two critical options documented in the knowledge base enable pipeline parallelism optimization: (1) \texttt{cb\_multiplier} (integer) increases Circular Buffer allocation by specified factors, allowing multiple operations to execute concurrently by expanding on-chip memory capacity; (2) \texttt{use\_dual\_core} (boolean) instructs the compiler to distribute operations between core A and core B core—executing DMAs on core A while core B performs vector instructions—improving throughput through heterogeneous execution. These options can be explored statically via \texttt{@triton.autotune} decorators that evaluate configuration combinations (e.g., BLOCK\_SIZE $\in$ \{32, 1024\}, \texttt{cb\_multiplier} $\in$ \{1, 8\}), with autotuning keyed to input dimensions (\texttt{key=["N"]}) to rerun when problem characteristics change.

\item \textbf{Compute Helper Functions}: MTIA provides three categories of optimized compute helpers documented in the knowledge base: (a) \texttt{unary\_elemwise\_compute(op, x)} supporting 30+ operations including mathematical functions (\texttt{exp}, \texttt{log}, \texttt{sqrt}), activations (\texttt{relu}, \texttt{gelu}, \texttt{sigmoid}), and logical operations; (b) \texttt{binary\_elemwise\_compute(op, x, y)} for tensor-tensor operations including arithmetic, comparisons, and ML-specific functions (\texttt{gelu\_backward\_tanh}, \texttt{log\_sigmoid\_backward}); (c) \texttt{binary\_elemwise\_const\_compute(op, x, const)} for tensor-scalar operations. These helpers compile to optimized vector instructions, providing higher performance than manual Triton implementations expressing equivalent semantics.

\item \textbf{Custom Type System}: MTIA kernels operate on device-specific data structures including \texttt{TensorView} (tensor metadata with shape, stride, addressing information), \texttt{CoreID} (PE identification and chip topology), and \texttt{ExecutionGrid} (kernel launch configuration). The knowledge base documents type definitions via \texttt{@core.struct\_type} decorators and associated utility functions, enabling kernels to directly manipulate MTIA runtime structures rather than relying on compiler-managed abstractions.
\end{itemize}

\textbf{Advanced Synchronization and Communication Primitives.} MTIA's multi-PE architecture requires explicit control over inter-PE data movement and synchronization—capabilities absent from standard Triton. The knowledge base documents four critical extensions:
\begin{itemize}[leftmargin=12pt]
\item \textbf{Cross-PE Broadcasting} (\texttt{direction} attribute in \texttt{tl.load}): Enables streaming memory between neighboring PEs, allowing multiple PEs reading identical memory to receive data from immediate neighbors rather than independent memory accesses. The \texttt{direction} parameter (\texttt{"down"} or \texttt{"right"}) specifies propagation direction through the PE grid. Complementary \texttt{tl.consume()} operator reads and discards memory, maintaining functional correctness by ensuring all participating PEs execute identical load sequences.

\item \textbf{Cross-PE Reduction} (\texttt{direction} attribute in \texttt{tl.store}): Extends \texttt{tl.store} to send computed results directly to neighboring PEs, enabling collaborative computation patterns where multiple program instances cooperate to produce results. This mechanism supports efficient row-wise and column-wise reductions across PE arrays.

\item \textbf{Runtime Barriers} (\texttt{tl.pe\_runtime\_barrier()}): Introduces runtime synchronization across all PEs, enabling cross-PE reduction mechanisms and eliminating kernel splitting for explicit synchronization. This primitive maps directly to \texttt{libjit\_fba\_runtime\_barrier()}, requiring careful placement to execute exactly once per PE in the physical grid.

\item \textbf{Explicit Tensor Copies} (\texttt{tl.copy()}): Creates deep tensor copies facilitating producer-consumer synchronization between MTIA cores. The compiler automatically detects data race conditions and fails compilation if copy operations prove insufficient for correctness.
\end{itemize}

\textbf{Knowledge Injection Mechanism.} The MTIA hardware subtree (\texttt{hardware/mtia/}) in the persistent knowledge base contains multiple documents spanning architectural overviews, language extensions, optimization patterns, and complete code examples. When the deep search sub-agent receives queries targeting MTIA hardware, it retrieves relevant documentation—libdevice API references for activation functions, cross-PE communication patterns for multi-PE kernels, custom type definitions for runtime structure manipulation. This retrieved content enters the LLM's context window during prompt synthesis, effectively teaching the model MTIA-specific programming idioms absent from pretraining data. The context memory sub-agent further refines retrieval based on runtime feedback: compilation errors citing undefined MTIA primitives trigger retrieval of language extension documentation; profiling results indicating suboptimal SFU utilization trigger retrieval of libdevice mapping tables.

This knowledge injection strategy proves critical for MTIA kernel generation quality. Without MTIA-specific documentation in context, LLMs generate standard Triton code targeting GPU semantics, producing compilation failures or functionally incorrect kernels when executed on MTIA hardware. With systematic knowledge base retrieval, \name successfully generates production-grade MTIA kernels leveraging hardware-specific features—SFU operations, inter-PE communication, dual-core synchronization—approaching hand-optimized performance while maintaining the productivity benefits of automated generation. This approach generalizes to emerging accelerator architectures: as new hardware platforms enter production, corresponding documentation injected into the knowledge base enables immediate LLM-based kernel generation without model retraining.

\subsection{File and Code Search}
\label{subsec:code-search}

The deep search and context memory sub-agents operationalize their retrieval strategies through a unified code search interface implemented via Model Context Protocol (MCP) tools~\cite{anthropic2024mcp}. This interface leverages Meta's production code search infrastructure—distributed systems (BigGrep, Glean~\cite{meta2024glean}) operating over fbsource, Meta's monolithic repository analogous to Google's monorepo~\cite{potvin2016google}—capable of querying billions of lines of code providing millisecond-latency retrieval through pre-built indices and optimized search algorithms. The search infrastructure automatically detects and follows code references embedded within knowledge base documentation: when retrieved content contains fbcode file paths or repository links, the tools automatically trigger secondary searches retrieving the referenced implementation code. This automatic dereferencing mechanism bridges curated documentation and production codebases, allowing sub-agents to seamlessly traverse from abstract optimization guidelines to concrete implementation examples without manual intervention.

\textbf{Search Modalities.} The code search interface exposes three search modes via MCP tool invocations: \texttt{STRMATCH} executes exact string matching for locating specific identifiers (function names, API calls, hardware operations); \texttt{REGEX} performs pattern-based queries for matching structural patterns (class definitions, function signatures, optimization templates); \texttt{FILENAME} locates files matching path patterns (hardware-specific modules, configuration files, test implementations). These modes execute against both the knowledge base filesystem and distributed production code indices, providing millisecond-latency retrieval across documentation and implementation corpora.

\textbf{Result Integration.} Search results return as file paths with code snippets including contextual lines (default: 1 leading, 1 trailing). When automatic dereferencing occurs, both the knowledge base documentation and retrieved production code integrate into prompt synthesis, providing the LLM with complementary information: abstract optimization principles from curated documentation alongside concrete implementation patterns from production systems. This enables generation of kernels satisfying theoretical optimization criteria while conforming to organizational coding conventions and platform-specific idioms observed in deployed infrastructure.
\subsection{Evaluation and Tooling Framework}
\label{subsec:evaluation}

\subsubsection{Kernel Output Format}
\name generates kernel implementations conforming to a standardized dual-implementation interface enabling automated correctness validation and performance benchmarking. Each generated artifact comprises three components: a PyTorch baseline, an optimized Triton kernel, and input data generation, structured to support systematic evaluation and downstream compilation integration.

\textbf{Dual Implementation Interface.} Generated kernels provide two \texttt{nn.Module} implementations with identical signatures\footnote{For training operators requiring gradient computation, both implementations must additionally provide \texttt{backward()} methods with matching input/output signatures for gradient propagation.}. 

\begin{pythoncode}
import torch
import triton
import triton.language as tl
# from triton_mtia.python.mtia.eager import mtia_triton_launcher
# mtia_triton_launcher.init()  # REQUIRED for MTIA execution

class PytorchModel(nn.Module):
    def forward(self, *args) -> torch.Tensor:
        return pytorch_impl(*args) # Baseline: standard PyTorch ops (torch.matmul, torch.sum)

@triton.jit
def optimized_kernel(...):
    pass  # Low-level Triton implementation (tl.load, tl.store, tl.dot)

class TritonModel(nn.Module):
    def forward(self, *args) -> torch.Tensor:
        return kernel_wrapper(*args) # Launch Triton kernel with grid configuration

def get_inputs() -> List[Tuple[torch.Tensor, ...]]:
    # Generate test cases across multiple scales
    return [(torch.randn(N, N, device="cuda"), ...) 
            for N in [512, 1024, 2048, 4096]]
\end{pythoncode}





The PyTorch baseline prioritizes correctness over performance, providing ground truth for numerical validation. The Triton implementation consists of a \texttt{@triton.jit} kernel, a wrapper managing grid configuration and kernel launch, and an \texttt{nn.Module} encapsulation. Input generation produces test cases spanning diverse sizes, exposing performance characteristics under varying computational and memory pressure.

\textbf{Design Rationale.} The \texttt{nn.Module} structure enables integration with PyTorch's \texttt{torch.compile} infrastructure, supporting hybrid optimization strategies combining compiler-driven graph transformations with hand-optimized kernels. Standardized interfaces enable automated evaluation: correctness validation executes both implementations on identical inputs, comparing outputs via \texttt{torch.allclose()} with precision-dependent tolerances; performance profiling measures execution time and computes speedup ratios. This structured evaluation framework transforms kernel assessment from manual validation to systematic automation, enabling tree search to evaluate thousands of variants while maintaining correctness guarantees.

\subsubsection{AI Hardware Interpreters}

\label{subsubsec:interpreters}

Generated Triton kernels require execution on target hardware platforms for correctness validation and performance profiling. To enable automated evaluation across heterogeneous accelerators (NVIDIA GPUs, AMD GPUs, MTIA), \name establishes dedicated interpreter environments providing standardized execution contexts with complete software stacks, compilation toolchains, and runtime dependencies for each hardware target.

\textbf{Bento-Based Interpreters.} Meta's Bento platform—the standard Jupyter notebook environment—bundles external libraries (PyTorch, Triton, CUDA/ROCm) with internal frameworks including hardware-specific software stacks (MTIA runtime, vendor-specific compilers) and build systems. \name configures three hardware-specific interpreters: \texttt{meta\_kernel\_gpu\_interpreter} (NVIDIA), \texttt{meta\_kernel\_amd\_interpreter} (AMD), and \texttt{meta\_kernel\_mtia\_interpreter} (MTIA). Each interpreter encapsulates platform-specific compilation toolchains (TritonGPU/AMDGPU/MTIA-MLIR backends), runtime libraries (\texttt{mtia\_triton\_launcher} for MTIA, cuDNN for NVIDIA), and profiling tools (NCU, Proton).

\begin{figure}
    \centering
    \includegraphics[width=1\linewidth]{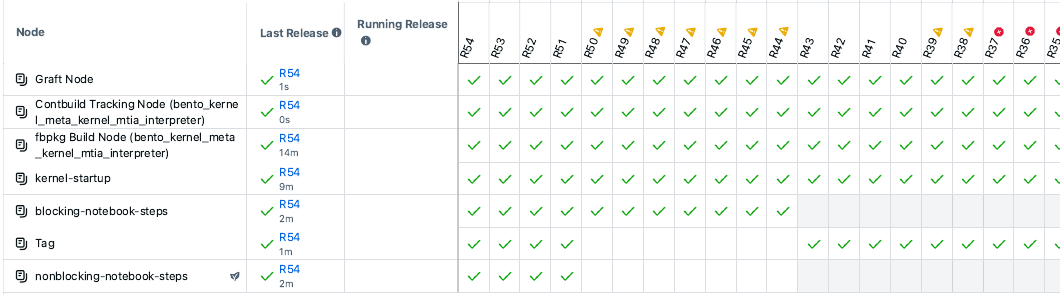}
\caption{Continuous deployment pipeline for \name interpreter environments via Meta's Conveyor system. The pipeline monitors dependency updates across Triton compiler backends, hardware runtime libraries, and build systems, automatically triggering daily rebuilds and deployments. Green checkmarks indicate successful releases, red crosses denote failed builds, and yellow warnings flag non-critical issues. The \texttt{bento\_kernel\_meta\_kernel\_mtia\_interpreter} node shows regular deployment cadence with occasional build failures, demonstrating automated recovery and continuous integration across multiple release candidates (R54, R53, etc.).}
\label{fig:conveyor_deployment}
\end{figure}

\textbf{Automated Deployment.} Interpreters integrate with Meta's Conveyor continuous deployment system~\cite{grubic2023conveyor}, which monitors dependency updates and automatically publishes new versions on regular schedules (Figure~\ref{fig:conveyor_deployment}). When underlying components update (Triton compiler, runtime libraries, build systems), Conveyor rebuilds and deploys interpreter packages with current dependencies. This eliminates manual environment maintenance: kernel artifacts submitted to interpreters execute directly against up-to-date software stacks. Interpreter isolation ensures reproducible profiling across tree search iterations.

This infrastructure abstracts hardware-specific complexities from kernel generation, enabling unified evaluation interfaces while maintaining platform-specific optimization capabilities.

\subsubsection{Evaluation Code Generation}
\label{subsubsec:eval-codegen}
\begin{figure}
    \centering
    \includegraphics[width=0.9\linewidth]{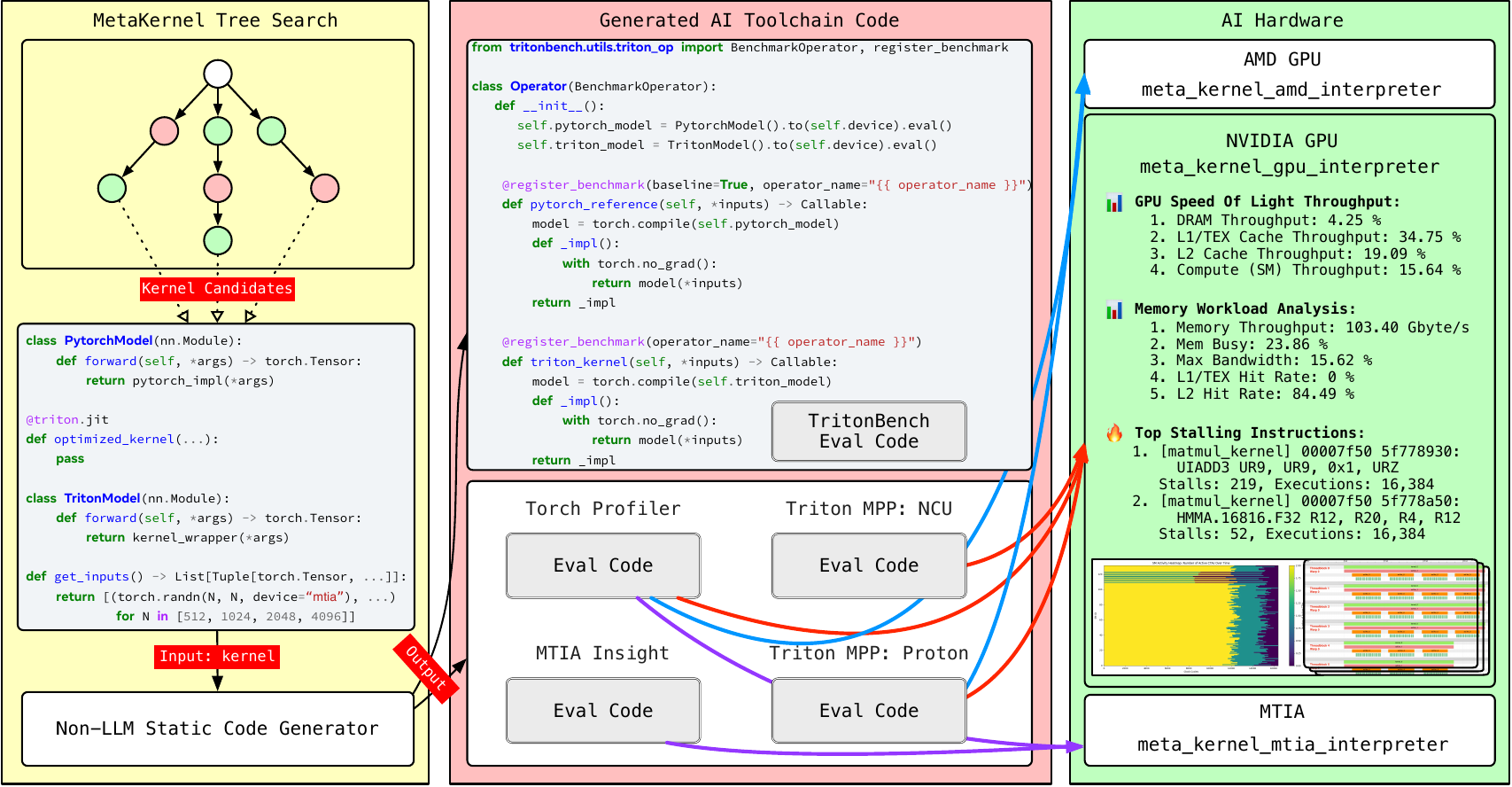}
\caption{End-to-end evaluation pipeline. Tree search generates kernel candidates with standardized dual implementations (PyTorch baseline, Triton optimized), executed on hardware interpreters (GPU, AMD, MTIA) collecting platform-specific profiling metrics via TritonBench, NCU, MPP, and MTIA Insight. Profiling feedback guides subsequent search iterations.}
\label{fig:evaluation_workflow}
\end{figure}

Kernel candidates generated by tree search conform to the standardized interface specification (\texttt{PytorchModel}, \texttt{TritonModel}, \texttt{get\_inputs()}), but require evaluation harness instrumentation for automated profiling. \name employs multiple profiling tools targeting different analysis objectives: TritonBench~\cite{meta2025tritonbench} validates correctness through numerical comparison against PyTorch baselines and measures speedup ratios across production input shapes; PyTorch Profiler~\cite{meta2021torchprofiler} captures system-level execution timelines including kernel launch overhead and host-device synchronization; NCU~\cite{nv-ncu} provides kernel-level hardware metrics including occupancy, memory throughput, and instruction mix; Proton~\cite{zhou2025proton} delivers intra-kernel instruction-level latency and pipeline behavior; MTIA Insight provides comprehensive MTIA-specific instrumentation: PE utilization, fixed-function accelerator metrics (DPE/SFU/MLU utilization and stall cycles), per-PE CPU runtime, cache analysis (CPU I/D-cache hit rates, branch prediction, LLC behavior), memory bandwidth (LLC/DRAM), and load-store throughput with per-PE read/write counters. \name employs a deterministic code generator transforming LLM-generated kernel implementations into platform-specific evaluation scripts invoking these profiling tool APIs (Figure~\ref{fig:evaluation_workflow}).

\textbf{Multi-Tool Evaluation Harness Generation.} The evaluation code generator accepts standardized kernel artifacts as input and produces executable Python scripts for each profiling tool. For TritonBench, the generator creates benchmark harnesses wrapping both \texttt{PytorchModel} and \texttt{TritonModel} within the \texttt{BenchmarkOperator} framework, configuring correctness validation (\texttt{baseline=True}) and speedup measurement. For Torch Profiler, generated scripts insert profiler contexts (\texttt{torch.profiler.profile()}) around kernel invocations capturing execution traces. For NCU and Proton (via Triton MPP), the generator synthesizes instrumentation invoking respective APIs with hardware-specific configurations. Each generated script imports models from standardized artifacts, instantiates profiler contexts, executes across test cases from \texttt{get\_inputs()}, and formats results as structured data consumable by the context memory sub-agent.

\textbf{Interpreter Execution Model.} Generated evaluation scripts leverage pre-deployed interpreter environments eliminating compilation overhead. Since hardware interpreters bundle complete toolchains (Triton compilers, profiling frameworks, runtime libraries) via Conveyor's continuous deployment, evaluation code executes immediately without dependency resolution or environment setup. This architectural separation—LLM-generated kernel logic versus deterministically-generated evaluation harness—provides critical benefits: (1) compilation occurs once during interpreter deployment rather than per-kernel evaluation (reducing latency from $\geq$10 minutes to seconds); (2) evaluation code remains consistent across kernel variants, ensuring reproducible profiling; (3) profiling tool APIs update independently through interpreter redeployment without modifying kernel generation prompts.

Figure~\ref{fig:evaluation_workflow} illustrates this workflow: tree search produces kernel candidates (left panel), the evaluation code generator transforms these into tool-specific harnesses invoking TritonBench, profilers, and hardware-specific instrumentation (center panel), and hardware interpreters execute generated evaluation code collecting platform-specific metrics (right panel) that feed back to guide subsequent tree search iterations.

\subsubsection{Unified Profiling: Triton MPP}

\name integrates Triton MPP (Multi-Pass Profiler) from Meta addressing fundamental GPU profiling challenges. Modern profiling workflows exhibit fragmentation: practitioners orchestrate IR-level tracers, assembly profilers (NCU), and binary instrumentation (NVBit), each with incompatible interfaces and vendor-specific assumptions. Profilers target human interpretation through textual reports and dashboards rather than structured data, forcing brittle text parsing unsuitable for automation.
MPP provides a compiler-centric abstraction unifying heterogeneous profiling tools. The framework composes analysis through job graphs: compiler transforms insert MLIR-level instrumentation, profiling passes collect metrics, trace synthesis produces structured output. This proves critical for modern Triton kernels employing TMA operations, warp-specialized pipelines, and overlapped data movement.

Traditional GPU profilers expose asynchronous behavior coarsely, failing to reveal instruction-level overlap between memory and computation. Direct wait insertion perturbs execution: initial synchronization disrupts overlap, cascading timing changes through subsequent operations. MPP addresses this through minimally-invasive profiling: capturing unmodified base traces, applying targeted probe passes isolating single instructions in specific iterations, guarding instrumentation to prevent interference, and fusing results for attribution. This profiles warp-group operations, async copies, and TMA transfers at TTGIR level with negligible perturbation.

MPP integration provides \name with structured, instruction-level performance data without vendor-specific parser implementations, transforming profiling from manual orchestration to programmatic composition.

\subsubsection{Agentic Debugging in JIT Flow}

Beyond profiling metrics, MTIA-Triton provides compiler introspection capabilities critical for kernel debugging and optimization. MTIA-Triton supports optional C++ code emission exposing the compiler's intermediate representation before final RISC-V binary generation:

\begin{pythoncode}
compiled_kernel = kernel[grid](x, *x.shape, BLOCK_SIZE=1024, cb_multiplier=8, emit_cxx=True)
cpp_source = compiled_kernel.asm["cpp"]
"""
/// This is a generated Triton C++ kernel.
extern "C" bool __mtia_is_core_b();
extern "C" uint64_t __mtia_scoped_block_print_info(uint64_t, uint64_t);
extern "C" v256_float16 __mtia_rvv_init256_fp16(float16_t);
extern "C" void __mtia_adjust_cb_read_pointer(uint32_t, uint32_t);
extern "C" void __mtia_adjust_cb_write_pointer(uint32_t, uint32_t);
...
"""
\end{pythoncode}

The emitted C++ code reveals low-level MTIA operations including RISC-V vector intrinsics, circular buffer pointer management, SFU initialization, and core affinity queries. 

\textbf{Interactive Debugging Workflow.} When generated kernels crash, fail correctness validation, or exhibit unexpected performance, the context memory sub-agent retrieves emitted C++ code alongside error diagnostics. The universal operator analyzes this representation—identifying incorrect buffer management, missing synchronization, or suboptimal SFU usage—and generates corrected C++ implementations. Modified C++ kernels can be tested immediately without full recompilation through MTIA's replay mechanism:

\begin{pythoncode}
from triton_mtia.python.mtia.eager.debug.replay_cpp import replay_cpp
# Rebuild modified C++ and launch with runtime arguments only
# (constexprs and compiler options already baked in)
replay_cpp(modified_cpp_source, compiled_kernel, 
           args=(1, 1, 1, x, *x.shape))  # PID grid + runtime args
\end{pythoncode}

This rapid iteration cycle—emit C++, modify, replay—eliminates full Triton recompilation overhead (constexpr resolution, MLIR lowering, backend code generation), enabling agents to validate hypothesized fixes within seconds rather than minutes. Compiler introspection transforms opaque execution failures into actionable optimization opportunities grounded in hardware-specific implementation details.

\subsubsection{Kernel Evaluation on FaaS}
\label{subsubsec:faas-evaluation}

Tree search execution decomposes each node expansion into two phases: kernel generation and kernel evaluation. Generation comprises prompt synthesis, knowledge base retrieval, and LLM invocation—CPU-bound operations requiring no accelerator access. Evaluation executes generated kernels on target hardware through TritonBench correctness validation, Torch Profiler timeline capture, and Triton MPP instruction-level analysis. This workload asymmetry—generation on CPU hosts versus evaluation on AI accelerators—motivates architectural disaggregation enabling independent, non-blocking evaluation execution on remote hardware.

\textbf{FaaS Integration for Remote Evaluation.} Kernel evaluation is an ideal FaaS workload: individual evaluation functions execute independently without inter-function dependencies or communication, unlike serverless databases requiring aggregation coordination~\cite{liao2023flock}. \name migrates kernel evaluation to Meta's FaaS (Function-as-a-Service) platform~\cite{sahraei2023xfaas}, which abstracts infrastructure complexity including service routing~\cite{saokar2023servicerouter}, Twine autoscaling~\cite{tang2020twine}, and lifecycle management. FaaS runtime auto-generates Thrift server interfaces for evaluation handlers, packaging them as fbpkg distributions with hardware interpreter dependencies. We extended FaaS's Tasklet resource model from CPU/RAM to include GPU resources, enabling hardware-specific evaluation functions targeting NVIDIA, AMD, and MTIA platforms. When tree search generates a kernel candidate, it asynchronously dispatches evaluation requests to FaaS endpoints corresponding to target hardware. Remote workers load pre-deployed interpreter environments, execute evaluation harnesses, and return structured results (correctness status, speedup ratios, profiling metrics) consumed by the context memory sub-agent.

\textbf{Benefits and Scalability.} Disaggregation addresses resource contention: a single host may run hundreds of generation agents but possess limited accelerators (8 GPUs or 24 MTIA devices per host). Without separation, agents serialize through available hardware—each occupying a device for 8-12 minutes (mostly idle during generation) while other agents queue. FaaS-based evaluation provides: (1) \textit{resource decoupling}—generation agents execute CPU-bound work locally while dispatching evaluation to remote accelerator pools, preventing device occupation during idle generation phases; (2) \textit{elastic capacity}—evaluation distributes across FaaS worker pools with hundreds of GPUs/MTIA devices rather than serializing through local hardware. This architecture maximizes both CPU (generation) and accelerator (evaluation) utilization, eliminating the mismatch between abundant generation parallelism and scarce local hardware resources.


\section{OSS Operator Evaluation}
\label{subsec:oss-benchmarks}

Kernel coverage—the availability of optimized implementations for standard operators—is a fundamental prerequisite for deploying models on emerging AI accelerators. Before optimizing for performance, the system must first demonstrate the ability to generate correct kernels across the operator set. This section evaluates \name's end-to-end capability to generate, validate, and benchmark kernels across heterogeneous hardware.

We curate a test suite of 160 ATen operators covering basic computational patterns: element-wise arithmetic (\texttt{torch.add}, \texttt{torch.div}), transcendental functions (\texttt{torch.cos}, \texttt{torch.exp}), reductions (\texttt{torch.amax}, \texttt{torch.allclose}), and activation primitives (\texttt{torch.ops.aten.elu}). While these operators are relatively simple, they represent the foundational building blocks required for PyTorch model execution and serve as an end-to-end validation of \name's kernel generation correctness. For each operator, \name generates Triton kernel implementations targeting three platforms: NVIDIA H100, AMD MI350, and MTIA v3. 
Generated kernels are validated against PyTorch reference implementations compiled with \texttt{torch.compile}. Numerical equivalence is verified using \texttt{torch.allclose} in TritonBench with precision-appropriate tolerances.

\name achieves 100\% correctness across all 480 operator-platform configurations (160 operators $\times$ 3 platforms). We further validate on KernelBench~\cite{ouyang2025kernelbench}, achieving 100\% pass rate across all three levels: Level 1 (single operators), Level 2 (fused operator patterns), and Level 3 (full model blocks).  While KernelBench originally targets CUDA kernel generation, these results demonstrate that \name reliably produces numerically correct Triton kernels across diverse architectures and operator complexities—from individual primitives to end-to-end model components—establishing the foundation for addressing the kernel coverage challenge on emerging hardware.

\begin{figure*}
    \centering
    \begin{subfigure}[b]{0.335\textwidth}
        \includegraphics[width=\textwidth]{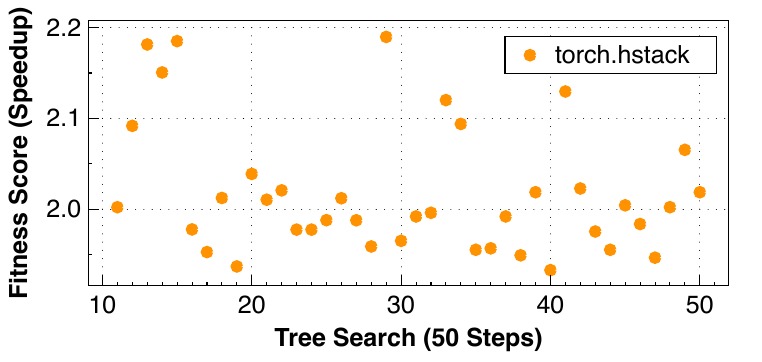}
    \end{subfigure}
    \hspace{-0.8em}
    \begin{subfigure}[b]{0.335\textwidth}
        \includegraphics[width=\textwidth]{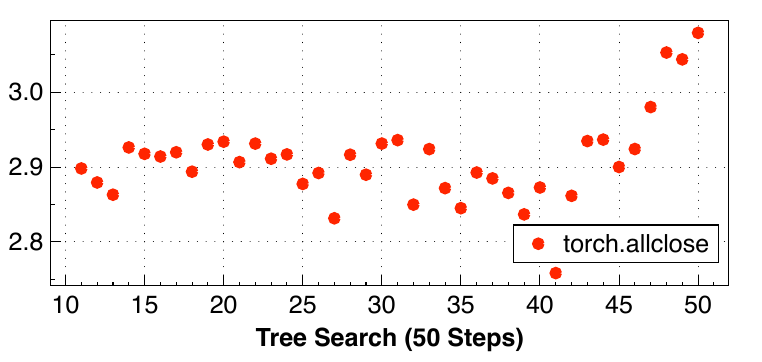}
    \end{subfigure}
    \hspace{-0.8em}
    \begin{subfigure}[b]{0.335\textwidth}
        \includegraphics[width=\textwidth]{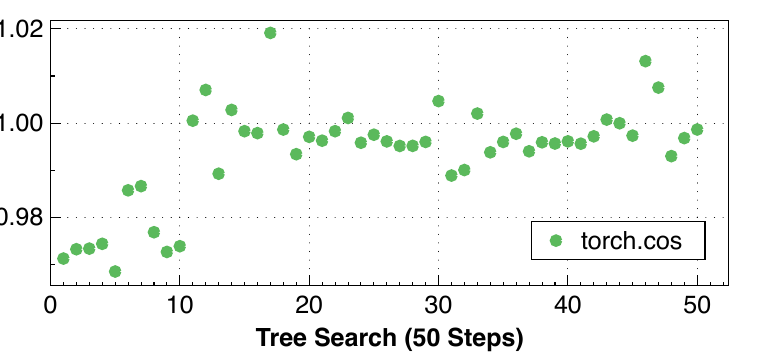}
    \end{subfigure}
    
    \vspace{0.5em}
    
    \begin{subfigure}[b]{0.335\textwidth}
        \includegraphics[width=\textwidth]{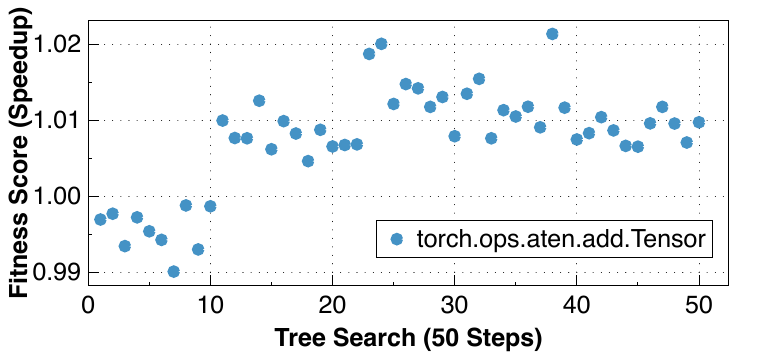}
    \end{subfigure}
    \hspace{-0.8em}
    \begin{subfigure}[b]{0.335\textwidth}
        \includegraphics[width=\textwidth]{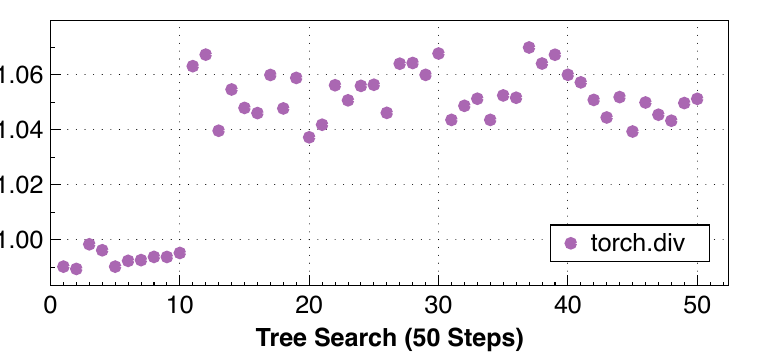}
    \end{subfigure}
    \hspace{-0.8em}
    \begin{subfigure}[b]{0.335\textwidth}
        \includegraphics[width=\textwidth]{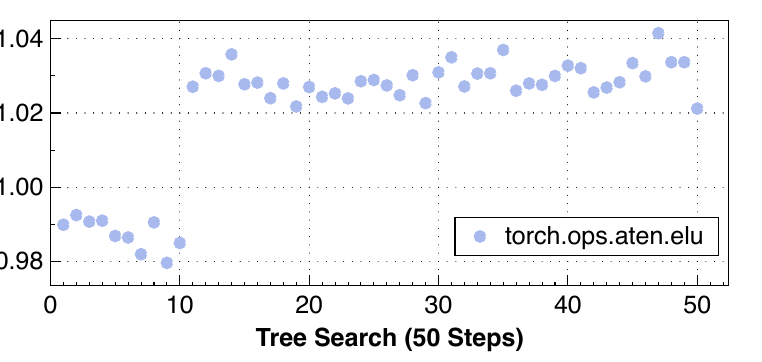}
    \end{subfigure}

    \caption{Fitness score trajectories during \name's tree search optimization for 6 representative ATen operators. The x-axis denotes search steps (50 total), and the y-axis shows the fitness score defined as the speedup ratio of the generated Triton kernel over the PyTorch baseline. The first 10 steps correspond to the draft phase (repeated sampling without memory context), while subsequent steps represent tree expansion with execution feedback.}
    \label{fig:aten-search}
\end{figure*}

Figure~\ref{fig:aten-search} shows optimization trajectories for six operators. The fitness score is defined as the speedup of the generated Triton kernel over the PyTorch reference.
The search operates in two phases. In the draft phase (steps 0--10), \name generates candidate kernels through independent sampling without feedback. In the tree expansion phase (steps 10--50), each node incorporates execution feedback—profiling data, compilation status, and correctness results—from its ancestors, enabling iterative refinement.
The trajectories exhibit operator-dependent behavior. \texttt{torch.cos} improves from 2.8× to 3.05× during tree expansion, indicating that feedback-guided search discovers superior implementations. \texttt{torch.ops.aten.add.Tensor} shows early-stage improvement (0.64× to 0.70×), demonstrating that iterative refinement benefits even initially suboptimal kernels. \texttt{torch.amax} and \texttt{torch.div} remain near 1.0× throughout, suggesting limited optimization headroom for these operators.
Four of six operators achieve fitness scores exceeding 1.0×, confirming that \name-generated kernels can outperform compiler-generated baselines.

These basic ATen operators serve primarily to validate \name's end-to-end correctness rather than to demonstrate optimization potential. As fundamental primitives, they offer limited headroom for improvement. In the following section, we evaluate \name on high-level ads operators, which compose multiple ATen primitives with ads-specific logic, exhibiting unique fusion opportunities and memory access patterns that yield substantially larger optimization potential and direct business impact.

\section{Monetization Case Study}
\label{sec:eval}

Figure~\ref{fig:kernel-evolve-speedups} presents \name's performance across production workloads, achieving 1.2-17× speedups over PyTorch baselines. We present detailed analysis of representative kernels: 1D convolution in convolutional transformers, operator fusion in WuKong's Optimized FM and InterFormer's PFFN, and data preprocessing operators (MapId, MBDT, Batch Event Truncate) across heterogeneous accelerators. 

Beyond these detailed case studies, several kernels achieve substantial speedups through more straightforward optimizations: expanded autotuning search spaces exploring block sizes and pipeline configurations, platform-specific compilation flags (MTIA's \texttt{cb\_multiplier}), and memory access improvements (cache modifiers, alignment hints). While individually less sophisticated, these optimizations demonstrate \name's ability to systematically explore configuration spaces that manual development often overlooks.

\begin{table}[t]
\centering
\small
\begin{tabular}{llccccc}
\toprule
\textbf{Precision} & \textbf{Tensor Shape} & \textbf{torch.conv1d} & \textbf{torch.conv2d} & \textbf{Triton} & \textbf{Speedup} & \textbf{Speedup} \\
& $(B \times C_{in} \times C_{out} \times L)$ & (ms) & (ms) & (ms) & \textbf{vs. conv1d} & \textbf{vs. conv2d} \\
\midrule
\multirow{9}{*}{FP16} 
& $64 \times 96 \times 96 \times 200$ & 0.03050 & 0.02019 & 0.01597 & \cellcolor{speedup}1.91× & \cellcolor{speedup}1.26× \\
& $128 \times 96 \times 96 \times 200$ & 0.03840 & 0.02490 & 0.01830 & \cellcolor{speedup}2.10× & \cellcolor{speedup}1.36× \\
& $256 \times 96 \times 96 \times 200$ & 0.05318 & 0.03347 & 0.02842 & \cellcolor{speedup}1.87× & \cellcolor{speedup}1.18× \\
& $512 \times 96 \times 96 \times 200$ & 0.08646 & 0.06006 & 0.04982 & \cellcolor{speedup}1.74× & \cellcolor{speedup}1.21× \\
& $1024 \times 96 \times 96 \times 200$ & 0.17226 & 0.11299 & 0.08406 & \cellcolor{speedup}2.05× & \cellcolor{speedup}1.34× \\
& \cellcolor{highlight}$2048 \times 96 \times 96 \times 200$ & \cellcolor{highlight}0.34243 & \cellcolor{highlight}0.24106 & \cellcolor{highlight}0.14864 & \cellcolor{speedup}\textbf{2.30×} & \cellcolor{speedup}\textbf{1.62×} \\
\cmidrule{2-7}
& \cellcolor{othershape}$32 \times 64 \times 64 \times 512^\dagger$ & \cellcolor{othershape}0.02768 & \cellcolor{othershape}0.01779 & \cellcolor{othershape}0.01264 & \cellcolor{speedup}2.19× & \cellcolor{speedup}1.41× \\
& \cellcolor{othershape}$32 \times 256 \times 256 \times 1024^\dagger$ & \cellcolor{othershape}0.07933 & \cellcolor{othershape}0.05485 & \cellcolor{othershape}0.06029 & \cellcolor{speedup}1.32× & \cellcolor{slowdown}0.91× \\
& \cellcolor{othershape}$64 \times 768 \times 768 \times 1024^\dagger$ & \cellcolor{othershape}0.71549 & \cellcolor{othershape}0.55354 & \cellcolor{othershape}1.12784 & \cellcolor{slowdown}0.63× & \cellcolor{slowdown}0.49× \\
\midrule
\multirow{9}{*}{FP32} 
& $64 \times 96 \times 96 \times 200$ & 0.03501 & 0.02531 & 0.02186 & \cellcolor{speedup}1.60× & \cellcolor{speedup}1.16× \\
& $128 \times 96 \times 96 \times 200$ & 0.04630 & 0.03248 & 0.03510 & \cellcolor{speedup}1.32× & \cellcolor{slowdown}0.93× \\
& $256 \times 96 \times 96 \times 200$ & 0.07168 & 0.05712 & 0.05789 & \cellcolor{speedup}1.24× & \cellcolor{slowdown}0.99× \\
& $512 \times 96 \times 96 \times 200$ & 0.15030 & 0.11517 & 0.11219 & \cellcolor{speedup}1.34× & \cellcolor{speedup}1.03× \\
& $1024 \times 96 \times 96 \times 200$ & 0.32077 & 0.24269 & 0.19725 & \cellcolor{speedup}1.63× & \cellcolor{speedup}1.23× \\
& \cellcolor{highlight}$2048 \times 96 \times 96 \times 200$ & \cellcolor{highlight}0.61411 & \cellcolor{highlight}0.46384 & \cellcolor{highlight}0.35594 & \cellcolor{speedup}\textbf{1.73×} & \cellcolor{speedup}\textbf{1.30×} \\
\cmidrule{2-7}
& \cellcolor{othershape}$32 \times 64 \times 64 \times 512^\dagger$ & \cellcolor{othershape}0.02730 & \cellcolor{othershape}0.01978 & \cellcolor{othershape}0.01571 & \cellcolor{speedup}1.74× & \cellcolor{speedup}1.26× \\
& \cellcolor{othershape}$32 \times 256 \times 256 \times 1024^\dagger$ & \cellcolor{othershape}0.13469 & \cellcolor{othershape}0.10326 & \cellcolor{othershape}0.13501 & \cellcolor{slowdown}1.00× & \cellcolor{slowdown}0.77× \\
& \cellcolor{othershape}$64 \times 768 \times 768 \times 1024^\dagger$ & \cellcolor{othershape}1.26237 & \cellcolor{othershape}1.04234 & \cellcolor{othershape}2.64502 & \cellcolor{slowdown}0.48× & \cellcolor{slowdown}0.39× \\
\bottomrule
\multicolumn{7}{l}{\footnotesize \colorbox{highlight}{Yellow}: production configuration. \colorbox{othershape}{Purple}$^\dagger$: randomly selected shapes (not optimization target).} \\
\end{tabular}
\caption{Conv1d kernel performance: \name-generated Triton kernel vs. PyTorch \texttt{conv1d} and \texttt{conv2d} baselines. The kernel is optimized for production ads ranking shapes (highlighted in yellow), achieving strong speedups. Performance on other shapes (highlighted in purple) varies: similar shapes benefit from the optimization, while out-of-distribution shapes show degraded performance.}
\label{tab:conv1d}
\end{table}

\subsection{Convolutional Transformer}
\label{subsec:conv-transformer}

Inspired by CNN Inceptions~\cite{szegedy2015going} and convolution-augmented transformers~\cite{gulati2020conformer}, the Convolutional Transformer architecture combines convolutional and transformer components to capture both local and global patterns in user sequential events for large-scale recommendation systems. 

The core of this architecture is a stack of 1D convolutional layers, which serve as the receptive field. Through multi-scale sliding windows with varying kernel sizes and strides, these conv1d layers compress long event sequences into shorter segment representations, extracting hierarchical patterns at different granularities. Given that conv1d operations dominate the computational workload, kernel-level optimization is critical for deployment at scale. To address this, we employ \name to automatically generates and tunes high-performance kernels on H100 GPUs through iterative refinement.

\textbf{Baselines and Evaluation Setup.} We compare \name-generated kernels against two PyTorch baselines on production shapes. The first baseline uses \texttt{torch.nn.functional.conv1d} directly. The second—a common optimization technique—reshapes input to 2D with \texttt{channels\_last} memory format and invokes \texttt{torch.nn.functional.conv2d}, mapping to cuDNN's heavily optimized Tensor Core path for NHWC convolutions. Table~\ref{tab:conv1d} evaluates performance across batch sizes with FP16 (serving) and FP32 (training) precision, verified with atol=$10^{-4}$, rtol=$5 \times 10^{-4}$.

\textbf{Performance Results.} 
On production shape $(B \times C_{in} \times C_{out} \times L) = (2048, 96, 96, 200)$, \name achieves 2.30× speedup over conv1d and 1.62× over the optimized conv2d baseline in FP16, with consistent gains across batch sizes (1.74-2.30× vs. conv1d). For FP32 training workloads, speedups reach 1.73× over conv1d and 1.30× over conv2d. The generated kernel is deliberately specialized: on out-of-distribution shapes (e.g., $64 \times 768 \times 768 \times 1024$), it underperforms baselines (0.49-0.63×), confirming that optimization targets production distributions rather than arbitrary inputs.

\begin{figure}
    \centering
    \includegraphics[width=0.7\linewidth]{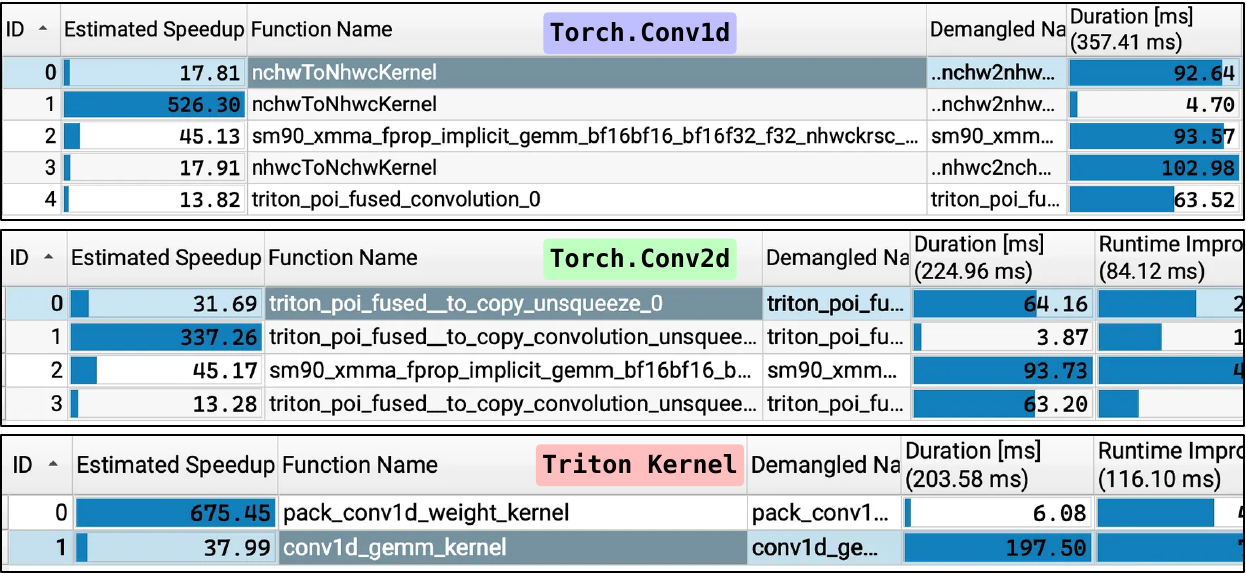}
    \caption{Profiling traces comparing conv1d implementations on production shape. PyTorch conv1d (top) launches five separate kernels including layout transformations and GEMM. PyTorch conv2d (middle) reduces to four kernels via optimized NHWC paths. \name (bottom) fuses operations into two kernels with cross-operation fusion. Note that durations shown in the profiling trace include profiling overhead and do not represent actual kernel latency.}

    \label{fig:conv1d-profiling}
\end{figure}

\textbf{Kernel Fusion as Primary Optimization.} Figure~\ref{fig:conv1d-profiling} and Table~\ref{tab:conv1d-kernels} reveal the source of these improvements through execution trace analysis. PyTorch conv1d launches five separate kernels: multiple layout transformations (\texttt{nchwToNhwcKernel}, \texttt{nhwcToNchwKernel}), Tensor Core GEMM (\texttt{sm90\_xmma\_fprop}), and a Triton-generated fusion kernel. Each launch incurs synchronization overhead and intermediate memory traffic. The conv2d baseline reduces this to four kernels via optimized NHWC paths but still requires separate operations for layout manipulation and computation.

\name fuses the entire operation into two kernels: weight preparation and the main convolution kernel. The Triton Conv1d kernel outperforms the PyTorch Conv2d workaround by eliminating memory layout conversions. The Conv2d approach launches four auxiliary kernels for unsqueeze, channels-last conversion, convolution, and squeeze operations---each requiring a full tensor pass and incurring significant global memory traffic. In contrast, Triton launches only two kernels: a lightweight weight-packing step and the fused convolution, operating directly on the native 1D layout. While cuDNN's implicit GEMM achieves high compute efficiency for the convolution itself, Triton delivers better end-to-end performance by avoiding redundant layout transformations and their associated memory overhead.

\begin{table}[t]
\footnotesize
\centering
\begin{tabular}{lll}
\toprule
\textbf{Implementation} & \textbf{Kernel} & \textbf{Operation} \\
\midrule
\texttt{torch.nn.Conv1d} 
& \texttt{nchwToNhwcKernel} & Convert input NCHW $\rightarrow$ NHWC \\
& \texttt{nchwToNhwcKernel} & Convert weights NCHW $\rightarrow$ NHWC \\
& \texttt{sm90\_xmma\_fprop\_implicit\_gemm} & Convolution (cuDNN implicit GEMM) \\
& \texttt{nhwcToNchwKernel} & Convert output NHWC $\rightarrow$ NCHW \\
& \texttt{triton\_poi\_fused\_convolution\_0} & Bias addition / post-processing \\
\midrule
\texttt{torch.nn.Conv2d} 
& \texttt{triton\_poi\_fused\_to\_copy\_unsqueeze\_0} & Layout conversion (unsqueeze + channels-last) \\
& \texttt{triton\_poi\_fused\_to\_copy\_convolution\_unsqueeze\_1} & Weight preparation (unsqueeze) \\
& \texttt{sm90\_xmma\_fprop\_implicit\_gemm} & Convolution (cuDNN implicit GEMM) \\
& \texttt{triton\_poi\_fused\_to\_copy\_convolution\_unsqueeze\_2} & Post-processing (squeeze 4D $\rightarrow$ 3D) \\
\midrule
\texttt{\name}
& \texttt{pack\_conv1d\_weight\_kernel} & Weight packing for GEMM-style access \\
\texttt{Triton Conv1d} & \texttt{conv1d\_gemm\_kernel} & Fused GEMM-style convolution \\
\bottomrule
\end{tabular}
\caption{Kernel breakdown comparison for conv1d implementations. PyTorch conv1d incurs significant layout conversion overhead. PyTorch conv2d reduces conversions through optimized NHWC paths. Triton conv1d eliminates redundant transformations through kernel fusion.}
\label{tab:conv1d-kernels}
\end{table}

\textbf{Complementary Optimizations.} Beyond fusion, the generated kernel employs architectural optimizations discovered through search. Expanded autotuning explores over 20 configurations across block sizes, warp counts, and pipeline stages, tailored to input dimensions and convolution parameters. A 3D grid launch parallelizes grouped convolution channels, eliminating inter-group dependencies. Double-buffered execution prefetches the next data blocks while computing current blocks, overlapping memory access with Tensor Core operations. Differentiated cache modifiers optimize memory hierarchy usage (\texttt{.ca} for streaming activations, \texttt{.cg} for reused weights).

\begin{figure*}
    \centering
    \includegraphics[width=1\linewidth]{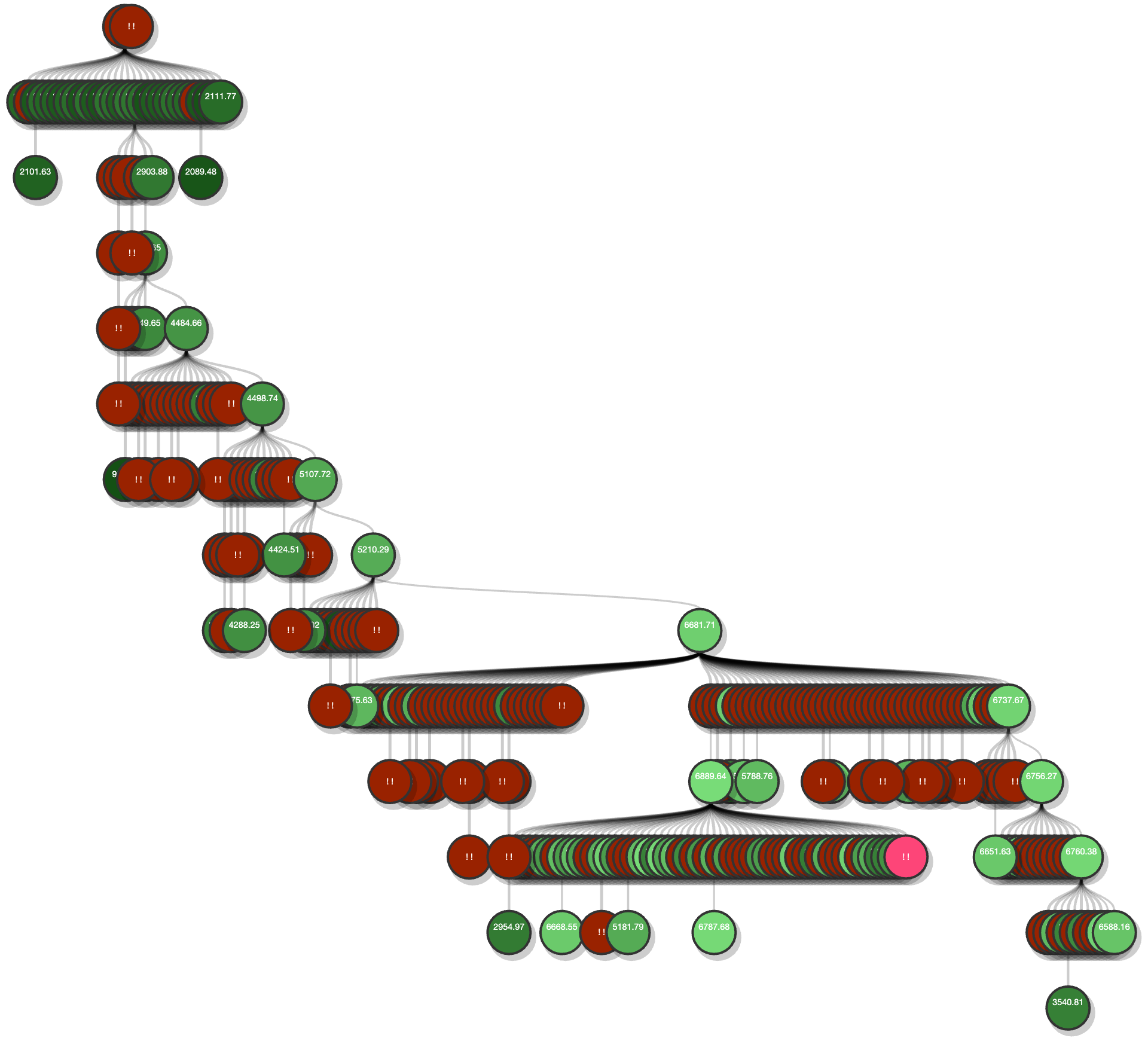}
\caption{Search tree visualization for \texttt{conv1d} kernel generation over 300 steps. Green: successful generation; Red: compilation/correctness failures.}
\label{fig:conv1d-tree}
\end{figure*}

\textbf{Search-Based Discovery.} Figure~\ref{fig:conv1d-tree} visualizes the optimization trajectory over 300 search steps, where the fitness score equals $1/\text{latency}$ (higher is better). Green nodes indicate successful generations; red nodes indicate compilation or correctness failures. Initial draft phases achieve fitness scores around 2000. As search progresses with accumulated execution feedback, scores improve systematically—reaching 4000, then 5000, and ultimately converging to 6889. This trajectory demonstrates that graph-based search with performance-guided selection discovers increasingly efficient implementations through inference-time scaling, automatically identifying the fusion strategies and tiling configurations that manual development would require weeks to explore.

Appendix~\ref{appendix:conv1d} provides source code for the PyTorch \texttt{conv1d} and \texttt{conv2d} baselines, the \name-generated Triton kernel, and TritonBench scripts for accuracy and speedup evaluation.

\subsection{Convolution on Heterogeneous Hardware}
\label{subsec:conv-hetero}

The diversity of hardware vendors and generations in large-scale model inference environments poses significant challenges for both enablement and optimization. \name's graph-based search and retrieval-augmented prompting are specifically designed to accommodate this heterogeneity, generating optimized kernels for AMD, NVIDIA, and MTIA accelerators from unified operator specifications. To validate cross-platform effectiveness, we evaluate the conv1d kernel from convolutional transformers (Section~\ref{subsec:conv-transformer}) across five hardware platforms spanning three vendors and multiple generations.

\begin{figure}
    \centering
    \includegraphics[width=1\linewidth]{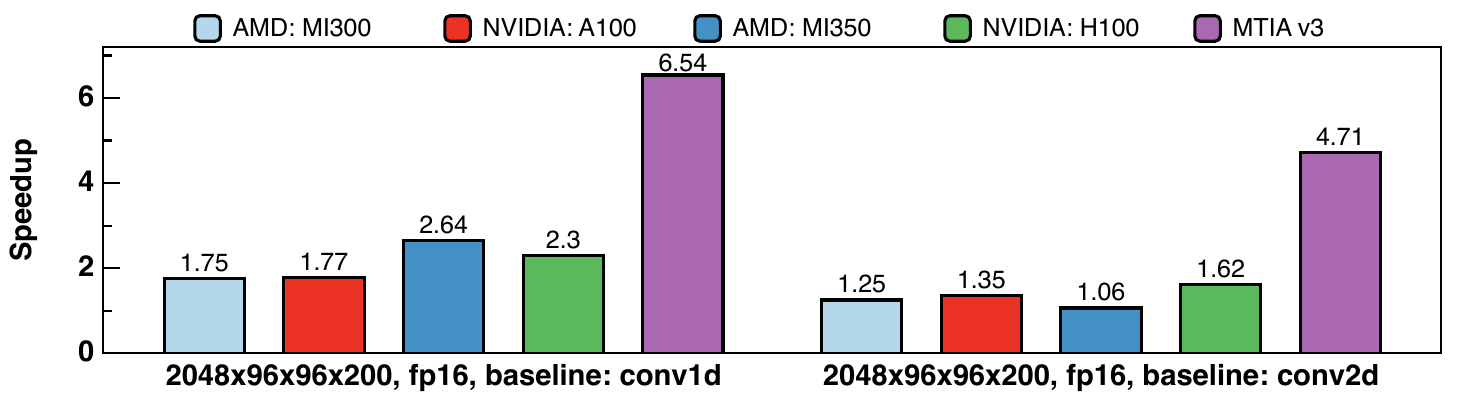}
    \caption{\name-generated kernels compared against PyTorch conv1d and optimized conv2d baselines, demonstrating up to 6.22× speedup across NVIDIA, AMD, and MTIA architectures.}
    \label{fig:conv1d-heterogeneous}
\end{figure}

\textbf{Cross-Platform Baseline Comparison.} Figure~\ref{fig:conv1d-heterogeneous} compares \name-generated kernels against two PyTorch baselines on production shape $(B \times C_{in} \times C_{out} \times L) = (2048, 96, 96, 200)$ with FP16 precision. The first baseline uses \texttt{torch.nn.functional.conv1d} directly. The second baseline employs a common optimization technique: reshape the input to 2D with \texttt{channels\_last} memory format and invokes \texttt{torch.nn.functional.conv2d}, which maps to cuDNN's heavily optimized Tensor Core path for NHWC convolutions. While mathematically equivalent to conv1d, this conv2d approach often provides superior performance on modern GPUs by exploiting vendor-optimized libraries.

\name-generated kernels achieve consistent speedups over the conv1d baseline across all platforms: 1.75× on AMD MI300, 2.30× on NVIDIA H100, 2.54× on AMD MI350, 1.77× on NVIDIA A100, and 6.54× on MTIA v3. Performance relative to the optimized conv2d baseline varies by platform: NVIDIA GPUs show modest improvements (1.62× on H100, 1.35× on A100), reflecting cuDNN's maturity on these architectures. AMD platforms demonstrate smaller gains (1.25× on MI300, 1.06× on MI350). MTIA v3 achieves the largest speedup at 4.71× over conv2d, demonstrating that \name's automated synthesis can effectively target custom accelerator architectures where vendor library coverage is less mature.

\textbf{Hardware-Specific Optimization Strategies.} The performance variations across platforms reflect fundamental architectural differences that \name's knowledge base encodes. On NVIDIA GPUs, generated kernels exploit Tensor Core operations through careful tile sizing and memory layout transformations. AMD platforms benefit from Infinity Cache-aware tiling that maximizes on-chip data reuse. MTIA kernels leverage specialized function units and inter-PE communication primitives absent from GPU programming models (Section~\ref{subsubsec:mtia-knowledge}). Critically, these hardware-specific optimizations emerge through systematic search guided by platform-specific documentation retrieved from the knowledge base—rather than manual per-platform tuning.

\textbf{Portability vs. Performance Trade-offs.} The results illustrate a fundamental challenge in heterogeneous deployment: optimizations targeting one platform may not transfer to others. \name addresses this challenge through shape-aware dispatch: for each platform, the system generates and validates platform-specific kernels during an offline optimization phase, then deploys the highest-performing variant while maintaining fallback paths to vendor libraries (conv1d/conv2d) when generated kernels underperform. This architecture ensures that automated synthesis delivers performance improvements where possible without risking regressions, enabling safe production deployment across diverse accelerator fleets.

\subsection{WuKong and InterFormer: Kernel Fusion}

\subsubsection{Optimized FM in WuKong}
\label{subsubsec:dcpp}

Optimized FM is a core computational primitive in Meta's Wukong recommendation model~\cite{zhang2024wukong} (Section 3.6). In factorization machine-based architectures, computing the pairwise dot product $XX^\top$ has $O(N^2D)$ complexity, prohibitive for real-world datasets with thousands of features. Wukong exploits the low-rank property of $XX^\top$ by introducing a learnable projection matrix $Y \in \mathbb{R}^{N \times K}$ where $K \ll N$, reducing output dimensionality from $N \times N$ to $N \times K$. Leveraging associativity, the computation reorders to:

\begin{equation}
\texttt{out} = X \cdot (X^\top Y)
\end{equation}

\begin{pythoncode}
def pytorch_ref_impl(x: torch.Tensor, y: torch.Tensor) -> torch.Tensor:
    """
    Reference PyTorch native implementation of Optimized FM kernel.
    """
    xty = torch.bmm(x.permute(0, 2, 1), y)  # (B, D, N) @ (B, N, K) = (B, D, K)
    return torch.bmm(x, xty)                 # (B, N, D) @ (B, D, K) = (B, N, K)
\end{pythoncode}

\noindent where $X \in \mathbb{R}^{B \times N \times D}$ and $Y \in \mathbb{R}^{B \times N \times K}$. Computing $X^\top Y$ first reduces complexity from $O(N^2D)$ to $O(NKD)$. This two-stage batched matrix multiplication presents a fusion opportunity: the intermediate result $X^\top Y \in \mathbb{R}^{B \times D \times K}$ can remain in registers or shared memory, eliminating global memory round-trips.

\textbf{Production Configuration.} We evaluate on production shapes extracted from a deployed Wukong variant: $(B, N, D, K) \in \{(1024, 24, 224, 2198), (1024, 40, 224, 448), (1024, 48, 224, 448)\}$. The PyTorch baseline with \texttt{torch.compile} generates two separate \texttt{extern\_kernels.bmm} calls, each independently loading inputs from HBM, performing computation, and writing results back—missing the fusion opportunity between operations.

\textbf{Kernel Optimization Strategy.} \name generates a fused Triton kernel exploiting two key optimizations:

\textit{(1) Operator Fusion Eliminating Intermediate Materialization.} The PyTorch baseline executes two independent matrix multiplications with an intermediate HBM round-trip. \name fuses both operations into a single kernel: inputs are loaded once, the intermediate result $X^\top Y$ remains in SRAM throughout computation, and only the final output writes to HBM. This reduces memory traffic by approximately 2×—eliminating one full read-write cycle of the intermediate tensor.

\textit{(2) Shape-Specific Tiling for SRAM Residency.} Rather than using PyTorch's standardized autotuning templates, \name generates custom tiling configurations tailored to production shapes. The kernel decomposes inputs into tiles sized to fit the complete computation chain (both multiplications) within SRAM capacity. For production configurations where $B=1024$, $D=224$ remain fixed, tile dimensions are optimized to maximize SRAM utilization while ensuring the fused operation executes entirely on-chip.

\begin{figure*}
    \centering
    \begin{subfigure}[b]{0.5\textwidth}
        \includegraphics[width=\textwidth]{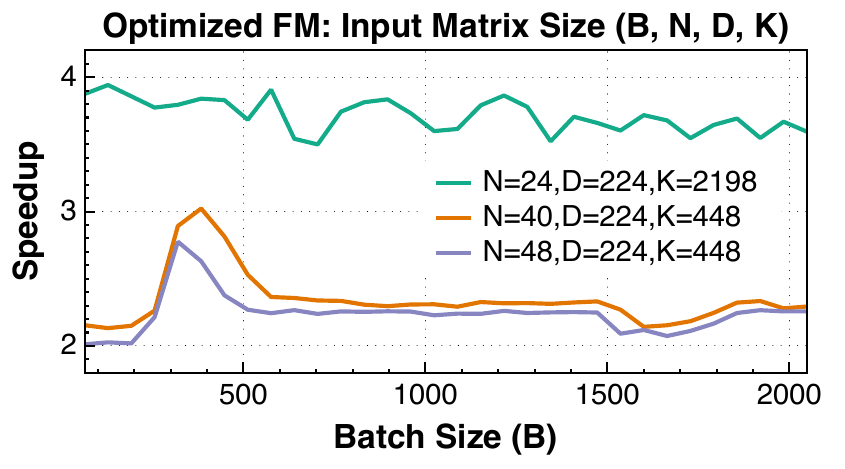}
    \end{subfigure}
    \hspace{-0.8em}
    \begin{subfigure}[b]{0.5\textwidth}
        \includegraphics[width=\textwidth]{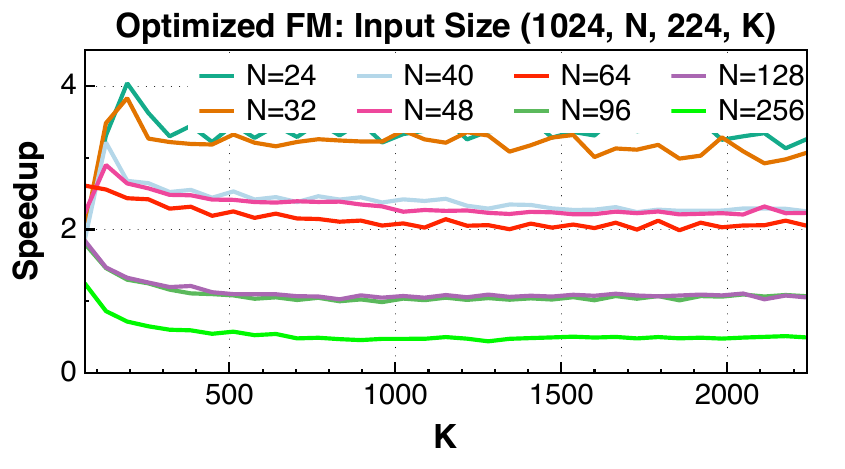}
    \end{subfigure}
    \caption{Optimized FM speedup on production shapes. Left: Performance across batch sizes for three representative $(N,D,K)$ configurations extracted from deployed Wukong models. Right: Speedup variation with output dimension $K$ at fixed batch size $B=1024$, showing strong performance for small-to-medium feature counts ($N \leq 64$) and degradation for larger $N$ where tiling overhead dominates.}
    \label{fig:dcpp-analysis-prod}
\end{figure*}

\textbf{Performance Analysis.} Figure~\ref{fig:dcpp-analysis-prod} analyzes speedup across production shape variations. The left panel shows speedup as a function of batch size $B$ for three representative configurations. The kernel achieves 3.6-3.9× speedup for small $N$ (24 features), maintaining stable performance as batch size scales from 128 to 2048. For medium $N$ (40, 48 features), speedup ranges from 2.1-3.0× at small batch sizes, gradually decreasing to 2.2-2.3× as $B$ increases—reflecting the trade-off between fusion benefits and tiling overhead.

The right panel examines speedup as a function of output dimension $K$ with fixed $B=1024$, $D=224$. Small $N$ values (24-32 features) maintain 3.0-3.5× speedup across the entire $K$ range (256-2304), demonstrating robust performance. Medium $N$ values (40-64 features) achieve 2.0-2.5× speedup, while larger $N$ (96-256 features) show diminishing returns, approaching 1× as $N$ increases. This degradation occurs because larger feature counts require more tiles to fit in SRAM—as the number of tiles grows, the overhead of tile management and accumulation eventually surpasses the benefits of on-chip computation, making direct HBM execution competitive.

\textbf{Deployment Strategy.} The generated kernel demonstrates consistent 2-4× speedups on production shapes where $N \leq 64$, covering the majority of deployed Wukong variants. For configurations with larger feature counts where tiling overhead dominates, the system falls back to PyTorch's unfused baseline. This shape-specific dispatch ensures performance gains on target workloads without risking regressions on out-of-distribution inputs. The results validate that \name's search-based optimization can discover fusion strategies and tiling configurations tailored to production distributions, achieving competitive performance with expert manual implementations while reducing development time from weeks to hours.

\subsubsection{PFFN in InterFormer}
\label{subsubsec:pffn}

Personalized FeedForward Network (PFFN) is a key component of the InterFormer architecture in Meta's ads ranking system~\cite{zeng2025interformer}. In recommendation models, user behavior sequences are inherently noisy—users browse items randomly, making pure sequential modeling ineffective. InterFormer addresses this by enabling bidirectional information flow between non-sequential features (e.g., user demographics) and sequential features (e.g., browsing history).

\textbf{Module Structure.} The PFFN module comprises five operations executed sequentially: (1) a feed-forward neural network (batched matrix multiplication with bias), (2) GELU activation, (3) root-mean-square normalization (RMSNorm), (4) another feed-forward layer, and (5) final RMSNorm. This operator chain processes tensors $X \in \mathbb{R}^{B \times N \times D}$ with weight matrices $W_1 \in \mathbb{R}^{B \times D \times K}$ and $W_2 \in \mathbb{R}^{B \times K \times D}$, where $B$ denotes batch size, $N$ sequence length, $D$ input dimension, and $K$ hidden dimension.

\textbf{Production Configuration.} We evaluate on shapes extracted from deployed InterFormer models: $(B, N, D, K) \in \{(1024, 200, 256, 160), (1024, 200, 192, 96), (1024, 400, 256, 160), (1024, 150, 96, 192)\}$. Production deployments exhibit consistent patterns with $B=1024$, $N \in [150, 400]$, $D \in [96, 256]$, and $K \in [96, 256]$. The PyTorch baseline with \texttt{torch.compile} generates two separate kernels: (1) \texttt{extern\_kernels.bmm} for matrix multiplication (single pass: load inputs, compute, write output), and (2) a two-pass fused Triton kernel \texttt{triton\_per\_fused\_rms\_norm\_add\_gelu} where the first pass loads data to perform bias addition and accumulate RMSNorm statistics, and the second pass reloads data to apply normalization. This results in three total memory round-trips: one for BMM and two for the fused operations. While PyTorch exploits fusion opportunities among element-wise operations, the multi-pass execution and kernel separation incur redundant memory traffic.

\textbf{Kernel Optimization Strategy.} \name generates two kernel variants targeting different operator chains: (1) fusing feed-forward network with RMSNorm, and (2) fusing feed-forward network, GELU, and RMSNorm. We select the highest-performing variant for production deployment, evaluating on FP16 precision matching production serving requirements. The generated kernel achieves performance improvements through two key optimizations:

\textit{(1) Shape-Specific Tiling for Target Distributions.} \name's search process incorporates production input shape ranges during kernel generation. The generated kernel employs customized tiling configurations that maximize SRAM utilization for target dimensions—in contrast to PyTorch's templated BMM kernel using generic tiling heuristics. For production shapes where $D \in [96, 256]$ and $K \in [96, 256]$, the specialized tiling ensures tiles remain SRAM-resident throughout computation, avoiding HBM fallback that occurs with one-size-fits-all tile sizes.

\textit{(2) Cross-Operation Tile Reuse.} 
\name generates a unified single-pass kernel that loads tiles once, performs the complete operator chain (matrix multiplication, bias addition, GELU, RMSNorm) while data resides in SRAM, and writes final results to HBM—requiring only one load and one write per tile.  
The PyTorch baseline executes PFFN through two separate kernels with three total passes: (1) the first kernel (\texttt{extern\_kernels.bmm}) loads inputs, performs matrix multiplication, and writes intermediate results; (2-3) the second kernel (\texttt{triton\_per\_fused\_rms\_norm\_add\_gelu}) executes in two passes—the first pass reloads data to perform bias addition and accumulate RMSNorm statistics, the second pass reloads data again to apply normalization.

\begin{figure*}
    \centering
    \begin{subfigure}[b]{0.5\textwidth}
        \includegraphics[width=\textwidth]{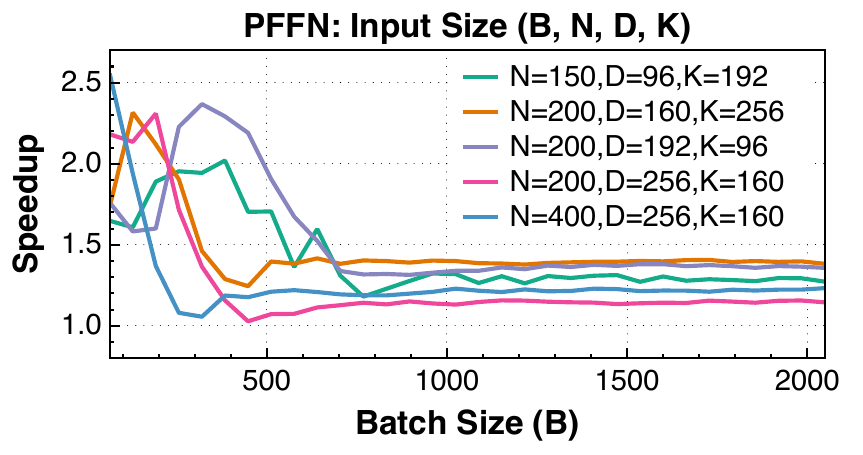}
    \end{subfigure}
    \hspace{-0.8em}
    \begin{subfigure}[b]{0.5\textwidth}
        \includegraphics[width=\textwidth]{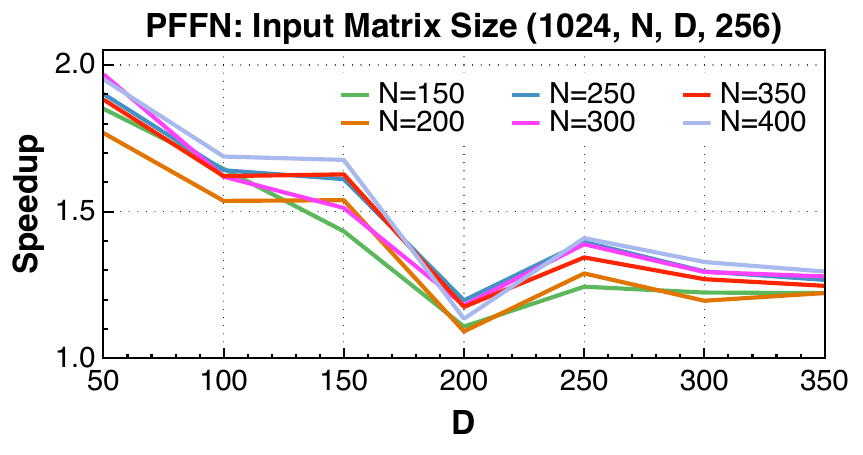}
    \end{subfigure}
\caption{PFFN speedup on production shapes. Left: Speedup as a function of batch size $B$ for five production $(N,D,K)$ configurations, showing peak performance of 2.0-2.4× at small batch sizes converging to 1.2-1.4× at large batches as kernel launch overhead amortization reduces fusion advantages. Right: Speedup variation with input dimension $D$ at fixed $B=1024, K=256$ across sequence lengths $N \in [150, 400]$, exhibiting non-monotonic behavior arising from tile size and SRAM capacity interactions.}

    \label{fig:pffn-analysis-prod}
\end{figure*}

\textbf{Performance Analysis.} Figure~\ref{fig:pffn-analysis-prod} analyzes speedup across production shape variations. The left panel shows performance as a function of batch size $B$ for five production configurations extracted from deployed InterFormer models. Peak speedups of 2.0-2.6× occur at small batch sizes ($B \leq 256$), where the fused kernel's reduced memory traffic dominates performance. As batch size increases beyond 512, speedup stabilizes at 1.2-1.4× across all configurations. This convergence reflects a fundamental trade-off: larger batches amortize kernel launch overhead for both optimized and baseline implementations, reducing the relative advantage of fusion as compute-to-memory ratio increases. The configuration $(N=200, D=256, K=160)$—representative of high-dimensional production embeddings—maintains consistent 1.2× speedup at large batch sizes, validating robust performance on primary deployment targets.

The right panel examines speedup as a function of input dimension $D$ with fixed $B=1024, K=256$ across varying sequence lengths $N \in [150, 400]$. Performance exhibits non-monotonic behavior: speedup peaks at 1.6-1.9× for small $D$ ($D \leq 100$), drops to a local minimum of 1.1-1.2× around $D=200$, then recovers to 1.2-1.4× for larger dimensions ($D > 200$). This pattern arises from the interplay between tile size and SRAM capacity. At small $D$, tiles fit comfortably in SRAM enabling effective fusion; at intermediate $D \approx 200$, tile dimensions approach SRAM limits causing partial spilling that negates fusion benefits; at large $D > 200$, the kernel adapts tiling strategy to maintain SRAM residency, recovering performance. Across all sequence lengths tested, curves follow similar trajectories—demonstrating that the optimization strategy generalizes across the $N$ dimension spanning production workloads.

Critically, all configurations maintain speedup $\geq 1.0×$ across the tested parameter space, with the majority achieving 1.2-2.0× improvements. The absence of performance regressions validates \name's shape-aware optimization approach: generated kernels exploit fusion opportunities when tile configurations permit on-chip execution, while avoiding pathological cases through adaptive tiling strategies discovered during search.


\textbf{Discussion.} The PFFN case study demonstrates \name's ability to discover non-obvious fusion opportunities through systematic search over operator compositions and tiling configurations. While human experts might identify the matrix multiplication and normalization fusion conceptually, determining the precise tile dimensions and reuse strategies that maximize SRAM occupancy across production shape distributions requires extensive trial-and-error—effort \name automates through graph-based search with execution feedback. The 1.5-2× speedups achieved on production workloads validate that automated synthesis can match expert-level kernel implementations while reducing development cycles from weeks to hours.

\subsection{Data Preprocessing Kernels on MTIA} 
\label{subsec:mtia-preprocessing}

MTIA is a custom silicon platform designed for both LLM and recommendation workloads. Unlike mature GPU ecosystems, MTIA presents unique kernel development challenges: not all PyTorch ATen operators are natively supported across hardware variants, and achieving optimal performance requires MTIA-specific tuning.

\paragraph{MTIA Kernel Coverage Challenge}
\label{subsubsec:mtia-coverage}

Running PyTorch models on MTIA requires kernel implementations for all ATen operators in the model graph. Missing kernels force either model rewrites or fallback to CPU execution, both unacceptable for production latency requirements. Table~\ref{tab:mtia-coverage} summarizes unsupported operators for two data preprocessing kernels across MTIA hardware generations.

\begin{table}[h!]
    \centering
    \small
    \begin{tabular}{lll}
        \toprule
        \textbf{AI Hardware} & \textbf{Data Preproc Operator} & \textbf{Missing ATen Ops} \\
        \midrule
        \multirow{2}{*}{MTIA v2i} 
        & \texttt{MapId} & \texttt{clamp.out}, \texttt{gather.out}, \texttt{sort.values\_stable}, \texttt{all.all\_out}, \texttt{\_unique2} \\
        & \texttt{MBDT} & \texttt{all.all\_out}, \texttt{unique\_consecutive} \\
        \midrule
        \multirow{2}{*}{MTIA v3} 
        & \texttt{MapId} & \texttt{clamp.out}, \texttt{sort.values\_stable}, \texttt{\_unique2} \\
        & \texttt{MBDT} & \texttt{unique\_consecutive} \\
        \bottomrule
    \end{tabular}
    \caption{Unsupported ATen operators on MTIA hardware for data preprocessing operators (MapId: MapIdTransform, MBDT: MergeBucketizedDenseTransform).}
    \label{tab:mtia-coverage}
\end{table}

\name addresses both model enablement and kernel optimization. On hardware with limited coverage (MTIA v2i), \name-generated kernels provide the missing implementations required for model execution. On hardware with higher coverage (MTIA v3), \name delivers performance gains through operator fusion and MTIA-specific tuning. The following sections evaluate two preprocessing kernels: MapIdTransform and MergeBucketizedDenseTransform (MBDT).

\subsubsection{MapId Transform}
\label{subsubsec:mapid}

MapIdTransform remaps sparse, high-cardinality categorical IDs to dense consecutive integers for embedding lookup.  Given a sorted mapping tensor $M$ containing known IDs, the kernel maps each input value to its 1-indexed position in $M$, reserving index 0 for unknown values.

\textbf{Algorithm.} For input tensor $V$ and sorted mapping $M$:
\begin{enumerate}[nosep,leftmargin=*]
    \item Binary search to find insertion index: $\text{idx}_j = \texttt{bucketize}(v_j, M)$
    \item Clamp index to valid range: $\text{idx}_j = \min(\text{idx}_j, |M|-1)$
    \item Validate match: if $M[\text{idx}_j] = v_j$, output $\text{idx}_j + 1$; else output $0$
\end{enumerate}

\begin{table}[h]
\centering
\small
\begin{tabular}{ll}
\toprule
\textbf{Input} & \texttt{values} = [100, 300, 500, 200, 999] \\
& \texttt{mapping} = [100, 200, 300, 400, 500] \\
\midrule
\textbf{Output} & [1, 3, 5, 2, 0] \\
\bottomrule
\end{tabular}
\captionof{table}{MapIdTransform example. Value 100 is at position 0 in mapping, outputs 1 (1-indexed). Value 300 is at position 2, outputs 3. Value 999 is not in mapping, outputs 0 (unknown).}
\label{tab:mapid-example}
\end{table}

The PyTorch reference implementation uses \texttt{torch.bucketize}, \texttt{torch.clamp}, \texttt{torch.gather}, and \texttt{torch.where}---operators partially unsupported on MTIA v2i (Table~\ref{tab:mtia-coverage}).

\begin{pythoncode}
class MapId(nn.Module):
    def forward(
        self,
        values: torch.Tensor,
    ) -> torch.Tensor:
        mapped_to_index = torch.clamp(
            torch.bucketize(values, self.mapping), max=self.mapping.numel() - 1
        )
        mapped_as_values = torch.gather(self.mapping, 0, mapped_to_index)
        mapped_values = torch.where(
            torch.eq(mapped_as_values, values), mapped_to_index + 1, 0
        )
        return mapped_values
\end{pythoncode}

\textbf{Generated Kernel Optimizations.} \name synthesizes a fused Triton kernel that consolidates four PyTorch operators into a single accelerator invocation, applying three MTIA-targeted optimizations:

\textit{(1) Operator Fusion.} The generated kernel fuses \texttt{bucketize}, \texttt{clamp}, \texttt{gather}, and \texttt{where} into a single kernel launch, eliminating three intermediate tensor materializations. Each thread block loads input values once, performs in-register binary search, validates matches, and writes final outputs---reducing global memory traffic by 4$\times$ compared to operator-by-operator execution.

\textit{(2) Compile-Time Loop Unrolling.} The binary search employs a fixed iteration bound of 20 steps, supporting mapping tables up to $2^{20}$ entries. This compile-time constant (\texttt{for \_ in range(20)}) enables aggressive loop unrolling by the Triton compiler, converting control flow into predicated straight-line code. Search bounds are maintained in registers using vectorized \texttt{tl.where} operations for branchless conditional updates:
\begin{pythoncode}
left = tl.where(search_active & (values > mapping_val), mid + 1, left)
right = tl.where(search_active & (values <= mapping_val), mid, right)
\end{pythoncode}

\textit{(3) Coalesced Block-Parallel Execution.} The kernel organizes work into contiguous blocks via \texttt{BLOCK\_SIZE} (a \texttt{tl.constexpr} parameter), where each program instance computes offsets as \texttt{block\_start + tl.arange(0, BLOCK\_SIZE)}. This layout ensures that adjacent threads access adjacent memory addresses, maximizing DRAM burst efficiency. Boundary conditions are handled through predicated loads and stores (\texttt{mask = offsets < n\_elements}), avoiding divergent control flow while maintaining full memory coalescing across the warp.

\begin{table}[t]
\centering
\small
\begin{tabular}{lcccc}
\toprule
\textbf{AI Hardware} & \textbf{Tensor Shape: \footnotesize{(UniqueIDs $\times$ Batch)}} & \textbf{PyTorch (ms)} & \textbf{Triton Kernel (ms)} & \textbf{Speedup} \\
\midrule
\multirow{10}{*}{MTIA v2i} 
& $100 \times 10000$ & 1.623 & 0.466 & \cellcolor{speedup}3.48× \\
& $500 \times 10000$ & 1.636 & 0.472 & \cellcolor{speedup}3.47× \\
& $1000 \times 10000$ & 1.641 & 0.480 & \cellcolor{speedup}3.42× \\
& $5000 \times 10000$ & 1.667 & 0.508 & \cellcolor{speedup}3.28× \\
\cmidrule{2-5}
& $10000 \times 2000$ & 0.399 & 0.514 & \cellcolor{slowdown}0.78× \\
& $10000 \times 4000$ & 0.720 & 0.521 & \cellcolor{speedup}1.38× \\
& $10000 \times 6000$ & 1.046 & 0.523 & \cellcolor{speedup}2.00× \\
& $10000 \times 8000$ & 1.367 & 0.520 & \cellcolor{speedup}2.63× \\
& $10000 \times 10000$ & 1.688 & 0.523 & \cellcolor{speedup}3.23× \\
& $10000 \times 50000$ & 8.090 & 1.989 & \cellcolor{speedup}4.07× \\
\midrule
\multirow{10}{*}{MTIA v3} 
& $100 \times 10000$ & 0.061 & 0.058 & \cellcolor{speedup}1.05× \\
& $500 \times 10000$ & 0.063 & 0.055 & \cellcolor{speedup}1.15× \\
& $1000 \times 10000$ & 0.063 & 0.050 & \cellcolor{speedup}1.26× \\
& $5000 \times 10000$ & 0.060 & 0.048 & \cellcolor{speedup}1.25× \\
\cmidrule{2-5}
& $10000 \times 2000$ & 0.039 & 0.035 & \cellcolor{speedup}1.11× \\
& $10000 \times 4000$ & 0.046 & 0.036 & \cellcolor{speedup}1.28× \\
& $10000 \times 6000$ & 0.049 & 0.037 & \cellcolor{speedup}1.32× \\
& $10000 \times 8000$ & 0.053 & 0.039 & \cellcolor{speedup}1.36× \\
& $10000 \times 10000$ & 0.063 & 0.048 & \cellcolor{speedup}1.31× \\
& $10000 \times 50000$ & 0.140 & 0.174 & \cellcolor{slowdown}0.80× \\
\bottomrule
\multicolumn{5}{c}{\footnotesize MTIA v2i / v3 : Triton Kernel. Color: \colorbox{speedup}{speedup}, \colorbox{slowdown}{regression}.} \\
\end{tabular}
\caption{MapIdTransform kernel performance: \name-generated Triton kernels vs. PyTorch baseline on MTIA v2i and v3.}
\label{tab:mapid_new}
\end{table}

\textbf{Performance Analysis.} Table~\ref{tab:mapid_new} compares \name-generated kernels against PyTorch on MTIA v2i and v3. As shown in Table~\ref{tab:mtia-coverage}, several ATen operators required by MapIdTransform lack native MTIA support, forcing PyTorch to execute CPU fallbacks with expensive host-device synchronization. \name synthesizes a fused Triton kernel executing entirely on-device, providing both functional enablement and performance optimization.

On MTIA v2i, \name achieves 3.28-3.48× speedup for large batches (batch size = 10000), with consistent performance across mapping table sizes. Performance scales with batch size: 0.78× at 2000 (launch overhead dominates), increasing to 1.38× at 4000, 2.00× at 6000, and 3.23× at 10000, reaching peak 4.07× at batch size 50000. This scaling reflects the trade-off between fixed kernel launch overhead and batch-dependent computation benefits.

On MTIA v3, latencies are substantially lower (0.035-0.174ms vs. 0.399-8.090ms on v1), reflecting improved hardware capabilities. Speedups are more modest at 1.05-1.36×, peaking at batch size 8000. The reduced gains arise from stronger PyTorch baselines on v2i due to improved native operator coverage and memory subsystem performance. 

Across both hardware generations, \name delivers robust speedups on large-batch workloads representative of production inference (up to 4.07$\times$ on v2i, 1.36$\times$ on v3). For edge cases where regressions occur on v2i and v3, runtime dispatch based on input dimensions ensures fallback to PyTorch, preventing performance degradation in deployment.


\subsubsection{MergeBucketizedDense Transform}

MBDT is a data preprocessing kernel that maps continuous features to discrete bin indices for embedding lookup in recommendation models. Given an input tensor and per-feature border lists, MBDT performs batched bucketization—a vectorized binary search assigning each value to its corresponding bin.

\textbf{Operation.} For input tensor $X \in \mathbb{R}^{F \times B}$ (F features, B batch size) and border lists $\{B_f\}_{f=1}^{F}$ where each $B_f = [b_1, b_2, \ldots, b_{K_f}]$ is sorted, the output $Y_{f,i}$ is:
\begin{equation}
Y_{f,i} = \min \left\{ k \mid X_{f,i} < B_f[k] \right\}
\end{equation}

\begin{figure}[t]
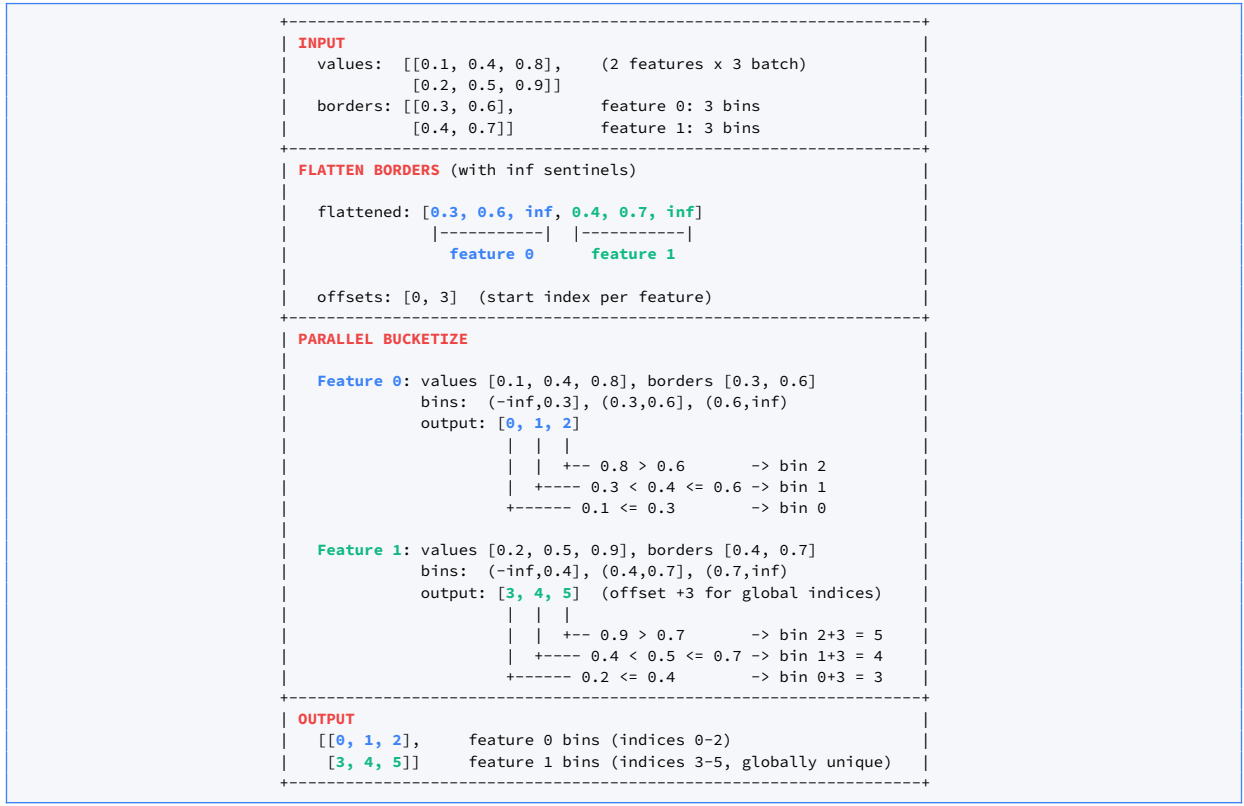

\centering
\begin{lstlisting}[style=mbdtstyle]
                            +-------------------------------------------------------------------+
                            | @INPUT@                                                             |
                            |   values:  [[0.1, 0.4, 0.8],    (2 features x 3 batch)            |
                            |             [0.2, 0.5, 0.9]]                                      |
                            |   borders: [[0.3, 0.6],         feature 0: 3 bins                 |
                            |             [0.4, 0.7]]         feature 1: 3 bins                 |
                            +-------------------------------------------------------------------+
                            | @FLATTEN BORDERS@ (with inf sentinels)                              |
                            |                                                                   |
                            |   flattened: [!0.3, 0.6, inf!, *0.4, 0.7, inf*]                       |
                            |               |-----------|  |-----------|                        |
                            |                 !feature 0!      *feature 1*                          |
                            |                                                                   |
                            |   offsets: [0, 3]  (start index per feature)                      |
                            +-------------------------------------------------------------------+
                            | @PARALLEL BUCKETIZE@                                                |
                            |                                                                   |
                            |   !Feature 0!: values [0.1, 0.4, 0.8], borders [0.3, 0.6]           |
                            |              bins:  (-inf,0.3], (0.3,0.6], (0.6,inf)              |
                            |              output: [!0, 1, 2!]                                    |
                            |                       |  |  |                                     |
                            |                       |  |  +-- 0.8 > 0.6       -> bin 2          |
                            |                       |  +---- 0.3 < 0.4 <= 0.6 -> bin 1          |
                            |                       +------ 0.1 <= 0.3        -> bin 0          |
                            |                                                                   |
                            |   *Feature 1*: values [0.2, 0.5, 0.9], borders [0.4, 0.7]           |
                            |              bins:  (-inf,0.4], (0.4,0.7], (0.7,inf)              |
                            |              output: [*3, 4, 5*]  (offset +3 for global indices)    |
                            |                       |  |  |                                     |
                            |                       |  |  +-- 0.9 > 0.7       -> bin 2+3 = 5    |
                            |                       |  +---- 0.4 < 0.5 <= 0.7 -> bin 1+3 = 4    |
                            |                       +------ 0.2 <= 0.4        -> bin 0+3 = 3    |
                            +-------------------------------------------------------------------+
                            | @OUTPUT@                                                            |
                            |   [[!0, 1, 2!],     feature 0 bins (indices 0-2)                    |
                            |    [*3, 4, 5*]]     feature 1 bins (indices 3-5, globally unique)   |
                            +-------------------------------------------------------------------+
\end{lstlisting}
\caption{MBDT execution example. \textcolor{red!70}{\textbf{Section headers}} in red. \textcolor{blue!70}{\textbf{Feature 0}} data in blue. \textcolor{green!50!black}{\textbf{Feature 1}} data in green. Borders are flattened with \texttt{inf} sentinels; output indices are offset per feature for global uniqueness.}
\label{fig:mbdt-example}
\end{figure}

\textbf{Example.} Figure~\ref{fig:mbdt-example} illustrates MBDT execution with 2 features × 3 batch elements. Feature 0 has borders $[0.3, 0.6]$ creating 3 bins; Feature 1 has borders $[0.4, 0.7]$ creating 3 bins. The preprocessing stage flattens all borders into a single array with \texttt{inf} sentinels marking boundaries, enabling $O(1)$ lookup of each feature's border range via offsets.

During parallel execution, each feature's values undergo binary search against their respective borders. For Feature 0, input $[0.1, 0.4, 0.8]$ maps to bins $[0, 1, 2]$: value 0.1 falls in bin 0 ($x \leq 0.3$), value 0.4 in bin 1 ($0.3 < x \leq 0.6$), and value 0.8 in bin 2 ($x > 0.6$). Feature 1's outputs are offset by 3 to ensure globally unique bin indices, yielding $[3, 4, 5]$ for inputs $[0.2, 0.5, 0.9]$. The final output tensor preserves the input shape with bin indices replacing continuous values.

\textbf{Implementation.} The core operation is a binary search over the borders array for each input value. The following pseudocode illustrates the sequential logic (the actual PyTorch baseline uses \texttt{torch.bucketize}):

\begin{pythoncode}
def mbdt_sequential(values, borders_list, offsets):
    # values: [F, B], borders_list: List[Tensor], offsets: [F]
    output = torch.empty_like(values, dtype=torch.int64)
    for f in range(num_features):
        for i in range(batch_size):
            # Binary search for bucket index
            idx = torch.bucketize(values[f, i], borders_list[f])
            output[f, i] = idx + offsets[f]
    return output
\end{pythoncode}

\noindent The offset parameter assigns globally unique bin indices across features (e.g., Feature 1's bins start at index 3 in the example above). The nested loops over features and batch elements present clear parallelization opportunities.

\begin{figure*}[t]
    \centering
    \begin{subfigure}[b]{0.49\textwidth}
        \includegraphics[width=\textwidth]{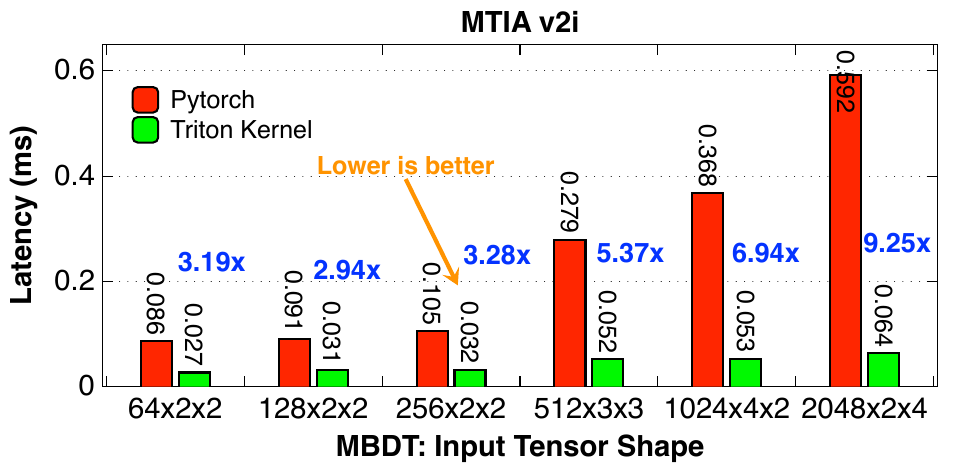}
    \end{subfigure}
    \hspace{-1em}
    \begin{subfigure}[b]{0.49\textwidth}
        \includegraphics[width=\textwidth]{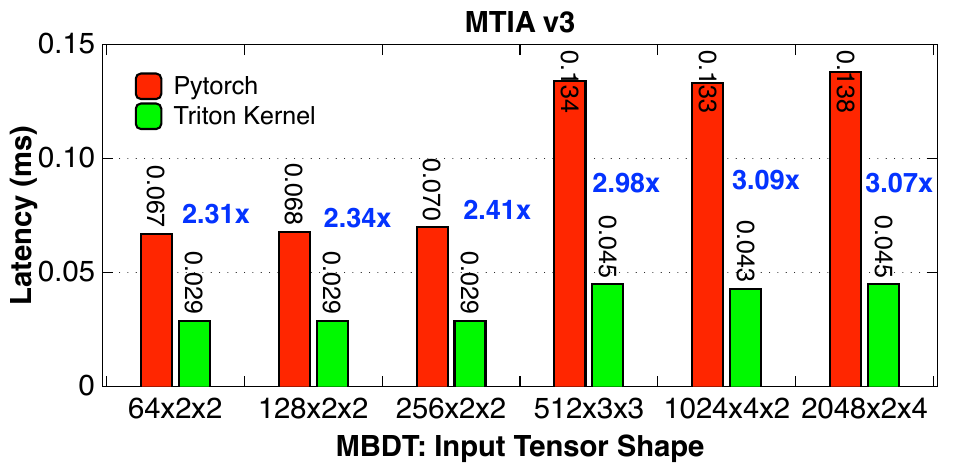}
    \end{subfigure}
    \caption{MBDT kernel latency comparison. Configuration format: Batch × Features × Borders. \name achieves 2.94--9.25× speedup on v2i and 2.31--3.09× on v3, with larger speedups at higher batch sizes.}
    \label{fig:mbdt-analysis}
\end{figure*}

\textbf{Performance Analysis.} Figure~\ref{fig:mbdt-analysis} compares \name-generated kernels against PyTorch (with \texttt{torch.compile}) across input configurations on MTIA v2i and v3. Configuration format is Batch × Features × Borders.

On MTIA v2i, \name achieves substantial speedups ranging from 2.94× to 9.25×. Speedup scales with input size: smaller configurations ($64 \times 2 \times 2$) achieve 3.19×, while larger configurations ($2048 \times 2 \times 4$) reach 9.25×. This scaling behavior reflects the kernel's ability to better amortize launch overhead and exploit parallelism as workload size increases.

On MTIA v3, absolute latencies are significantly lower (0.029--0.045ms vs. 0.027--0.064ms on v1), reflecting the next-generation hardware's improved compute throughput. \name achieves consistent speedups of 2.31--3.09× across all configurations. The lower speedup magnitude compared to v1 is expected: v2i has higher native operator coverage (Table~\ref{tab:mtia-coverage}), resulting in a stronger PyTorch baseline with less room for optimization. Nevertheless, kernel fusion and vectorized execution still deliver meaningful gains.

\textbf{Generated Kernel Optimizations.} \name generates a fused Triton kernel with several MTIA-specific optimizations:

\textit{Kernel Fusion.} The entire bucketization pipeline—border lookup, binary search, and offset computation—executes in a single kernel launch, eliminating inter-kernel communication.

\textit{Vectorized Counting.} Instead of scalar binary search, the kernel uses SIMD-vectorized counting: \texttt{values > border\_val} is applied to blocks of 64--256 elements simultaneously. For typical small border arrays (3--10 elements), this $O(n)$ approach outperforms $O(\log n)$ binary search due to reduced control flow overhead and branch-free execution.

\textit{Adaptive Block Sizing.} Block size is tuned based on input dimensions (64 for small, 128 for medium, 256 for large inputs) to maximize Processing Element utilization and ensure hardware saturation.

\textit{Register-Resident Computation.} Intermediate results (left/right counts, averages) remain in registers throughout computation. No intermediate tensor allocations occur; results write directly to the output buffer.

\textit{MTIA-Specific Patterns.} The kernel avoids constructs that fail MTIA compilation (e.g., \texttt{tl.where} in loops), using direct boolean-to-int conversion instead. Memory loads are coalesced with proper masking, and small border arrays are cached across processing blocks.

\subsubsection{Summary}
\label{subsubsec:mtia-summary}

The MapIdTransform and MBDT evaluations demonstrate \name's dual value proposition on emerging hardware platforms: \textit{enablement} and \textit{optimization}. On MTIA v2i, where native operator coverage is limited, \name-generated kernels provide the only viable on-device execution path—without them, PyTorch falls back to CPU for unsupported operators, incurring order-of-magnitude latency penalties. On MTIA v3, where coverage is more complete, \name still delivers 2--3× speedups through kernel fusion and hardware-specific tuning.

These results highlight a key insight: as new accelerators emerge, the gap between hardware availability and software ecosystem maturity creates a critical need for automated kernel generation. Traditional approaches—waiting for vendor libraries or hand-tuning by kernel experts—cannot scale to the diversity of operators and hardware variants in production. \name addresses this by generating correct, optimized kernels from high-level operator specifications, reducing the time from hardware deployment to production readiness from months to hours.

Looking forward, we envision \name as a foundational tool for heterogeneous accelerator ecosystems. As Meta continues to deploy next-generation MTIA hardware alongside NVIDIA and AMD GPUs, the ability to rapidly generate and optimize kernels across platforms becomes increasingly critical.

\subsection{Sequence Learning: Batch Event Truncate}
\label{subsubsec:event-truncate}

Event-based features (EBF)~\cite{reddy2024seqlearning} are a time-ordered sequence of interactions of a specific type, such as ad\_impression, where each interaction contains multiple feature values (e.g., ad display format types, page id). EBF encodes user behavior sequences in ads ranking models as nested jagged tensors. Each batch contains three tensors: \texttt{outer\_lengths} specifies the number of events per user, \texttt{inner\_lengths} specifies the number of attributes per event, and \texttt{values} stores the flattened attribute data.

\textbf{Operation.} Figure~\ref{fig:event-truncate-example} illustrates the Batch Event Truncate operation with multiple features. The input batch contains three users with [3, 4, 1] events respectively, and two features per event: Feature 0 (e.g., ad display format) with variable-length attributes, and Feature 1 (e.g., page id) with uniform single attributes. For Feature 0, User 0 has events with [1, 0, 2] attributes containing values [1, 2, 3]; User 1 has events with [0, 3, 1, 1] attributes containing values [4, 5, 6, 7, 8]; User 2 has a single event with 1 attribute containing value [9]. For Feature 1, all events have single attributes, with values [1, 2, 3] for User 0, [4, 5, 6, 7] for User 1, and [8] for User 2.
When truncating to $N$=2 events, the operator retains only the first two events per user across all features simultaneously. For User 0, the third event is discarded—removing 2 attributes (values [2, 3]) from Feature 0 and 1 attribute (value [3]) from Feature 1. For User 1, the third and fourth events are removed—discarding attributes [1, 1] (values [7, 8]) from Feature 0 and attributes [1, 1] (values [6, 7]) from Feature 1. User 2 is unchanged since it has fewer than $N$ events.

The operation requires coordinated index arithmetic across three nested levels and multiple features—a pattern poorly suited to standard tensor primitives. In production, sequences can reach 200 events with 5--32 features, making efficient batched truncation critical for serving latency and motivating a custom Triton kernel that processes all features in a single launch.

\definecolor{batch0}{RGB}{230, 243, 255}  
\definecolor{batch1}{RGB}{255, 243, 230}  
\definecolor{batch2}{RGB}{230, 255, 230}  
\definecolor{truncated}{RGB}{255, 180, 180}  

\begin{figure}[t]
\centering
\small
\begin{tabular}{lll}
\toprule
& \textbf{Feature 0} (ad format) & \textbf{Feature 1} (page id) \\
\midrule
\multicolumn{3}{l}{\textbf{Input} (3 users with [3, 4, 1] events)} \\
\midrule
\texttt{outer\_len} & \texttt{[\colorbox{batch0}{3}, \colorbox{batch1}{4}, \colorbox{batch2}{1}]} & \texttt{[\colorbox{batch0}{3}, \colorbox{batch1}{4}, \colorbox{batch2}{1}]} \\[0.8em]
\texttt{inner\_len} & \texttt{[\colorbox{batch0}{1, 0, \colorbox{truncated}{2}}, \colorbox{batch1}{0, 3, \colorbox{truncated}{1, 1}}, \colorbox{batch2}{1}]} & \texttt{[\colorbox{batch0}{1, 1, \colorbox{truncated}{1}}, \colorbox{batch1}{1, 1, \colorbox{truncated}{1, 1}}, \colorbox{batch2}{1}]} \\[0.8em]
\texttt{values} & \texttt{[\colorbox{batch0}{1, \colorbox{truncated}{2, 3}}, \colorbox{batch1}{4, 5, 6, \colorbox{truncated}{7, 8}}, \colorbox{batch2}{9}]} & \texttt{[\colorbox{batch0}{1, 2, \colorbox{truncated}{3}}, \colorbox{batch1}{4, 5, \colorbox{truncated}{6, 7}}, \colorbox{batch2}{8}]} \\
\midrule
\multicolumn{3}{l}{\textbf{Output} (truncate to $N$=2 events)} \\
\midrule
\texttt{outer\_len} & \texttt{[\colorbox{batch0}{2}, \colorbox{batch1}{2}, \colorbox{batch2}{1}]} & \texttt{[\colorbox{batch0}{2}, \colorbox{batch1}{2}, \colorbox{batch2}{1}]} \\[0.8em]
\texttt{inner\_len} & \texttt{[\colorbox{batch0}{1, 0}, \colorbox{batch1}{0, 3}, \colorbox{batch2}{1}]} & \texttt{[\colorbox{batch0}{1, 1}, \colorbox{batch1}{1, 1}, \colorbox{batch2}{1}]} \\[0.8em]
\texttt{values} & \texttt{[\colorbox{batch0}{1}, \colorbox{batch1}{4, 5, 6}, \colorbox{batch2}{9}]} & \texttt{[\colorbox{batch0}{1, 2}, \colorbox{batch1}{4, 5}, \colorbox{batch2}{8}]} \\
\bottomrule
\end{tabular}
\caption{Batch Event Truncate with multiple features. Colors indicate users (\colorbox{batch0}{User 0}, \colorbox{batch1}{User 1}, \colorbox{batch2}{User 2}). Nested \colorbox{truncated}{red} highlights data discarded when truncating to $N$=2 events—showing which user's data is being removed.}
\label{fig:event-truncate-example}
\end{figure}

The original PyTorch implementation processes each feature event sequence independently in a loop—no batched variant existed due to the complexity of coordinating index arithmetic across nested jagged tensors. \name automatically generates a batched Triton kernel that processes multiple features in parallel, a non-trivial optimization that would require significant manual engineering effort.

\textbf{Performance Analysis.} Table~\ref{tab:event-truncate} compares the \name-generated batched Triton kernel against the non-batched PyTorch baseline. The PyTorch implementation processes each feature sequentially, while the Triton kernel batches all features into a single launch.

\begin{table}[t]
\centering
\small
\begin{tabular}{lcccccc}
\toprule
\textbf{Description} & \textbf{Feature Counts} & \textbf{Events} & \textbf{Max Length} & \textbf{PyTorch} & \textbf{Triton} & \textbf{Speedup} \\
& & & & (ms) & (ms) & \\
\midrule
Single feature & 1 & 200 & 100 & 0.148 & 0.313 & 1.0× \\
Single feature & 1 & 200 & 200 & 0.148 & 0.109 & \cellcolor{speedup}1.4× \\
\midrule
Prod multi-feature & 5 & 200 & 100 & 0.788 & 0.571 & \cellcolor{speedup}1.4× \\
Prod multi-feature & 9 & 200 & 200 & 1.443 & 0.148 & \cellcolor{speedup}\textbf{9.8×} \\
\midrule
Large feature count & 32 & 200 & 100 & 5.078 & 2.548 & \cellcolor{speedup}2.0× \\
Large feature count & 32 & 200 & 200 & 5.085 & 0.350 & \cellcolor{speedup}\textbf{14.5×} \\
\bottomrule
\end{tabular}
\caption{Batch Event Truncate performance: \name-generated batched Triton kernel vs. non-batched PyTorch baseline.}
\label{tab:event-truncate}
\end{table}

Two factors drive the performance difference. First, when no truncation is needed (Max $N$ $\geq$ actual event count), PyTorch loops through each batch element comparing lengths individually, while the batched kernel performs a single vectorized comparison—yielding 9.8× and 14.5× speedups at higher feature counts. Second, when truncation is required, the batched kernel uses constant kernel launches for parallel processing versus PyTorch's sequential iteration, achieving 1.4--2.0× speedups. 

In production end-to-end benchmarks, the batched kernel achieves 2× speedup over the PyTorch implementation. Notably, speedup scales with batch size, indicating that the batched kernel enables further model scaling by supporting additional event-based features without proportional latency increase. These results demonstrate that \name can generate efficient batched implementations for operators where only sequential baselines exist, enabling significant latency reductions in production serving.

\section{Future Directions}
\label{sec:future}
\name demonstrates that LLM agents can generate production-quality kernels for heterogeneous accelerators, but this work represents only the first step toward a broader vision: \textit{fully automated, heterogeneous hardware-aware code generation that scales across the entire AI infrastructure stack at Meta}.

\textbf{Heterogeneous Hardware at Scale.} As Meta's AI infrastructure evolves to include next-generation MTIA, AMD MI series, ARM CPUs, and future NVIDIA architectures, the diversity of optimization targets will grow exponentially. We envision \name as the unified kernel generation layer across this heterogeneous fleet—automatically adapting to new hardware through updated specifications rather than manual engineering. This requires developing hardware abstraction primitives that capture memory hierarchies, compute capabilities, and ISA constraints in a format amenable to LLM reasoning.

\textbf{From Operators to Models.} Current optimizations target individual operators and small modules, but the greatest performance gains lie at model level. Future work will extend \name to reason about cross-layer fusion, global memory allocation, and end-to-end computation graphs. Combined with model transformation techniques—quantization, sparsity, architecture search—this enables co-optimization of model structure and kernel implementation, potentially discovering novel operator compositions that neither humans nor traditional compilers would identify.

\textbf{Deeper Code Generation.} Triton provides a productive abstraction, but certain optimizations require lower-level control. Extending \name to modify MLIR dialects, direct PTX/SASS, or hardware diagnostic routines would unlock performance-critical scenarios where Triton's abstractions become limiting. This vertical integration—from high-level DSLs to bare-metal code—positions LLM agents as general-purpose compilers rather than domain-specific tools.

\textbf{Massively Parallel Search.} The current tree search explores candidates sequentially, but kernel generation is inherently parallelizable. Relaxing consistency guarantees enables thousands of candidates to be evaluated simultaneously across distributed infrastructure, with eventual convergence to optimal solutions. This "infinite-width" search paradigm, combined with inference-time scaling laws, suggests that kernel quality may improve predictably with compute investment—a compelling property for production systems with strict performance targets.

\textbf{Hardware-Specific Adaptation.} Foundation models lack knowledge of proprietary accelerators like MTIA. Reinforcement learning from execution feedback—rewarding compilation success, correctness, and latency—offers a path to adapt generic models to specialized hardware without exposing proprietary details. This approach could enable rapid onboarding of new accelerators: deploy hardware, collect execution traces, fine-tune the agent, and generate optimized kernels within days rather than months.

\textbf{End-to-End Production Integration.} The ultimate goal is seamless integration from model definition to production deployment. \name should interface with AOT Inductor, model serving infrastructure, and continuous integration pipelines—automatically generating, validating, and deploying kernels as models and hardware evolve. This closed-loop system would reduce the human effort required to maintain performance across Meta's rapidly evolving AI stack.

\textbf{Sustainable AI Infrastructure.} As LLM-based generation scales, resource efficiency becomes critical. Future work will quantify token consumption per kernel, optimize prompt design for minimal inference cost, and track carbon footprint across the search process~\cite{wang2025catransformerscarbonawaretransformers}. These metrics enable principled trade-offs between optimization depth and environmental impact, aligning \name with broader sustainability goals in AI infrastructure~\cite{wu2022sustainableaienvironmentalimplications}.

\section{Related Work}
\label{sec:related}

\textbf{Kernel Optimization.} High-performance kernel development has traditionally relied on vendor-optimized libraries (cuBLAS, cuDNN, rocBLAS) and auto-tuning frameworks. Halide decouples algorithm specification from scheduling, enabling portable optimization across hardware targets~\cite{ragan2013halide}. TVM extends this with learned cost models for automated schedule search~\cite{chen2018tvm}. Triton provides a Python-embedded DSL that abstracts GPU programming at the block level while exposing performance-critical tiling decisions~\cite{tillet2019triton}. Recent work introduces higher-level and lower-level abstractions: NVIDIA's CuTe DSL~\cite{nvdia-cute} as part of CUTLASS v4 provides composable layout and tensor abstractions for Tensor Core programming with Python JIT compilation. Meta's TLX (Triton Low-Level Extensions)~\cite{tlx} adds warp-aware intrinsics and explicit pipeline control for NVIDIA GPU Hopper and Blackwell architectures. OpenAI's Gluon dialect~\cite{gluon} exposes lower-level layout encoding within the Triton compiler stack whereas TileLang~\cite{wang2504tilelang} offers a composable tiled programming model with automatic layout inference across NVIDIA and AMD GPUs and Helion~\cite{helion} compiles high-level PyTorch-like syntax to autotuned Triton code. These abstractions reduce development effort but still require substantial domain expertise for novel kernel transformations and struggle to generalize across heterogeneous hardware without manual adaptation.

\textbf{LLM-Based Code and Kernel Generation.} Beyond traditional kernel optimization, earlier work have  applied evolutionary computations to improve general-purpose GPU code, such as machine learning kernels, by designing code transformation operators at the LLVM / MLIR level~\cite{Liou-GECCO2020, Liou-GI2019, Liou-IISWC2022, Liou-TACO2020}. And, more recently, large language models (LLM) have demonstrated remarkable capabilities in general-purpose code synthesis~\cite{chen2021evaluating, li2022competition, roziere2023code,fair-CodeGen2025}. Recent work extends these capabilities to performance-critical domains: AlphaCode~\cite{li2022competition,alphacode2} achieves competitive programming performance through large-scale sampling, while CodeRL~\cite{le2022coderl} incorporates execution feedback via reinforcement learning. These advances establish the foundation for applying LLMs to kernel development, where correctness and performance speedup are both essential.

\textbf{Inference Time Scaling.} Several recent systems start to apply LLMs specifically for GPU kernel synthesis. KernelBench~\cite{ouyang2025kernelbench} benchmarks LLM capabilities across operator, fusion, and model-level difficulty tiers. 
AutoTriton~\cite{li2025autotriton} applies RL to Triton programming and KernelLLM~\cite{kernelllm2025} explored supervised
baseline for Triton kernel generation from
PyTorch modules, while TritonRL~\cite{woo2025tritonrl} trains models with execution-guided rewards. 
KernelAgent~\cite{kernelagent} was designed around KernelLLM to generate verified Triton kernels from PyTorch programs based on a multi-agent system.
GEAK~\cite{wang2025geak} targets AMD MI300X through agentic workflows, and Kevin~\cite{baronio2025kevin} employs multi-turn RL for CUDA generation. More recently, TritorX has been introduced as an agentic AI system that aims to generate functionally correct Triton kernels from PyTorch
ATen operators~\cite{hammond2025agenticoperatorgenerationml} -- an important first step to enable automatic kernel generation for production deployed models to run on MTIAs. AlphaEvolve~\cite{novikov2025alphaevolvecodingagentscientific} leverages evolutionary search with LLMs to optimize select stages of TPU/GPU kernels. While these systems demonstrate competitive results on isolated benchmarks, they target single hardware platforms with synthetic workloads, lacking heterogeneous hardware support, production operator coverage, and deployment infrastructure integration required for industry-scale adoption.

Finally, recent work demonstrates that model performance improves predictably with increased test-time compute~\cite{snell2024scaling}. Chain-of-thought prompting~\cite{wei2022chain}, tree-of-thought search~\cite{yao2023tree}, and self-consistency decoding~\cite{wang2022self} can enable complex reasoning through multi-path exploration. OpenAI's o1 and DeepSeek-R1~\cite{guo2025deepseek} demonstrate substantial gains on mathematical and coding benchmarks through scaled inference. \name builds on these insights, applying a tree-based search algorithm with execution feedback to systematically explore kernel optimization spaces for a wide collection of heterogeneous AI hardware. Other search algorithms, such as, ~\cite{Toledo-NeurIPS2025}, can be easily integrated to further improve \name's design space exploration and performance optimization results. 

Overall, \name differs from key prior work along the following three axes:
\begin{itemize}
    \item \textit{First}, it targets \emph{heterogeneous hardware at scale} --- generating optimized kernels for NVIDIA GPUs, AMD GPUs, and Meta's custom MTIA accelerators from unified operator specifications. MTIA presents a unique challenge: as a proprietary architecture absent from public training corpora, effective kernel generation requires systematic knowledge injection of hardware constraints and programming idioms (Section~\ref{subsubsec:mtia-knowledge}).
    \item \textit{Second}, \name addresses \emph{production operator diversity} beyond canonical benchmarks. Ads ranking models employ 200+ preprocessing operators with irregular access patterns and data-dependent control flow --- operators that determine deployment architecture rather than merely affecting performance (Section~\ref{sec:intro}).
    \item \textit{Third}, it provides \emph{deployment-integrated optimization} with continuous validation, multi-level profiling (system, kernel, intra-kernel), and serving infrastructure compatibility, enabling safe production rollout.
\end{itemize}

To our knowledge, \name is the 
first LLM-based kernel coding system deployed at-scale for business-critical 
recommendation model inference.

\section{Conclusion}
\label{sec:conclusion}
This paper presents \name, an agentic kernel generation framework addressing the three-dimensional diversity challenge in large-scale AI infrastructure: hardware heterogeneity across vendors and generations, model architectural diversity spanning retrieval to ranking stages, and kernel diversity from preprocessing to compute-intensive primitives. By formulating kernel optimization as graph-based search with retrieval-augmented prompting and persistent knowledge bases, \name achieves 100\% correctness across 480 operator-platform configurations and delivers 1.25-17× speedups on production workloads, while reducing development time from weeks to hours.

Beyond performance gains, \name demonstrates a fundamental shift in how we approach heterogeneous accelerator deployment. Incomplete kernel coverage forces costly disaggregated architectures with 10-20ms network overhead; automated synthesis eliminates these architectural penalties, transforming kernel availability from a deployment blocker to an enabler. As AI infrastructure evolves toward increasingly diverse accelerator fleets—integrating custom silicon, next-generation GPUs, and specialized processors—the gap between hardware innovation and software ecosystem maturity widens. Manual kernel development cannot scale to this combinatorial explosion of operators, models, and platforms.

We envision a future where LLM agents serve as the universal compilation layer for heterogeneous AI systems, automatically adapting to new hardware through knowledge injection rather than manual porting. \name represents a first step toward this vision: demonstrating that agents can generate production-quality kernels for proprietary accelerators absent from training corpora, achieve competitive performance through inference-time scaling, and operate continuously in mission-critical infrastructure serving more than hundreds of trillions of daily inferences. The insights from this deployment—multi-granularity profiling integration, automated validation frameworks, and architecture-aware optimization—establish design principles for AI-assisted systems software that we believe will prove essential as the hardware landscape continues to fragment and diversify. The path forward requires not just better kernels, but fundamentally new approaches to bridging the hardware-software gap at scale.

\newpage

\section{Contributors}
\newcommand{\role}[1]{\par\vspace{3mm}\noindent\textbf{#1}\par\vspace{1mm}}
\label{sec:contributors}

\role{Project Leads}
\begin{itemize}
    \item Gang Liao, Gaoxiang Liu
\end{itemize}

\role{Core Contributors}
\begin{itemize}
    \item \textbf{Monetization Infra and Ranking}: Gang Liao, Hongsen Qin, Ying Wang, Yavuz Yetim, Jia Jiunn Ang, Xiayu Yu, Yihan He, Feng Shi, Zewei Jiang, Chunli Fu, Ruichao Xiao, Dianshi Li, Gaoxiang Liu
    \item \textbf{FAIR SysML}: Alicia Golden, Michael Kuchnik, Samuel Hsia, Carole-Jean Wu
    \item \textbf{MTIA Software}: Roman Levenstein, Kunming Ho, Haishan Zhu, Alec Hammond, Richard Li, Ajit Mathews, Kaustubh Gondkar
\end{itemize}

\role{Contributors}
\begin{itemize}
    \item \textbf{Ad Use Cases}: Zhou Fang, Abdul Zainul-Abedin, Ketan Singh
    \item \textbf{Serverless Compute}: Sean Zhang, Noah Weller, Zach Marine, Wyatt Cook
    \item \textbf{AI Compilers \& Runtimes}: Hongtao Yu, Wenyuan Chi, Barney Huang
\end{itemize}

\role{Supervision}
\begin{itemize}
    \item Liyuan Li, Nathan Yan, Varna Puvvada, Uladzimir Pashkevich, Matt Steiner
\end{itemize}

\section{Acknowledgments}
  Yoram Bachrach, Rick Chang, Yuanwei Fang, Jun Ge, Danilo Carvalho Grael, Karen Hambardzumyan, Nathan Hu, Keren Huang, Joe Isaacson, Martin Josifoski, Minjang Kim, Irene Liu, Alexey Loginov, Abhi Pandey, Srivatsan Ramesh, Praveen Ramachandran, Mark Saroufim, Dev (Devashish) Shankar, Rithin Shetty, Bidit Sharma, Jake Siso, Oleksandr Stashuk, Tejas Venkateswaran, Edan Toledo, Laura Wang, Shiguang Wang, Zhaodong Wang, Shawn Xu, Feixiong Zhang, Xudong Zhang, Mingjie Zhu



\clearpage
\newpage
\bibliographystyle{assets/plainnat}
\bibliography{paper}

\clearpage
\newpage
\beginappendix

\section{1D Convolution}
\label{appendix:conv1d}

\textbf{Note:} The candidate presented here was randomly selected from a pool of high-performing solutions rather than representing the single best-performing variant. Due to data confidentiality constraints, production data could not be used directly. Instead, we employ synthetic data that preserves the statistical properties (Min, Max, Mean, and STD) of the original production dataset. This synthetic dataset has been validated to yield equivalent speedup characteristics to those observed with production data. Additionally, the internal Triton trunk version used in production differs slightly from the open-source release.

\begin{pythoncode}
"""
Auto-generated TritonBench Operator for conv1d

================================================================================
GENERATION METADATA
================================================================================
Generated by:    kernel_evolve CLI (TritonBench Operator generator)
Generated at:    2025-11-21 03:14:31
Operator name:   conv1d
Precision:       fp16
Device:          cuda
Output dir:      /tmp/kernel_evolve_eval
================================================================================

This file was automatically generated by the kernel_evolve evals framework.
DO NOT EDIT THIS FILE MANUALLY - regenerate it using the CLI tool.
"""

import argparse
import sys
from typing import Any, Callable, Generator, List, Optional

# ============================================================================
# Kernel Evolve Generated Models
# ============================================================================

import torch
import torch.nn as nn
import triton
import triton.language as tl
from tritonbench.utils.triton_op import BenchmarkOperator, register_benchmark

def generate_synthetic_production_data(dtype=torch.float32, device="cuda"):
    """
    Generate synthetic data matching production data characteristics.

    Creates realistic test data with controlled statistical properties
    suitable for performance evaluation.

    Production data statistics (analyzed 2025-12-12):
    - Input: shape=(2048, 96, 200), dtype=torch.float32, mean=0.006413, std=0.041105
    - Weight: shape=(96, 96, 3), dtype=torch.float32, mean=-0.000024, std=0.047870
    - Bias: shape=(96,), dtype=torch.float32, mean=-0.003198, std=0.033662

    Args:
        dtype: Data type for tensors (default: torch.float32, matching production data)
        device: Device to place tensors on (default: "cuda")

    Returns:
        Tuple of (input_tensor, weight_tensor, bias_tensor)
    """
    torch.manual_seed(42)

    # Production data shape parameters
    max_batch_size = 2048
    in_channels = 96
    out_channels = 96
    seq_length = 200
    kernel_size = 3

    input_full = (
        torch.randn(max_batch_size, in_channels, seq_length, device=device, dtype=dtype)
        * 0.041105
        + 0.006413
    )

    weight_tensor = (
        torch.randn(out_channels, in_channels, kernel_size, device=device, dtype=dtype)
        * 0.047870
        - 0.000024
    )

    bias_tensor = (
        torch.randn(out_channels, device=device, dtype=dtype) * 0.033662 - 0.003198
    )

    return input_full, weight_tensor, bias_tensor

# Global cache to avoid re-packing weights on every call
# Set DISABLE_WEIGHT_CACHE=True to measure "cold start" performance including weight packing
_WEIGHT_PACK_CACHE = {}
_DISABLE_WEIGHT_CACHE = True  # Set to True to benchmark without caching

class PytorchModel(nn.Module):
    def __init__(self) -> None:
        super().__init__()

    def forward(self, input_tensor, weight_tensor, bias_tensor, use_conv2d=False):
        if use_conv2d:
            input_tensor = input_tensor.unsqueeze(2).to(
                memory_format=torch.channels_last
            )
            weight_tensor = weight_tensor.unsqueeze(2)
            return torch.nn.functional.conv2d(
                input_tensor,
                weight_tensor,
                bias_tensor,
                stride=(1, 1),
                padding=(0, 1),
                dilation=1,
            ).squeeze(2)
        else:
            return torch.nn.functional.conv1d(
                input_tensor,
                weight_tensor,
                bias_tensor,
                stride=1,
                padding=1,
                dilation=1,
                groups=1,
            )


# ===========================
# Autotune configs for conv kernel
# Expanded and tuned for M-heavy shapes and small K
# ===========================
autotune_configs = [
    # Small/latency-oriented
    triton.Config(
        {
            "BLOCK_M": 32,
            "BLOCK_N": 64,
            "BLOCK_K": 32,
            "GROUP_M": 8,
            "WARP_SPECIALIZE": False,
        },
        num_warps=2,
        num_stages=3,
    ),
    triton.Config(
        {
            "BLOCK_M": 32,
            "BLOCK_N": 128,
            "BLOCK_K": 32,
            "GROUP_M": 8,
            "WARP_SPECIALIZE": False,
        },
        num_warps=4,
        num_stages=2,
    ),
    triton.Config(
        {
            "BLOCK_M": 64,
            "BLOCK_N": 32,
            "BLOCK_K": 32,
            "GROUP_M": 8,
            "WARP_SPECIALIZE": False,
        },
        num_warps=2,
        num_stages=4,
    ),
    # Balanced
    triton.Config(
        {
            "BLOCK_M": 64,
            "BLOCK_N": 64,
            "BLOCK_K": 32,
            "GROUP_M": 4,
            "WARP_SPECIALIZE": False,
        },
        num_warps=4,
        num_stages=2,
    ),
    triton.Config(
        {
            "BLOCK_M": 64,
            "BLOCK_N": 128,
            "BLOCK_K": 32,
            "GROUP_M": 4,
            "WARP_SPECIALIZE": False,
        },
        num_warps=8,
        num_stages=2,
    ),
    triton.Config(
        {
            "BLOCK_M": 128,
            "BLOCK_N": 64,
            "BLOCK_K": 32,
            "GROUP_M": 4,
            "WARP_SPECIALIZE": False,
        },
        num_warps=4,
        num_stages=3,
    ),
    triton.Config(
        {
            "BLOCK_M": 128,
            "BLOCK_N": 128,
            "BLOCK_K": 32,
            "GROUP_M": 4,
            "WARP_SPECIALIZE": False,
        },
        num_warps=8,
        num_stages=3,
    ),
    # Deeper K tiles
    triton.Config(
        {
            "BLOCK_M": 64,
            "BLOCK_N": 64,
            "BLOCK_K": 64,
            "GROUP_M": 8,
            "WARP_SPECIALIZE": False,
        },
        num_warps=4,
        num_stages=3,
    ),
    triton.Config(
        {
            "BLOCK_M": 128,
            "BLOCK_N": 64,
            "BLOCK_K": 64,
            "GROUP_M": 8,
            "WARP_SPECIALIZE": False,
        },
        num_warps=4,
        num_stages=4,
    ),
    triton.Config(
        {
            "BLOCK_M": 64,
            "BLOCK_N": 128,
            "BLOCK_K": 64,
            "GROUP_M": 8,
            "WARP_SPECIALIZE": False,
        },
        num_warps=8,
        num_stages=3,
    ),
    triton.Config(
        {
            "BLOCK_M": 128,
            "BLOCK_N": 128,
            "BLOCK_K": 64,
            "GROUP_M": 8,
            "WARP_SPECIALIZE": False,
        },
        num_warps=8,
        num_stages=4,
    ),
    # M-heavy problems
    triton.Config(
        {
            "BLOCK_M": 256,
            "BLOCK_N": 64,
            "BLOCK_K": 32,
            "GROUP_M": 8,
            "WARP_SPECIALIZE": False,
        },
        num_warps=8,
        num_stages=4,
    ),
    triton.Config(
        {
            "BLOCK_M": 256,
            "BLOCK_N": 128,
            "BLOCK_K": 32,
            "GROUP_M": 8,
            "WARP_SPECIALIZE": False,
        },
        num_warps=8,
        num_stages=4,
    ),
    triton.Config(
        {
            "BLOCK_M": 256,
            "BLOCK_N": 64,
            "BLOCK_K": 64,
            "GROUP_M": 8,
            "WARP_SPECIALIZE": False,
        },
        num_warps=8,
        num_stages=5,
    ),
    triton.Config(
        {
            "BLOCK_M": 256,
            "BLOCK_N": 128,
            "BLOCK_K": 64,
            "GROUP_M": 8,
            "WARP_SPECIALIZE": False,
        },
        num_warps=8,
        num_stages=5,
    ),
    # Aggressive N-tiling
    triton.Config(
        {
            "BLOCK_M": 128,
            "BLOCK_N": 256,
            "BLOCK_K": 32,
            "GROUP_M": 4,
            "WARP_SPECIALIZE": False,
        },
        num_warps=16,
        num_stages=3,
    ),
    triton.Config(
        {
            "BLOCK_M": 64,
            "BLOCK_N": 256,
            "BLOCK_K": 32,
            "GROUP_M": 8,
            "WARP_SPECIALIZE": False,
        },
        num_warps=16,
        num_stages=3,
    ),
    # Larger K tiles to saturate tensor cores
    triton.Config(
        {
            "BLOCK_M": 128,
            "BLOCK_N": 128,
            "BLOCK_K": 128,
            "GROUP_M": 4,
            "WARP_SPECIALIZE": False,
        },
        num_warps=8,
        num_stages=4,
    ),
    triton.Config(
        {
            "BLOCK_M": 256,
            "BLOCK_N": 64,
            "BLOCK_K": 128,
            "GROUP_M": 8,
            "WARP_SPECIALIZE": False,
        },
        num_warps=8,
        num_stages=5,
    ),
    # Very large M-tiles for huge sequences/batches
    triton.Config(
        {
            "BLOCK_M": 512,
            "BLOCK_N": 64,
            "BLOCK_K": 64,
            "GROUP_M": 8,
            "WARP_SPECIALIZE": False,
        },
        num_warps=8,
        num_stages=5,
    ),
    triton.Config(
        {
            "BLOCK_M": 512,
            "BLOCK_N": 128,
            "BLOCK_K": 64,
            "GROUP_M": 8,
            "WARP_SPECIALIZE": False,
        },
        num_warps=16,
        num_stages=5,
    ),
]


# ===========================
# Weight packing kernel: [Cout, Cin_g, Ksz] -> [Cin_g*Ksz, Cout]
# Vectorized copy with 2D tiling. This improves global load/store efficiency.
# ===========================
@triton.jit

def pack_conv1d_weight_kernel(
    w_in_ptr,  
    w_out_ptr, 
    Cout,
    Cin_g,
    Ksz,
    BLOCK_R: tl.constexpr,  # tile along rows (Cin_g*Ksz)
    BLOCK_C: tl.constexpr,  # tile along cols (Cout)
):
    R = Cin_g * Ksz  # rows of packed matrix
    C = Cout  # cols of packed matrix
    pid_r = tl.program_id(0)
    pid_c = tl.program_id(1)

    offs_r = pid_r * BLOCK_R + tl.arange(0, BLOCK_R)  # [BR]
    offs_c = pid_c * BLOCK_C + tl.arange(0, BLOCK_C)  # [BC]
    mask_r = offs_r < R
    mask_c = offs_c < C

    # Treat input as A: [Cout, Cin_g*Ksz]; row-stride ldA = Cin_g*Ksz
    ldA = R
    a_ptrs = w_in_ptr + offs_c[:, None] * ldA + offs_r[None, :]  # [BC, BR]
    a_mask = mask_c[:, None] & mask_r[None, :]
    # Cache in L2 to promote reuse across blocks that share Cout slices
    a = tl.load(a_ptrs, mask=a_mask, other=0.0, cache_modifier=".cg")

    # Output is B: [Cin_g*Ksz, Cout]; row-stride ldB = Cout
    ldB = C
    b_ptrs = w_out_ptr + offs_r[None, :] * ldB + offs_c[:, None]  # [BC, BR]
    tl.store(b_ptrs, a, mask=a_mask)


# ===========================
# Conv1D GEMM-style kernel with packed weights (3D tiling grid)
# input:  [B, Cin, Lin] contiguous
# weight_packed: [Cin_g*Ksz, Cout] contiguous
# output: [B, Cout, Lout]
# ===========================
@triton.autotune(
    configs=autotune_configs,
    key=["M", "N", "K", "stride", "dilation", "kernel_size", "Cin_g"],
)
@triton.jit
def conv1d_gemm_kernel(
    input_ptr,  
    weight_packed_ptr,  
    bias_ptr,  
    output_ptr,  
    # Dimensions (constexpr for specialization)
    B: tl.constexpr,
    Cin: tl.constexpr,
    Cout: tl.constexpr,
    Lin: tl.constexpr,
    Lout: tl.constexpr,
    G: tl.constexpr,
    # Matmul dims
    M: tl.constexpr,  # M = B * Lout
    N: tl.constexpr,  # N = out_channels_per_group
    K: tl.constexpr,  # K = Cin_g * kernel_size
    Cin_g: tl.constexpr,  # in_channels per group
    # Conv hyperparams
    kernel_size: tl.constexpr,
    stride: tl.constexpr,
    padding: tl.constexpr,
    dilation: tl.constexpr,
    has_bias: tl.constexpr,
    # Tunables (from autotuner)
    BLOCK_M: tl.constexpr,
    BLOCK_N: tl.constexpr,
    BLOCK_K: tl.constexpr,
    GROUP_M: tl.constexpr,
    WARP_SPECIALIZE: tl.constexpr,
):
    # Ensure TensorCore-friendly K tile
    tl.static_assert(BLOCK_K % 16 == 0)

    # Tile indices from 3D launch grid
    pid_m = tl.program_id(0)
    gid = tl.program_id(1)  # group id
    pid_n = tl.program_id(2)

    # Tile offsets
    offs_m = pid_m * BLOCK_M + tl.arange(0, BLOCK_M)  # [BM]
    offs_n = pid_n * BLOCK_N + tl.arange(0, BLOCK_N)  # [BN]
    mask_m = offs_m < M
    mask_n = offs_n < N

    # Decompose M -> (batch_idx, out_pos)
    batch_idx = offs_m // Lout  # [BM]
    out_pos = offs_m % Lout  # [BM]
    pos_base = out_pos * stride - padding  # [BM]

    # Group channel ranges for this gid
    in_ch_start = gid * Cin_g
    out_ch_start = gid * N

    # Strides
    batch_in_stride = Cin * Lin
    batch_out_stride = Cout * Lout

    # Precompute base pointers
    batch_in_offs = batch_idx * batch_in_stride  # [BM]
    # Base for [BM, 1] rows at first input channel of this group
    input_batch_group_base = input_ptr + batch_in_offs[:, None] + (in_ch_start * Lin)
    out_ch_idx = out_ch_start + offs_n  # [BN]

    # Weight base column pointers: [1, BN]
    w_col_base = weight_packed_ptr + out_ch_idx[None, :]

    # Accumulator in fp32
    acc = tl.zeros((BLOCK_M, BLOCK_N), dtype=tl.float32)

    mask_m_b = mask_m[:, None]
    mask_n_b = mask_n[None, :]

    # Loop over kernel taps then Cin_g in BLOCK_K chunks
    # Important: hoist row validity outside K-loop to reduce mask materialization
    for t in tl.static_range(0, kernel_size):
        tap_incr = t * dilation
        input_pos = pos_base[:, None] + tap_incr  # [BM, 1]
        # Valid rows for this tap
        row_valid_mask = (input_pos >= 0) & (input_pos < Lin)
        row_valid_mask = row_valid_mask & mask_m_b  # [BM, 1]

        # Double-buffer the (x, w) tiles across the Cin_g loop for better overlap
        # Prefetch the first chunk
        cin_off0 = 0 + tl.arange(0, BLOCK_K)
        k_mask0 = cin_off0 < Cin_g

        cin_incr0 = cin_off0 * Lin
        in_ptrs0 = input_batch_group_base + cin_incr0[None, :] + input_pos
        in_mask0 = row_valid_mask & k_mask0[None, :]
        x0 = tl.load(in_ptrs0, mask=in_mask0, other=0.0, cache_modifier=".ca")

        k_linear0 = cin_off0 * kernel_size + t
        w_ptrs0 = w_col_base + k_linear0[:, None] * Cout
        w_mask0 = k_mask0[:, None] & mask_n_b
        # Cache weights in L2 since the same Cout slice is reused for nearby tiles
        w0 = tl.load(w_ptrs0, mask=w_mask0, other=0.0, cache_modifier=".cg")

        # Process remaining chunks
        for cstart in tl.static_range(BLOCK_K, Cin_g, BLOCK_K):
            # Compute with current tiles
            acc = tl.dot(x0, w0, acc)

            # Prefetch next
            cin_off1 = cstart + tl.arange(0, BLOCK_K)
            k_mask1 = cin_off1 < Cin_g

            cin_incr1 = cin_off1 * Lin
            in_ptrs1 = input_batch_group_base + cin_incr1[None, :] + input_pos
            in_mask1 = row_valid_mask & k_mask1[None, :]
            x1 = tl.load(in_ptrs1, mask=in_mask1, other=0.0, cache_modifier=".ca")

            k_linear1 = cin_off1 * kernel_size + t
            w_ptrs1 = w_col_base + k_linear1[:, None] * Cout
            w_mask1 = k_mask1[:, None] & mask_n_b
            w1 = tl.load(w_ptrs1, mask=w_mask1, other=0.0, cache_modifier=".cg")

            # Swap buffers
            x0 = x1
            w0 = w1

        # Final dot for tail (or the only chunk if Cin_g <= BLOCK_K)
        acc = tl.dot(x0, w0, acc)

    # Bias add (fused epilogue)
    if has_bias:
        b = tl.load(bias_ptr + out_ch_idx, mask=mask_n, other=0.0).to(tl.float32)
        acc += b[None, :]

    # Store output [B, Cout, Lout] - cast to input dtype
    batch_out_offs = batch_idx * batch_out_stride
    out_ptrs = (
        output_ptr
        + batch_out_offs[:, None]
        + out_ch_idx[None, :] * Lout
        + out_pos[:, None]
    )
    out_mask = mask_m_b & mask_n_b
    # Cast accumulator back to input precision
    tl.store(out_ptrs, acc, mask=out_mask)


def _maybe_pack_weight(weight_tensor, in_channels_per_group, kernel_sz):
    """
    Packs weight from [Cout, Cin_g, Ksz] to [Cin_g*Ksz, Cout] using Triton.
    Caches the result based on data_ptr + shape + device to avoid repeated packing.
    Supports fp32, fp16, and bf16 dtypes.
    """
    Cout, Cin_g, Ksz = weight_tensor.shape
    assert Cin_g == in_channels_per_group and Ksz == kernel_sz

    # Check if caching is disabled
    if not _DISABLE_WEIGHT_CACHE:
        cache_key = (
            int(weight_tensor.data_ptr()),
            tuple(weight_tensor.shape),
            weight_tensor.device,
            weight_tensor.dtype,
        )
        packed = _WEIGHT_PACK_CACHE.get(cache_key, None)
        if (
            packed is not None
            and packed.is_cuda
            and packed.dtype == weight_tensor.dtype
        ):
            return packed

    # Allocate packed tensor with same dtype as input
    packed = torch.empty(
        (Cin_g * Ksz, Cout), device=weight_tensor.device, dtype=weight_tensor.dtype
    )

    # Launch Triton packing kernel
    BR, BC = 128, 128

    def grid(meta):
        R = Cin_g * Ksz
        C = Cout
        gr = (R + BR - 1) // BR
        gc = (C + BC - 1) // BC
        return (gr, gc)

    pack_conv1d_weight_kernel[grid](
        weight_tensor,
        packed,
        Cout,
        Cin_g,
        Ksz,
        BLOCK_R=BR,
        BLOCK_C=BC,
    )

    if not _DISABLE_WEIGHT_CACHE:
        _WEIGHT_PACK_CACHE[cache_key] = packed
    return packed


def kernel_function(
    input_tensor,
    weight_tensor,
    bias_tensor=None,
    stride=1,
    padding=1,
    dilation=1,
    groups=1,
):
    """
    Optimized Triton 1D convolution using GEMM tiling with packed weights.
    Supports fp32, fp16, and bf16 dtypes.
    """
    # Validate dtype consistency
    assert input_tensor.dtype in (
        torch.float32,
        torch.float16,
        torch.bfloat16,
    ), f"input_tensor must be fp32, fp16, or bf16, got {input_tensor.dtype}"
    assert (
        weight_tensor.dtype == input_tensor.dtype
    ), f"weight_tensor dtype {weight_tensor.dtype} must match input_tensor dtype {input_tensor.dtype}"
    if bias_tensor is not None:
        assert (
            bias_tensor.dtype == input_tensor.dtype
        ), f"bias_tensor dtype {bias_tensor.dtype} must match input_tensor dtype {input_tensor.dtype}"

    # Dimensions
    batch_size, in_channels, input_length = input_tensor.shape
    out_channels, in_channels_per_group, kernel_sz = weight_tensor.shape

    # Validate inputs
    assert in_channels_per_group * groups == in_channels, (
        f"in_channels_per_group ({in_channels_per_group}) * groups ({groups}) != "
        f"in_channels ({in_channels})"
    )
    assert (
        out_channels % groups == 0
    ), f"out_channels ({out_channels}) must be divisible by groups ({groups})"
    out_channels_per_group = out_channels // groups

    # Output length per PyTorch formula
    output_length = (
        (input_length + 2 * padding - dilation * (kernel_sz - 1) - 1) // stride
    ) + 1

    # Allocate output with same dtype as input
    output = torch.empty(
        (batch_size, out_channels, output_length),
        device=input_tensor.device,
        dtype=input_tensor.dtype,
    )

    # Bias handling
    has_bias = bias_tensor is not None
    if not has_bias:
        bias_tensor = torch.zeros(
            out_channels, device=input_tensor.device, dtype=input_tensor.dtype
        )

    # Pack weights once and cache
    weight_packed = _maybe_pack_weight(weight_tensor, in_channels_per_group, kernel_sz)

    # GEMM dims
    M = batch_size * output_length
    N = out_channels_per_group
    K = in_channels_per_group * kernel_sz

    # Grid calculation (pure Python)
    def grid(meta):
        BM = meta["BLOCK_M"]
        BN = meta["BLOCK_N"]
        gm = (M + BM - 1) // BM
        gn = (N + BN - 1) // BN
        return (gm, groups, gn)

    # Launch kernel
    conv1d_gemm_kernel[grid](
        input_tensor,
        weight_packed,
        bias_tensor,
        output,
        batch_size,
        in_channels,
        out_channels,
        input_length,
        output_length,
        groups,
        M,
        N,
        K,
        in_channels_per_group,  # Cin_g as constexpr
        kernel_sz,
        stride,
        padding,
        dilation,
        has_bias,
    )

    return output


class TritonModel(nn.Module):
    def __init__(self):
        super().__init__()

    def forward(self, input_tensor, weight_tensor, bias_tensor):
        return kernel_function(
            input_tensor,
            weight_tensor,
            bias_tensor,
            stride=1,
            padding=1,
            dilation=1,
            groups=1,
        )


def get_inputs():
    """
    Generate test inputs for the conv1d benchmark.

    Returns:
        List of input tuples for benchmarking.
    """
    # Generate synthetic production-like data
    # Shape matches original production data: (2048, 96, 200)
    # Use torch.float32 to match production data dtype
    input_full, weight_tensor, bias_tensor = generate_synthetic_production_data(
        dtype=torch.float32, device="cuda"
    )

    print("Generated synthetic production data:")
    print(f"  Input shape: {input_full.shape}, dtype: {input_full.dtype}")
    print(f"  Weight shape: {weight_tensor.shape}, dtype: {weight_tensor.dtype}")
    print(f"  Bias shape: {bias_tensor.shape}, dtype: {bias_tensor.dtype}")
    print(f"  Input is contiguous: {input_full.is_contiguous()}")

    # Create multiple batch sizes by slicing the full input
    # This matches the original production data loading approach
    batch_sizes = [64, 128, 256, 512, 1024, 2048]
    inputs = []

    for batch_size in batch_sizes:
        # Slice input to desired batch size
        input_tensor = input_full[:batch_size, :, :]
        if not input_tensor.is_contiguous():
            input_tensor = input_tensor.contiguous()
        inputs.append((input_tensor, weight_tensor, bias_tensor))
    return inputs


# ============================================================================
# TritonBench Operator Integration
# ============================================================================


class Operator(BenchmarkOperator):
    """
    TritonBench operator for conv1d.

    This operator wraps the kernel_evolve generated TritonModel and PytorchModel
    into the TritonBench evaluation framework.
    """

    DEFAULT_METRICS = ["latency"]
    DEFAULT_PRECISION = "fp32"

    def __init__(
        self,
        tb_args: argparse.Namespace,
        extra_args: Optional[List[str]] = None,
    ) -> None:
        super().__init__(tb_args, extra_args)

        self.pytorch_model = PytorchModel().to(self.device).eval()
        self.triton_model = TritonModel().to(self.device).eval()

    def get_input_iter(self) -> Generator:
        """
        Generate test inputs using the shared get_inputs() function.

        Note: get_inputs() is an independent function that both models access.
        This ensures consistent test data generation across all implementations.
        """
        inputs = get_inputs()

        def to_device_dtype(x):
            if isinstance(x, torch.Tensor):
                return x.to(device=self.device, dtype=self.dtype)
            return x

        for input_tuple in inputs:
            converted_tuple = tuple(to_device_dtype(tensor) for tensor in input_tuple)
            yield converted_tuple

    def get_x_val(self, example_inputs: Any) -> str:
        """Extract x-axis value for plotting."""
        if not example_inputs or len(example_inputs) == 0:
            return "0"

        first_input = example_inputs[0]
        if isinstance(first_input, torch.Tensor):
            shape = first_input.shape

            if not shape:
                return "1"

            return "x".join(map(str, shape[: len(shape)]))

        return "Unknown"

    @register_benchmark(baseline=True, operator_name="conv1d")
    def pytorch_reference(self, *inputs) -> Callable:
        """
        PyTorch reference implementation (baseline).

        Uses the PytorchModel generated by kernel_evolve.
        """
        model = torch.compile(self.pytorch_model, mode="max-autotune-no-cudagraphs")

        def _impl():
            with torch.no_grad():
                return model(*inputs)

        return _impl

    @register_benchmark(baseline=False, operator_name="conv1d")
    def pytorch_reference_conv2d(self, *inputs) -> Callable:
        """
        PyTorch reference implementation (baseline).

        Uses the PytorchModel generated by kernel_evolve.
        """
        model = torch.compile(self.pytorch_model, mode="max-autotune-no-cudagraphs")

        def _impl():
            with torch.no_grad():
                return model(*inputs, use_conv2d=True)

        return _impl

    @register_benchmark(operator_name="conv1d")
    def triton_kernel(self, *inputs) -> Callable:
        """
        Triton kernel implementation.

        Uses the TritonModel generated by kernel_evolve.
        """
        model = torch.compile(self.triton_model, mode="max-autotune-no-cudagraphs")

        def _impl():
            with torch.no_grad():
                return model(*inputs)

        return _impl


# ============================================================================
# Evaluation Entry Point
# ============================================================================


def run():
    """Run benchmark with latency, speedup, and accuracy evaluation."""

    try:
        from tritonbench.utils.parser import get_parser

        override_args = [
            "--device",
            "cuda",
            "--mode",
            "fwd",
            "--metrics",
            "latency,speedup,accuracy",
            "--precision",
            "fp16",
            "--benchmark-name",
            "conv1d",
            "--atol",
            "1e-4",
            "--rtol",
            "5e-4",
        ]

        parser = get_parser()
        tb_args, extra_args = parser.parse_known_args(override_args)

        op = Operator(tb_args=tb_args, extra_args=extra_args)
        print("Running TritonBench conv1d benchmark...")
        op.run()

        # Print TritonBench output table
        print("\n" + str(op.output))

        # Also print formatted performance table with explicit headers
        if hasattr(op, "output") and op.output:
            print("\n" + "=" * 80)
            print("PERFORMANCE SUMMARY TABLE")
            print("=" * 80)

            ub_dict = op.output.userbenchmark_dict
            benchmark_name = op.output.benchmark_name
            x_vals = op.output.x_vals

            # Print table header
            print(
                f"{'x_val':<15} {'pytorch_reference-latency':<30} {'triton_kernel-latency':<30} {'triton_kernel-speedup':<25}"
            )
            print("-" * 100)

            # Print data rows
            for x_val in x_vals:
                latency_key = (
                    f"tritonbench_{benchmark_name}[x_{x_val}-triton_kernel]_latency"
                )
                speedup_key = (
                    f"tritonbench_{benchmark_name}[x_{x_val}-triton_kernel]_speedup"
                )
                ref_latency_key = (
                    f"tritonbench_{benchmark_name}[x_{x_val}-pytorch_reference]_latency"
                )

                triton_latency = ub_dict.get(latency_key, "N/A")
                speedup = ub_dict.get(speedup_key, "N/A")
                ref_latency = ub_dict.get(ref_latency_key, "N/A")

                print(
                    f"{x_val:<15} {str(ref_latency):<30} {str(triton_latency):<30} {str(speedup):<25}"
                )

            print("=" * 80 + "\n")
    except Exception as e:
        print(f"ERROR: Test failed with exception: {e}")
        print("REFERENCE_TEST_FAILED")
        print("KERNEL_TEST_FAILED")
        print("FITNESS_SCORE: 0.0")
        return False

    # Validate accuracy and report metrics
    try:
        if not hasattr(op, "output") or not op.output:
            print("ERROR: No output available from benchmark")
            print("KERNEL_TEST_FAILED")
            print("FITNESS_SCORE: 0.0")
            return False

        ub_dict = op.output.userbenchmark_dict
        benchmark_name = op.output.benchmark_name
        x_vals = op.output.x_vals

        print("\n" + "=" * 80)
        print("DETAILED METRICS PER INPUT SIZE")
        print("=" * 80)

        accuracy_passed = True
        all_latencies = []
        all_speedups = []

        for x_val in x_vals:
            latency_key = (
                f"tritonbench_{benchmark_name}[x_{x_val}-triton_kernel]_latency"
            )
            speedup_key = (
                f"tritonbench_{benchmark_name}[x_{x_val}-triton_kernel]_speedup"
            )
            accuracy_key = (
                f"tritonbench_{benchmark_name}[x_{x_val}-triton_kernel]_accuracy"
            )

            latency = ub_dict.get(latency_key)
            speedup = ub_dict.get(speedup_key)
            accuracy = ub_dict.get(accuracy_key)

            print(f"\nInput size: {x_val}")
            print(f"  Latency: {latency}")
            print(f"  Speedup: {speedup}")
            print(f"  Accuracy: {accuracy}")

            if latency is not None:
                all_latencies.append(latency)
            if speedup is not None:
                all_speedups.append(speedup)

            if accuracy is not None and accuracy == 0.0:
                accuracy_passed = False
                print(f"  ACCURACY FAILED for input size {x_val}")

        print("\n" + "=" * 80)
        print("SUMMARY STATISTICS")
        print("=" * 80)

        avg_latency_key = f"tritonbench_{benchmark_name}[triton_kernel]-latency-avg"
        avg_speedup_key = f"tritonbench_{benchmark_name}[triton_kernel]-speedup-avg"

        avg_latency = ub_dict.get(avg_latency_key)
        avg_speedup = ub_dict.get(avg_speedup_key)

        print(f"Average Latency: {avg_latency}")
        print(f"Average Speedup: {avg_speedup}")
        print(
            f"Accuracy: {'PASS' if accuracy_passed else 'FAIL'} (all input sizes must pass)"
        )
        print("=" * 80 + "\n")

        if not accuracy_passed:
            print(
                "ERROR: NUMERICAL MISMATCH BETWEEN TRITON KERNEL AND PYTORCH REFERENCE"
            )
            print("DEBUG: One or more input sizes failed accuracy check")
            print("KERNEL_TEST_FAILED")
            print("FITNESS_SCORE: 0.0")
            return False

        print(
            "SUCCESS: Triton kernel output matches PyTorch reference implementation"
        )
        print("KERNEL_TEST_PASSED")

        fitness_score = 1.0
        if avg_speedup is not None and avg_speedup > 0:
            fitness_score = max(0.1, min(avg_speedup, 10.0))

        print(f"FITNESS_SCORE: {fitness_score}")

        return True

    except Exception as e:
        print(f"ERROR: Test failed with exception: {e}")
        print("KERNEL_TEST_FAILED")
        print("FITNESS_SCORE: 0.0")
        return False


if __name__ == "__main__":
    success = run()
    sys.exit(0 if success else 1)
\end{pythoncode}

\end{document}